\documentclass[times,twocolumn,final]{elsarticle}

\newcommand{\eqal}[2]{\\\begin{equation}\begin{aligned} \label{eq:#1} #2 \end{aligned}\end{equation}\\}
\newcommand{\Tone}{T\textsubscript{1}}
\newcommand{\Ttwo}{T\textsubscript{2}}
\newcommand{\Tonec}{T\textsubscript{1c}}

\newcommand{\TtwoPDTone}{T\textsubscript{2},~PD~$\rightarrow$~T\textsubscript{1}}
\newcommand{\TonePDTtwo}{T\textsubscript{1},~PD~$\rightarrow$~T\textsubscript{2}}
\newcommand{\ToneTtwoPD}{T\textsubscript{1},~T\textsubscript{2}~$\rightarrow$~PD}
\newcommand{\ToneTtwoFlairTonec}{T\textsubscript{1},~T\textsubscript{2},~FLAIR~$\rightarrow$~T\textsubscript{1c}}
\newcommand{\ToneTtwoTonecFlair}{T\textsubscript{1},~T\textsubscript{2},~T\textsubscript{1c}~$\rightarrow$~FLAIR}
\newcommand{\ToneFlairTonecTtwo}{T\textsubscript{1},~FLAIR,~T\textsubscript{1c}~$\rightarrow$~T\textsubscript{2}}
\newcommand{\TtwoFlairTonecTone}{T\textsubscript{2},~FLAIR,~T\textsubscript{1c}~$\rightarrow$~T\textsubscript{1}}
\newcommand{\ASC}{A~$\rightarrow$~S~$\rightarrow$~C}
\newcommand{\ACS}{A~$\rightarrow$~C~$\rightarrow$~S}
\newcommand{\SCA}{S~$\rightarrow$~C~$\rightarrow$~A}
\newcommand{\SAC}{S~$\rightarrow$~A~$\rightarrow$~C}
\newcommand{\CSA}{C~$\rightarrow$~S~$\rightarrow$~A}
\newcommand{\CAS}{C~$\rightarrow$~A~$\rightarrow$~S}

\usepackage{medima}

\usepackage[switch,columnwise]{lineno}
\usepackage[colorlinks]{hyperref}
\usepackage{amssymb}
\usepackage{latexsym}
\usepackage{float}
\usepackage{siunitx}
\usepackage{gensymb}
\usepackage{xcolor}
\usepackage{ragged2e}
\usepackage[utf8]{inputenc} 
\usepackage{url}            
\usepackage{amsfonts}       
\usepackage{lipsum}
\usepackage{amsmath}
\usepackage{framed,multirow}
\usepackage{xcolor}
\usepackage{subcaption,booktabs}
\graphicspath{{./manuscript_figures/}}
\definecolor{mint}{rgb}{0.24, 0.71, 0.54}
\definecolor{newcolor}{rgb}{.8,.349,.1}

\journal{Preprint}

\let\oldequation\equation
\let\oldendequation\endequation

\renewenvironment{equation}
  {\linenomathNonumbers\oldequation}
  {\oldendequation\endlinenomath}

\hypersetup{
  colorlinks,
  citecolor=Red,
  linkcolor=Red,
  urlcolor=Red}
\begin{document}

\verso{Yurt and Ozbey \textit{et~al.}}

\begin{frontmatter}

\title{Progressively Volumetrized Deep Generative Models for Data-Efficient Contextual Learning of MR Image Recovery}%

\author[1,2,*]{Mahmut Yurt}
\author[1,2,*]{Muzaffer Özbey}
\author[1,2]{Salman UH Dar}
\author[1,2,3]{Berk Tınaz}
\author[2,4]{Kader K Oguz}
\author[1,2,5]{Tolga Çukur\corref{cor1}}
\cortext[cor1]{Corresponding author, 
  e-mail: cukur@ee.bilkent.edu.tr  }
\address[1]{Department of Electrical and Electronics Engineering, Bilkent University, Ankara 06800, Turkey}
\address[2]{National Magnetic Resonance Research Center (UMRAM), Bilkent University, Ankara 06800, Turkey}
\address[3]{Department of Electrical and Computer Engineering, University of Southern California, Los Angeles 90089, USA}
\address[4]{Department of Radiology, Hacettepe University, Ankara 06100, Turkey}
\address[5]{Neuroscience Program, Bilkent University, Ankara 06800, Turkey}
\address[*]{denotes equal contribution}

\begin{abstract}
Magnetic resonance imaging (MRI) offers the flexibility to image a given anatomic volume under a multitude of tissue contrasts. Yet, scan time considerations put stringent limits on the quality and diversity of MRI data. The gold-standard approach to alleviate this limitation is to recover high-quality images from data undersampled across various dimensions, most commonly the Fourier domain or contrast sets. A primary distinction among recovery methods is whether the anatomy is processed per volume or per cross-section.  Volumetric models offer enhanced capture of global contextual information, but they can suffer from suboptimal learning due to elevated model complexity. Cross-sectional models with lower complexity offer improved learning behavior, yet they ignore contextual information across the longitudinal dimension of the volume.  Here, we introduce a novel progressive volumetrization strategy for generative models (ProvoGAN) that serially decomposes complex volumetric image recovery tasks into successive cross-sectional mappings task-optimally ordered across individual rectilinear dimensions. ProvoGAN effectively captures global context and recovers fine-structural details across all dimensions, while maintaining low model complexity and improved learning behaviour. Comprehensive demonstrations on mainstream MRI reconstruction and synthesis tasks show that ProvoGAN yields superior performance to state-of-the-art volumetric and cross-sectional models.
\end{abstract}


\end{frontmatter}

\newcommand\tab[1][0.8cm]{\hspace*{#1}}
\newcommand\tabx[1][0.6cm]{\hspace*{#1}}
\newcommand{\MSaccess}{}

\section{Introduction}
Magnetic resonance imaging (MRI) is a clinically preferred modality that produces volumetric images of a given anatomy under diverse tissue contrasts \citep{intro_ref}. As MR acquisitions are intrinsically slow, there has been persistent interest in recovery methods to improve quality and diversity of images derived from accelerated imaging protocols \citep{intro_jcye,jcye_survey}. Two mainstream MRI recovery problems with pervasive applications are reconstruction and synthesis \citep{Griswold2002,Pruessmann1999,lustig2008,patch_based_one_to_one_4,mustgan,loc_sens_nn_1,patch_based_one_to_one_2,example_based}. While reconstruction aims to recover high-quality images from undersampled k-space acquisitions \citep{lustig2007}, synthesis aims to recover high-quality images of unacquired tissue contrasts from images of collected contrasts \citep{com_sen_mr_tissue}. Learning-based models have offered performance leaps in both recovery tasks, given their ability to solve inverse problems \citep{yi2019generative,litjens_survey,choi2020stargan}. However, the trade-off between sensitivity to spatial context and model complexity introduces a dilemma regarding the use of volumetric versus cross-sectional recovery models \citep{3Dreview}. The primary aim of this study is to introduce a novel volumetrization approach to achieve the contextual sensitivity of volumetric models while maintaining on par complexity with cross-sectional models.   

\par
Among learning-based models, a native recovery approach is to perform a single-shot global mapping between source and target volumes \citep{str_artifacts,3d_rec_1,3d_synth_2,3D_cgan,eagan,chen2021synthesizing,sood20213d,kustner2020cinenet,el2020deep,chong2021synthesis}. Volumetric models leverage spatial correlations across all dimensions to better capture contextual information \citep{str_artifacts,3d_rec_1,3d_synth_2,3D_cgan}. Introduction of these contextual priors can theoretically lead to more consistent and accurate recovery across the volume. However, three-dimensional (3D) models involve substantially more parameters than their two-dimensional (2D) counterparts \citep{3Dreview,mustgan}. Furthermore, each volume constitutes a single training sample for a 3D model, whereas it would yield several tens of samples for a 2D model. Taken together, these factors render heavier demand for training data and impair the learning process for volumetric models \citep{3Dreview}.

\par
A less demanding approach in terms of training data for learning-based MRI recovery is to perform a spatially-localized mapping between individual cross-sections \citep{Akcakaya2019,chartsias2018c,Cheng2018,assess_importance,Mardani2019b,Mardani2017,zhan2021,Joyce2017c,chen2021wavelet,li2021modified,zhan2021lr,wang2021review,Hyun2018,assess_importance,Quan2018c,Schlemper2017}. Volumes are split along a specific rectilinear orientation, and cross-sectional models are then trained to learn the 2D mapping \citep{Dar2019,darsynergistic,Hammernik2017,Han2018,collagan,nn_one_to_one_2}. Since a lower-dimensional mapping is to be learned, cross-sectional models are of lower complexity and have reduced demand for training data \citep{mustgan}. This facilitates the learning process, and often results in more detailed mappings along the transverse dimensions within cross-sections compared to 3D models. Yet, 2D models do not fully utilize context across the longitudinal dimension, even when simultaneously processing multiple neighboring cross-sections \citep{3d_synth_2,3D_cgan,Dar2019}. This results in inconsistency and errors across separately recovered cross-sectional images \citep{str_artifacts,3d_rec_1,eagan}.

\par
An effective alternative to either approach is to build hybrid architectures that bridge 2D and 3D models. A group of studies in this domain have proposed aggregated models that fuse the outputs of parallel streams, where the streams are cross-sectional models in three orthogonal orientations \citep{volume_volume_fuse,provolike_1}. Pseudo target volumes are first recovered separately by the 2D streams, and a 3D fusion network then produces the final target volume \citep{volume_volume_fuse,provolike_1}. Other studies have instead proposed transfer of learned model weights from 2D to 3D models \citep{shan20183,liu20183d}. A 2D model is first pretrained for a cross-sectional recovery task at a selected orientation, the learned weights are then used to initialize the convolutional kernels in 3D models \citep{shan20183,liu20183d}. While both approaches can improve learning behavior, they involve a volumetric processing component that elevates memory requirements and places practical constraints on model complexity, potentially limiting sensitivity to detailed image features. 

\par
\begin{figure*}[!t]
\centering
\includegraphics[width=0.85\textwidth]{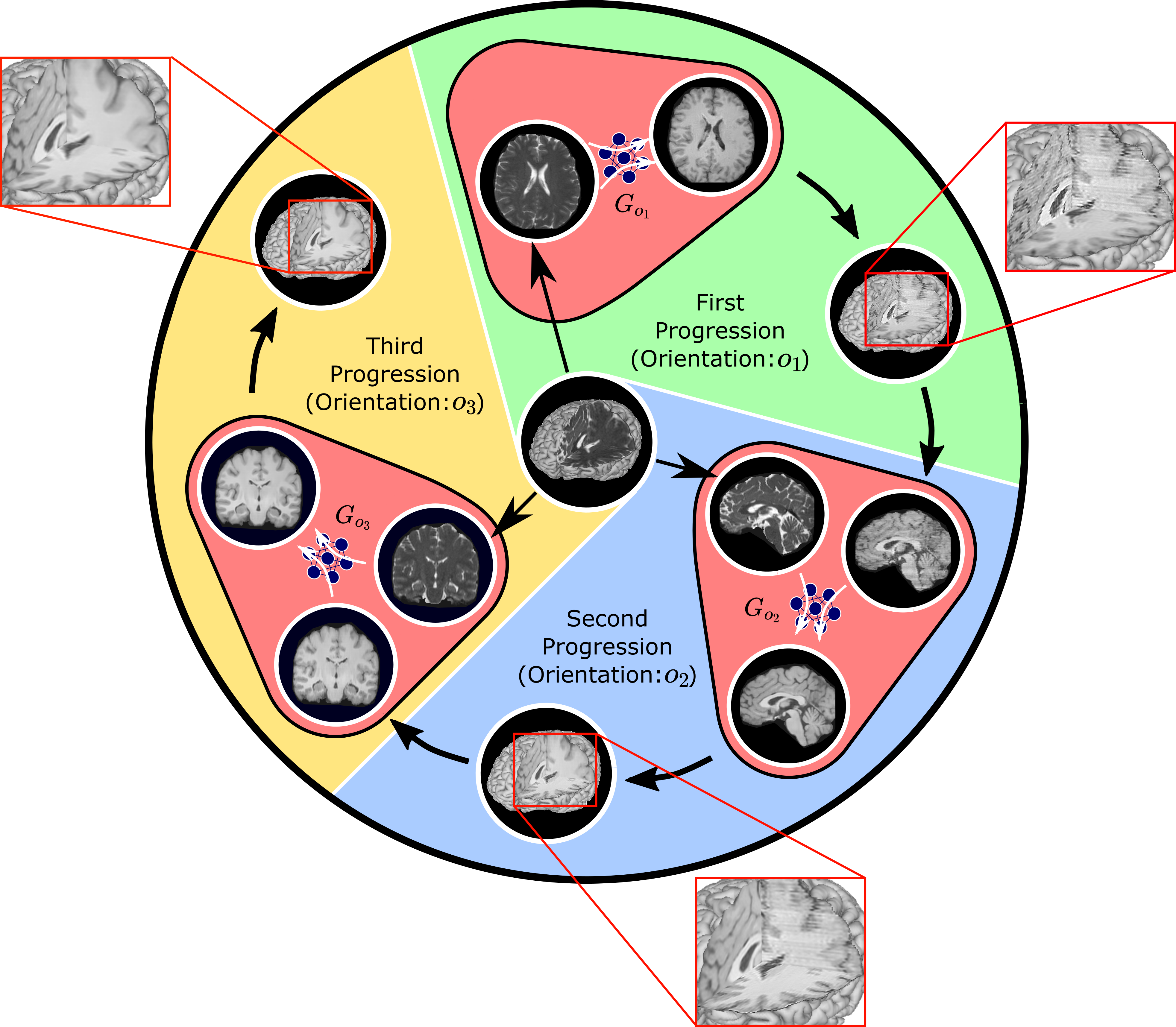}
\caption{ProvoGAN decomposes complex volumetric image recovery tasks into a cascade of progressive cross-sectional subtasks defined across the rectilinear orientations (axial, coronal, and sagittal). Given a specific order of progression sequence (axial $\rightarrow$ sagittal $\rightarrow$ coronal is given here for demonstration), ProvoGAN first learns a cross-sectional mapping in the first orientation, and processes cross-sections within the entire source volume to estimate the target volume. This volumetric estimate is then divided into cross-sections in the second orientation, and a separate cross-sectional model is learned in the second orientation. The volumetric estimate from the second progression is then fed onto the final progression in which a third cross-sectional model is learned for final recovery (see Supp. Fig. 1 for further details). The sequential implementation of the progressive cross-sectional models enables ProvoGAN to gradually improve capture of fine-structural details in each orientation and to ensure global contextual consistency within the volume while at the same time manifesting reduced model complexity and improved learning behaviour of cross-sectional mapping.}
\label{fig:main_fig}
\end{figure*}

Here, we propose a novel progressive volumetrization strategy for deep generative models (ProvoGAN) for contextual learning of MR image recovery. To improve learning efficiency by lowering model complexity, ProvoGAN serially decomposes volumetric recovery tasks into a sequence of cross-sectional subtasks (e.g., axial, coronal, sagittal) for the first time in literature.~\footnote{We presented a preliminary conception of the idea in the IEEE International Symposium on Biomedical Imaging (ISBI) on April 4, 2020.} For a given subtask in a  selected  orientation, the source volume is split across the respective longitudinal dimension, and a 2D model is trained to map between cross-sectional source and target images. The predicted pseudo cross-sections are reformatted into a volume and then input to the next subtask as spatial priors (Fig. \ref{fig:main_fig}). This progressive nature empowers ProvoGAN to recover fine-structural details in each orientation while ensuring contextual consistency across the volume. Furthermore, the progression order of the subtasks is adaptively optimized to enhance task-specific performance.  To ensure a high degree of realism, we primarily employ ProvoGAN to volumetrize a recent conditional generative adversarial network based on the ResNet architecture \citep{Dar2019}. Note that ProvoGAN can be viewed as a model-agnostic strategy, so it can be extended to volumetrize other 2D network models as also demonstrated here. Comprehensive demonstrations are provided for mainstream reconstruction and synthesis tasks in multi-contrast MRI protocols. Our results indicate that ProvoGAN yields enhanced recovery performance compared to cross-sectional, volumetric, and hybrid approaches in terms of image quality. Importantly, ProvoGAN maintains these performance benefits while at the same time offering reduced model complexity and improved learning behaviour. 

\subsection*{Contributions}
\begin{itemize}
    \item To our knowledge, ProvoGAN is the first volumetrized model for MRI recovery that serially decomposes a global 3D mapping into a sequence of progressive 2D mappings. 
    \item ProvoGAN maximizes task performance via adaptive ordering of the progression sequence of 2D mappings across rectilinear orientations. 
    \item ProvoGAN embodies a model-agnostic learning strategy, so it can be implemented to volumetrize various 2D network architectures.
    \item Demonstrations on mainstream reconstruction and synthesis tasks indicate that ProvoGAN yields superior performance to several prior 2D, 3D and hybrid models. 
\end{itemize}

\section{Methods}
	
\subsection{Generative Adversarial Networks}
Generative adversarial networks (GAN) are generative models composed of two subnetworks. The first subnetwork is a generator ($G$) that aims to synthesize fake samples closely mimicking a target data distribution, while the second subnetwork is a discriminator ($D$) that aims to detect whether a given data sample has been drawn from the target distribution or not \citep{Goodfellow2014a}. These subnetworks are trained alternately in a two player zero-sum min-max game in an adversarial setup:
\eqal{GAN_Equation}{
	L_{\mathrm{GAN}}=\mathrm{E}_y[\mathrm{log}(D(y))]+\mathrm{E}_z[\mathrm{log}(1-D(G(z)))]
}
where $L_{\mathrm{GAN}}$ is the adversarial loss function, $\mathrm{E}$ denotes expectation, $z$ denotes a random noise vector sampled from a prior distribution, and $y$ denotes an arbitrary real sample drawn from the target domain. In practice, the log-likelihood terms are replaced with squared-loss terms to improve stability \citep{lsgan}:
\eqal{GAN_Equation_2}{
	L_{\mathrm{GAN}}=-\mathrm{E}_y[(D(y)-1)^2]-\mathrm{E}_z[D(G(z))^2]
}
where $D$ is trained to maximize $L_{\mathrm{GAN}}$, whereas $G$ is trained to minimize it. 
\par
While the basic GAN model synthesizes target data samples given a random noise input, recent studies on computer vision \citep{pix2pix,cycleGAN} and medical imaging \citep{Dar2019,collagan,diamondgan,mra_synth,mmgan,mr_tra_seg,3D_cgan,eagan} have demonstrated that conditional GAN (cGAN) models \citep{condgans} are highly effective in image-to-image translation tasks. The central aim in these tasks is to synthesize data samples from the target image domain, given data samples from a separate source image domain. The cGAN model is therefore modified to condition both $G$ and $D$ on the source domain image: 
\eqal{cGAN_Equation}{
	L_{\mathrm{cGAN}}=-\mathrm{E}_{x,y}[(D(x,y)-1)^2]-\mathrm{E}_x[D(x,G(x))^2]
}
where $x$ denotes the source domain image, and $y$ denotes the target domain image. When paired images from the source and target domains are available, a pixel-wise loss between the ground truth and synthesized images can also be included:
\eqal{cGAN_Equation_with_pixel_wise_loss}{
	L_{\mathrm{cGAN}}=&-\mathrm{E}_{x,y}[(D(x,y)-1)^2]-\mathrm{E}_x[D(x,G(x))^2]\\&+\mathrm{E}_{x,y}[||y-G(x)||_1]
}
The pixel-wise loss is typically based on the mean-absolute error to reduce sensitivity to outliers and alleviate undesirable smoothing. The mapping learned by the cGAN model grows more accurate as the statistical dependence between source and target domains gets stronger \citep{mustgan}.

\subsection{MR Image Recovery via Volumetric GANs}
\label{vGAN_section}
As MR images are intrinsically volumetric, a comprehensive approach for 3D MR image recovery is to use volumetric GAN (vGAN) models that perform a global mapping between source and target volumes \citep{str_artifacts,3d_rec_1,3d_synth_2,3D_cgan}. To learn this mapping, vGAN models commonly employ complex generator $G_V$ and discriminator $D_V$ modules containing 3D convolutional kernels. The loss function is defined over the entire volume in an adversarial setup: 
\eqal{}{
	L_{\mathrm{vGAN}}=&-\mathrm{E}_{X,Y}[(D_V(X,Y)-1)^2]-\mathrm{E}_{X}[D_V(X,G_V(X))^2]\\&+\mathrm{E}_{X,Y}[||Y-G_V(X)||_1]
}
where $X$ denotes the source and $Y$ denotes the target volumetric images. For MRI reconstruction, $X$ is typically the Fourier reconstruction of undersampled acquisitions, and $Y$ is the fully-sampled reference volume. For MRI synthesis, $X$ is the source contrast volume, and $Y$ is the target contrast volume. Note that, in MRI reconstructions, an additional constraint is introduced to enforce consistency of acquired and recovered k-space data:
\eqal{}{
	F_u(G_V(X)):=F_u(X)
}
where $F_u$ denotes the partial Fourier operator that is defined at the acquired k-space points.
\par
Due to their 3D nature, vGAN models can better incorporate contextual information across MRI volumes by leveraging spatial correlations across separate cross-sections  \citep{3d_synth_2,3D_cgan,eagan}. This contextual prior can lead to elevated consistency across the volume and increased accuracy in recovery performance. That said, learning in 3D network models is inherently more difficult since they involve substantially more parameters \citep{3Dreview}. The learning process might be further impaired by data scarcity as the entire volume of each subject is taken as a single training sample \citep{3Dreview}. These limitations often cause vGAN models to settle on suboptimal parameter sets, compromising recovery performance.

\subsection{MR Image Recovery via Cross-Sectional GANs}
\label{sGAN_section}
A more focused approach for 3D MRI recovery is based on cross-sectional GAN (sGAN) models that perform localized mappings between 2D cross-sectional images within source and target volumes \citep{Dar2019,mustgan,shin2018medical,mmgan}. These 2D images are typically taken to be individual cross-sections within the volume in a specific rectilinear orientation, i.e., axial, sagittal or coronal. To learn this 2D mapping, sGAN models employ relatively simpler generator $G_S$ and discriminator $D_S$ modules containing 2D convolutional kernels. The loss function is defined for individual cross-sections in an adversarial setup with a pixel-wise loss: 
\eqal{}{
	L_{\mathrm{sGAN}}=&-\mathrm{E}_{x_o^i,y_o^i}[(D_{S}(x_o^i,y_o^i)-1)^2]-\mathrm{E}_{x_o^i}[D_{S}(x_o^i,G_{S}(x_o^i))^2]\\&+\mathrm{E}_{x_o^i,y_o^i}[||y_o^i-G_{S}(x_o^i)||_1]
}
where $x_o^i$ and $y_o^i$ denote the $i$th cross-sections within the source and target volumes in orientation $o$. As with sGAN models, $x_o^i$-$y_o^i$ are taken as cross-sectional images for undersampled and fully-sampled acquisitions in MRI reconstruction, and $x_o^i$-$y_o^i$ are taken as cross-sectional images of source and target contrasts in MRI synthesis. Consistency between acquired and recovered data can again be enforced during reconstruction via the following procedure: 
\eqal{}{
	F_u(G_S(x_o^i)):=F_u(x_o^i)
}
where $F_u$ denotes the partial Fourier operator that is defined at the acquired k-space points. Once the mapping between the source and target cross-sections is learned, cross-sections of the target volumes are independently generated, and then the target volumes are recovered by concatenating the generated cross-sections. 
\par
Due to their 2D nature, sGAN models are less complex and so they naturally have lower demand for data \cite{mustgan}. Individual cross-sections within a subject's volume are taken as separate training samples, expanding the effective size of the dataset. As a result, more detailed cross-sectional mapping can be learned. However, this advantage comes at the expense of neglecting global contextual information across the volume \citep{3d_synth_2,3D_cgan,eagan}. Therefore, sGAN models might suffer from inconsistency or inaccuracy of recovered images across cross-sections.

\subsection{Progressively Volumetrized GAN}
Here, a novel architecture is proposed to address the limitations of volumetric and cross-sectional GAN models. The proposed model, named progressively volumetrized GAN (ProvoGAN), decomposes complex volumetric image recovery tasks into a series of simpler cross-sectional tasks (Fig. \ref{fig:main_fig}). The cross-sectional recovery tasks are defined in separate orientations, and are implemented sequentially via cascaded 2D GAN models. We consider rectilinear cross-sections of volumetric MRI datasets in this study, so the selected orientations are axial, coronal and sagittal. Given a specific order of the three orientations ($o_1$, $o_2$, $o_3$), ProvoGAN first learns a 2D recovery model in orientation $o_1$. The entire source volume is processed by this model to estimate the target volume. Afterwards, this volumetric estimate is separated into cross-sections in orientation $o_2$, and a separate 2D recovery model is trained. The estimated target volume for $o_2$ is then fed onto the final stage, where a third 2D recovery model is trained in orientation $o_3$.

\par
The cascaded 2D models in ProvoGAN are trained sequentially in the three rectilinear orientations, where the 2D model weights at earlier orientations are frozen upon training. This learning strategy empowers ProvoGAN to progressively recover fine-structural details at each orientation, while bypassing the need for computationally expensive calculation of error gradients across the entire volume and across all orientations. Therefore, ProvoGAN offers the ability to efficiently capture global contextual information without drastically elevating computational demand. At the same time, this step-wise training can increase sensitivity to progression order. Therefore, progression order across orientations is adaptively tuned to maximize performance in specific tasks. Detailed formulation of the ProvoGAN model is provided below. 

\textbf{\\~\\First Progression:} 
ProvoGAN first learns a cross-sectional mapping between the source-target volumes in $o_1$ via a generator ($G_{o_1}$) and a discriminator ($D_{o_1}$). The source and target cross-sections in $o_1$ are extracted with a division block ($d_{o_1})$.
\eqal{}{
	x_{o_1}^i\in\{x_{o_1}^1,x_{o_1}^2\dots,x_{o_1}^I\}=d_{o_1}(X)\\
	y_{o_1}^i\in\{y_{o_1}^1,y_{o_1}^2\dots,y_{o_1}^I\}=d_{o_1}(Y)
}
where $X$ denotes the source volume, $Y$ denotes the target volume, $x_{o_1}^i$ denotes the $i$th cross-section of the source volume in $o_1$, $y_{o_1}^i$ denotes the $i$th cross-section of the target volume in $o_1$, and $I$ denotes the total number of cross-sections within the volumes in $o_1$. $G_{o_1}$ then learns to recover the cross-sections of the target volume from the corresponding cross-sections of the source volume.
\eqal{}{
	\hat{y}_{p_1,o_1}^i = G_{o_1}(x^i_{{o_1}})
}
where $\hat{y}_{p_1, o_1}^i$ denotes the $i$th cross-section of the target volume in $o_1$ recovered via the first progression. Meanwhile, $D_{o_1}$ learns to distinguish between the real and fake cross-sections.
\eqal{DEquation1}{
	D_{o_1}(x_{o_1}^i, m)\in[0,1]
}
where $m$ is either generated ($\hat{y}_{p_1,o_1}^i$) or ground truth ($y_{o_1}^i$) target cross-section. To simultaneously train $G_{o_1}$ and $D_{o_1}$, a loss function ($L_{o_1}$) consisting of adversarial and pixel-wise losses is used.
\eqal{}{
	L_{{o_1}}=&-\mathrm{E}_{x^i_{o_1},y^i_{o_1}}[(D_{o_1}(x^i_{o_1},y^i_{o_1})-1)^2]\\&-\mathrm{E}_{x^i_{o_1}}[D_{o_1}(x^i_{o_1},G_{o_1}(x^i_{o_1}))^2]\\&+\mathrm{E}_{x^i_{o_1},y^i_{o_1}}[||y^i_{o_1}-G_{o_1}(x^i_{o_1})||_1]
}
Once $G_{o_1}$ and $D_{o_1}$ are properly trained, cross-sections in $o_1$ for the target volume are independently generated, and then combined with a concatenation block ($c_{o_1}$) to recover the entire target volume.
\eqal{}{
	\hat{Y}_{p_1} = c_{o_1}(\hat{y}_{p_1,o_1}^1,\dots,\hat{y}_{p_1,o_1}^I) 
}
where $\hat{Y}_{p_1}$ denotes the target volume recovered after the first progression.
\\~\\
\textbf{Second Progression: }Having learned the cross-sectional mapping in $o_1$, ProvoGAN then learns a separate recovery model in the second orientation $o_2$ to gradually enhance capture of fine-structural details and spatial correlations. The prediction for the target volume generated in the first progression is also incorporated as an input to the generator $G_{o_2}$ to leverage global contextual priors.
\eqal{}{
	\hat{y}_{p_2,o_2}^j = G_{o_2}(x^j_{o_2},\hat{y}_{p_1,o_2}^j)
}
where $x_{o_2}^j$ denotes the $j$th cross-section of the source volume in ${o_2}$, $\hat{y}_{p_1,o_2}^j$ denotes the $j$th cross-section in $o_2$ of the target volume recovered in the first progression, and $\hat{y}_{{p_2},{o_2}}^j $ denotes the $j$th cross-section in $o_2$ of the target volume recovered in the second progression. Meanwhile, discriminator $D_{o_2}$ learns to distinguish between the generated and real cross-sections. 
\\~\\
\textbf{Third Progression: }Lastly, ProvoGAN learns a cross-sectional mapping in the third orientation $o_3$. As in the second progression, the prediction from the previous progression is incorporated into the mapping as prior information. Therefore, the third generator $G_{o_3}$ receives as input the cross-sections in $o_3$ of the source volume and the previously recovered volume:
\eqal{}{
	\hat{y}_{{p_3},{o_3}}^k = G_{o_3}(x^k_{{o_3}},\hat{y}_{{p_2},{o_3}}^k)
}
where $x^k_{{o_3}}$ denotes the $k$th cross-section of the source volume in $o_3$, $\hat{y}_{p_2,o_3}^k$ denotes the $k$th cross-section in $o_3$ of the target volume recovered in the second progression, and $\hat{y}_{{p_3},{o_3}}^k $ denotes the $k$th cross-section in $o_3$ of the target volume recovered in the third progression. Meanwhile, discriminator $D_{o_3}$ learns to distinguish between the generated and real cross-sections. The final output volume $\hat{Y}_{p_3}$ of the proposed method is recovered by combining the generated cross-sections in $o_3$ via a concatenation block $c_{o_3}$:
\eqal{}{
	\hat{Y}_{p_3} = c_{o_3}(\hat{y}_{p_3,o_3}^1,\dots,\hat{
		y}_{p_3,o_3}^K) 
}
where $K$ denotes the total number of cross-sections in $o_3$. Note that, in MRI reconstruction, an additional constraint is introduced after each progression to enforce consistency of the acquired and recovered k-space data via the following~procedure.
\eqal{}{
	F_u(\hat{Y}_{p_n}):=F_u(X)
} 
where $F_u$ denotes the partial Fourier operator defined on the sampling mask utilized to acquire $X$, and $n$ denotes the ongoing progression index. Meanwhile, an additional consistency between the progressions is enforced in the form of residual learning for MRI synthesis, where the generator models in the second and third progressions learn to predict the cross-sectional residuals between the target volume and the previously synthesized target volume.
\eqal{}{
	\hat{y}_{{p_{n}},{o_{n}}} = \hat{y}_{{p_{n-1}},{o_n}}+G_{o_n}(x_{{o_n}},\hat{y}_{{p_{n-1}},{o_n}})          
}
\subsection{Datasets}
\label{datasets_section}
We demonstrated the proposed ProvoGAN approach on a public brain dataset, an in vivo knee dataset, and an in vivo brain dataset. The public dataset, IXI (\url{https://brain-development.org/ixi-dataset/}), consisted of coil-combined magnitude multi-contrast brain MR images of healthy subjects. The in vivo knee dataset \citep{kneedataset} consisted of multi-coil complex knee MR images of healthy subjects. The in vivo brain dataset contained multi-contrast brain MR images of both healthy subjects and glioma patients. Further details about each dataset are provided below.\\~\\
\textbf{IXI Dataset: }\Tone-, \Ttwo-, and proton-density (PD-) weighted brain MR images of $52$ subjects were used, where $37$ subjects were reserved for training, $5$ for validation, and $10$ for testing. \Tone-weighted images were acquired sagittally with repetition time $=9.813$ ms, echo time $=4.603$ ms, flip angle $=8\degree$, spatial resolution $=0.94\times 0.94\times 1.2$ mm$^3$, and matrix size $=256\times256\times150$. \Ttwo-weighted images were acquired axially with repetition time $=8178$ ms, echo time $=100$ ms, flip angle $= 90\degree$, spatial resolution $=0.94\times0.94\times1.20$ mm$^3$, and matrix size $= 256\times256\times150$. PD-weighted images were acquired axially with repetition time $=8178.34$ ms, echo time $=8$ ms, flip angle $=90\degree$, spatial resolution $=0.94\times0.94\times1.2$ mm$^3$, and matrix size $=256\times256\times150$. Since the images of separate contrasts were spatially unregistered in this dataset, \Ttwo- and PD-weighted images were registered onto \Tone-weighted images using FSL \citep{fsl_1,fsl_2} via an affine transformation.  For synthesis images were further registered onto the Montreal Neurological Institute (MNI) template of \Tone-weighted images with an isotropic resolution of 1 mm$^3$.  Registration was performed based on mutual information loss.
\\~\\
\textbf{In vivo Knee Dataset: }PD-weighted multi-coil knee MR images of $20$ subjects were used, where $12$ subjects were reserved for training, $3$ for validation, and $5$ for testing. Images were sagittally acquired with $8$ receive coils, repetition time $=1550$ ms, echo time $=25.661$ ms, spatial resolution $=0.5\times0.5\times0.6$ mm$^3$, and matrix size $=320\times320\times256$. MRI scans were performed in the Richard M. Lucas Center at Stanford University, California, United States on 3T GE scanners.
\\~\\
\textbf{In vivo Brain Dataset: }\Tone-weighted, contrast enhanced \Tone-weighted (\Tonec), \Ttwo-weighted, and FLAIR coil-combined brain MR images of $11$ healthy subjects, $12$ glioma patients with homogenous tumor, and $62$ glioma patients with heterogenous tumor were used. $55$ subjects were reserved for training (healthy: $8$, homogenous: $7$, heterogenous: $40$), $15$ for validation (healthy: $2$, homogenous: $2$, heterogenous: $11$), and $15$ for testing (healthy: $2$, homogenous: $2$, heterogenous: $11$). Data augmentation was performed to prevent class imbalance among the three subject groups. Augmentation was achieved by rotating the volumes around their longitudinal axis by a random angle in the range $[-10\degree, 10\degree]$, and repeated $10$ times for healthy subjects, $9$ times for glioma patients with homogenous tumor, and performed once for glioma patients with heterogenous tumor.  MRI exams were performed in the Department of Radiology at Hacettepe University, Ankara, Turkey, on Siemens and Philips scanners under a diverse set of protocols with varying spatial resolution across both contrast sets and subjects. Specifically, the prescribed resolutions included $1\times1\times1$ mm$^3$, $0.9\times0.9\times1.5$ mm$^3$, $0.9\times0.9\times1.8$ mm$^3$, $0.9\times0.9\times2.2$ mm$^3$ for \Tone- and \Tonec-weighted images, and $0.3\times0.3\times5$ mm$^3$, $0.4\times0.4\times5$ mm$^3$, $0.5\times0.5\times5$ mm$^3$, $0.6\times0.6\times5$ mm$^3$, $0.7\times0.7\times5$ mm$^3$ for \Ttwo-weighted and FLAIR images.  For demonstrations, all images were registered onto the MNI template of \Tone-weighted images with an isotropic resolution of $1\mathrm{\,mm^3}$. Registration was performed via FSL \citep{fsl_1,fsl_2} using affine transformation based on mutual information loss. Imaging protocols were approved by the local ethics committee at Hacettepe University. All participants provided written informed consent.
\par
For MRI reconstruction, volumes in the IXI and in vivo knee datasets were retrospectively undersampled with variable-density sampling patterns for  acceleration factors ($R=4,8,12,16$). A sampling density function across k-space was taken a bi-variate normal distribution with mean at the center of k-space. The variance of the distribution was adjusted to achieve the expected sampling rate given $R$. The in-plane orientation was designated as axial. For MRI synthesis, all brain images were further skull stripped using FSL \citep{fsl_1,fsl_2} with functional intensity threshold of $0.5$, and vertical gradient intensity threshold of $0$.

\subsection{Competing Methods}
\label{comp_met_section}
To demonstrate the performance of ProvoGAN in MR image recovery, we compared it against several state-of-the-art 3D models (vGAN, SC-GAN, REPLICA), 2D models (sGAN, RefineGAN, SPIRiT, SparseMRI), and hybrid models (M$^3$NET, TransferGAN). Baselines implemented for both reconstruction and synthesis included sGAN, vGAN, M$^3$NET, and TransferGAN. Meanwhile, task-specific baselines were RefineGAN, SPIRiT, and SparseMRI in MRI reconstruction, and SC-GAN and REPLICA in MRI synthesis. 

The main effect that we seeked in comparing ProvoGAN against sGAN and vGAN was the benefit of progressive volumetrization over purely 2D or 3D processing. To improve reliability of these comparisons, we wanted to control for potential confounds from secondary factors such as network architecture or loss function. Therefore, the sGAN and vGAN models embodied consistent generator-discriminator architectures and loss functions with ProvoGAN (see Supp. Text 1,2 and Supp. Fig. 1,2 for details).  
\\~\\
\textbf{vGAN: } A learning-based volumetric GAN model that performs a global one-shot mapping between source and target volumes (see Section \ref{vGAN_section}). vGAN was implemented with a ResNet-based generator and a PatchGAN discriminator. 
\\~\\
\textbf{sGAN: }A learning-based cross-sectional GAN model that performs a localized mapping between cross-sections of the source and target volumes (see Section \ref{sGAN_section}). sGAN contained a ResNet-based generator and a PatchGAN discriminator. 
\\~\\ \textbf{RefineGAN: }A learning-based cross-sectional GAN model proposed for MRI reconstruction \citep{Quan2018c}. RefineGAN uses a cycle-consistency loss for acquired k-space samples in addition to adversarial and pixel-wise image loss to improve reconstruction quality. The overall architecture and loss terms were taken from \cite{Quan2018c}, but a ResNet-based generator was implemented to enable fair comparisons against ProvoGAN as it was observed here to yield higher reconstruction quality. 
\\~\\ \textbf{SC-GAN: }A learning-based volumetric GAN model proposed for MRI synthesis \citep{str_artifacts}. SC-GAN leverages self-attention modules to improve capture of long-range spatial dependencies. SC-GAN was implemented with a U-Net based generator and a PatchGAN discriminator as described in \cite{str_artifacts}, where the encoder and decoder components in the generator and the intermediate layer in the discriminator contained a self-attention module.
\\~\\ \textbf{M\textsuperscript{3}NET: }A learning-based hybrid model proposed for MRI segmentation \citep{volume_volume_fuse}. First, M\textsuperscript{3}NET separately learns orthogonal cross-sectional mappings in three rectilinear orientations (i.e., axial, coronal, sagittal). Using these 2D mappings as parallel streams, it fuses their outputs with a 3D fusion module to recover the target volume. The overall architecture, 3D fusion module, and loss functions were adopted from \cite{volume_volume_fuse}, where 2D models were implemented with ResNet-based generators and PatchGAN discriminators as they were observed to yield enhanced performance in this study. 
\\~\\ \textbf{TransferGAN: }A learning-based hybrid GAN model proposed for low-dose CT denoising \citep{shan20183}. TransferGAN pretrains a 2D model for image recovery in a specific orientation, and then performs domain transfer from 2D onto 3D by transferring model weights. The transfer learning procedure was implemented as described in \cite{shan20183}, with 2D-3D models implemented as conditional GANs using ResNet-based generators and PatchGAN discriminators for fair comparison against ProvoGAN.
\\~\\ \textbf{SparseMRI: }A compressed sensing-based cross-sectional method for single-coil MRI reconstruction \citep{lustig2007}. SparseMRI enforces transform domain sparsity as prior information during reconstruction from undersampled acquisitions. Here, SparseMRI was implemented as described in \cite{lustig2007}.
\\~\\ \textbf{SPIRiT: }A compressed sensing-based cross-sectional method for multi-coil MRI reconstruction \citep{Lustig2010}. SPIRiT employs k-space interpolation kernels to estimate missing k-space samples. Here, SPIRiT was implemented as described in \cite{Lustig2010}.
\\~\\ \textbf{REPLICA: }A compressed sensing-based volumetric method for multi-contrast MRI synthesis \citep{Jog2017b}. REPLICA performs a nonlinear intensity transformation in multi-resolution feature space via a regression ensemble based on random forests. Here, REPLICA was implemented as described in \cite{Jog2017b}.
\\~\\
In single-coil reconstruction, learning-based models  were trained to recover a magnitude image given real and imaginary parts of the undersampled image. In multi-coil reconstruction, learning-based models  were first trained to recover a coil-combined magnitude image given real and imaginary parts of coil-combined Fourier reconstructions of undersampled acquisitions. A complex image was then formed by mapping the phase of the coil-combined undersampled image onto the predicted magnitude image. Coil combination was performed using sensitivity maps estimated via ESPIRiT \citep{espirit}. A multi-coil complex image was obtained by projecting the coil-combined network prediction onto individual coils with the estimated sensitivity maps. Data-consistency was enforced in Fourier domain using the multi-coil complex images. In synthesis, learning-based models  were trained to recover the magnitude image of the target contrast given magnitude images of the source contrasts.

The volumetric vGAN, SC-GAN, and REPLICA methods received as input volumetric source images. 
The cross-sectional sGAN-A, sGAN-C, sGAN-S, RefineGAN, SPIRiT, and SparseMRI methods received as input individual cross-sections of source volumes. M$^3$NET received cross-sectional inputs, aggregated them across the volume and finally processed the entire volume. TransferGAN received cross-sectional inputs during pretraining of the 2D model, and instead received volumetric inputs during training of the 3D model. Details regarding the dimensionality of input data to each method are provided in Supp. Text 3.

ProvoGAN, vGAN, and sGAN were implemented in Python 2.7 using PyTorch 0.4 and NumPy 1.14 libraries. Implementations of RefineGAN and SC-GAN were adopted from \cite{Quan2018c} and \cite{str_artifacts} respectively, and performed in Python 3.6 using PyTorch 1.10 and Numpy 1.19 libraries. Implementations of M$^3$NET and TransferGAN were adapted from \cite{volume_volume_fuse} and \cite{shan20183} respectively, and performed in Python 2.7 using the PyTorch 0.4 and Numpy 1.14 libraries. SparseMRI and SPIRiT were implemented in MATLAB using the toolboxes available at \url{https://people.eecs.berkeley.edu/~mlustig/Software.html}. REPLICA was also implemented in MATLAB using the toolboxes shared by \cite{Jog2017b}. All implementations were run on workstations equipped with Intel(R) Core(TM) i7-7800X @ 3.50GHz and i7-6850K @ 3.60GHz CPUs, and nVidia GeForce GTX 1080 Ti and RTX 2080 Ti GPUs. Quantitative performance assessments based on PSNR and SSIM were performed in Python 2.7 using the Scikit-image 0.14 library. Please note that the code and data to build and demonstrate ProvoGAN and competing deep-learning models will be publicly available at \url{http://github.com/icon-lab/mrirecon} upon publication.

\subsection{Experiments}
\label{experiments_section}
\textbf{\\Task-Specific Progression Order in ProvoGAN: }Experiments were performed on ProvoGAN to optimize its progression order across the rectilinear orientations for specific tasks. To do this, multiple independent ProvoGAN models were trained while varying the progression order: 1)~\ACS, 2)~\ASC, 3)~\CAS, 4)~\CSA, 5)~\SAC, 6)~\SCA, where A denotes the axial, C denotes the coronal, and S denotes the sagittal orientation. Performance of these models were evaluated on the validation set via PSNR measurements. The experiments were performed separately for all synthesis and reconstruction tasks, and the progression orders optimized for specific tasks were used in all evaluations thereafter.
\\~\\
\textbf{MRI Reconstruction: }Reconstruction experiments were performed on the IXI and in vivo knee datasets to compare ProvoGAN against sGAN, vGAN, RefineGAN, SparseMRI, and SPIRiT.  In the IXI dataset, the proposed and competing methods were demonstrated separately for single-coil reconstruction of \Tone- and \Ttwo-weighted images with four distinct acceleration factors ($R=4,8,12,16$). Meanwhile, in the in vivo knee dataset, the proposed and competing methods were demonstrated for multi-coil reconstruction of PD-weighted images again with ($R=4,8,12,16$). Note that a single sGAN model was trained in the axial orientation (sGAN-A) given the axial readout direction. 
\\~\\

\textbf{MRI Synthesis: }Synthesis experiments were performed on the IXI and in vivo brain datasets to demonstrate ProvoGAN against sGAN, vGAN, SC-GAN, and REPLICA. All synthesis experiments were conducted on coil-combined magnitude images. In the IXI dataset, three synthesis tasks were considered: 1)~\TtwoPDTone, 2)~\TonePDTtwo, 3)~\ToneTtwoPD. In the in vivo brain dataset, four synthesis tasks were considered: 1)~\TtwoFlairTonecTone, 2)~\ToneFlairTonecTtwo, 3)~\ToneTtwoTonecFlair, 4)~\ToneTtwoFlairTonec. For each task, three independent sGAN models were implemented to recover target cross-sections in separate orientations: sGAN-A for the axial, sGAN-C for the coronal, sGAN-S for the sagittal orientation.
\\~\\

\textbf{Progressive Volumetrization versus Hybrid Models: }Experiments were conducted on the IXI dataset to demonstrate ProvoGAN against M\textsuperscript{3}NET and TransferGAN. Reconstruction experiments were conducted for \Tone- and \Ttwo-weighted image recovery tasks at four distinct acceleration factors ($R=4,8,12,16$). Meanwhile, synthesis experiments were conducted for the many-to-one recovery tasks of \TtwoPDTone, \TonePDTtwo, and \ToneTtwoPD.
\\~\\

    \textbf{Radiological Evaluation: }To assess the clinical value of the recovered images, an expert radiologist ($25+$ years of experience) gave opinion scores to the images while blinded to the method name and order of presentation. Reconstructed images were evaluated for single-coil reconstructions  of \Tone- and \Ttwo-weighted acquisitions at $R=8$ in the IXI dataset, and multi-coil reconstructions of PD-weighted acquisitions at $R=8$ in the in vivo knee dataset. Synthesized images were evaluated for \TtwoPDTone$~$ in IXI and \ToneTtwoTonecFlair$~$ in the in vivo brain datasets. From each recovered volume, intermediate axial, coronal, and sagittal cross-sections were randomly selected, and the image quality was rated as the similarity to ground truth images on a five-point scale (5: perfect match, 4: good, 3: moderate, 2: limited, 1: very poor, 0: unacceptable). 
\\~\\
\textbf{Multi-Cross-Section Models: }To demonstrate the benefit of leveraging contextual priors by incorporating multiple neighboring cross-sections at the input level, variants of ProvoGAN and sGAN, referred to as ProvoGAN(multi) and sGAN(multi), were implemented, which receive as input $n_c$ consecutive cross-sections to recover the corresponding central cross-section in the target volume. Here $n_c=3$ was selected as higher number of cross-sections did not yield a notable benefit in recovery performance \citep{Dar2019}. Experiments were performed on the IXI dataset for reconstruction of \Tone- and \Ttwo-weighted images with distinct acceleration factors ($R=4,8,12,16$), and for many-to-one synthesis tasks (\TtwoPDTone, \TonePDTtwo, \ToneTtwoPD). The ordering of the progressions across the orientations in ProvoGAN was optimized via PSNR measurements in the validation set. Three separate sGAN(multi) models were implemented in each individual rectilinear orientation: sGAN(multi)-A for the axial, sGAN(multi)-C for the coronal, and sGAN(multi)-S for the sagittal orientation.
\\~\\
\textbf{Cross-Sectional Models of Varying Complexity: }An additional analysis was performed on ProvoGAN and sGAN to examine recovery performance as a function of the complexity of convolutional layers. Several variants of ProvoGAN and sGAN were implemented while the number of network weights in individual convolutional layers were scaled by $n_f\in\{1/16,1/9,1/4,1,4,9,16\}$, where the kernel size, number of layers, number of hidden units were kept fixed but the number of filters were modified. This resulted in seven distinct ProvoGAN and sGAN pairs: ProvoGAN($n_f$)-sGAN($n_f$). Experiments were performed on the IXI dataset for single-coil reconstruction of \Tone-weighted  acquisitions undersampled at $R=8$ and a many-to-one synthesis task of \TtwoPDTone. The ordering of the progressions across the orientations in ProvoGAN($n_f$) was optimized via PSNR measurements in the validation set. Separate sGAN models were trained in the individual rectilinear orientations (axial, coronal, and sagittal) for each model complexity level: sGAN($n_f$)-A, sGAN($n_f$)-C, sGAN($n_f$)-S.
\\~\\ 
\textbf{Data Efficiency: } The complexity of volumetric models can elevate the amount of data samples required for successful training. Instead, ProvoGAN comprises sequential cross-sectional models that are of lower complexity and that can be trained effectively with fewer data. Experiments were conducted to comparatively demonstrate the data efficiency of ProvoGAN against vGAN. To do this, models were trained using data from varying number of subjects ($n_T\in\{5, 15, 25\}$), yielding ProvoGAN($n_T$) and vGAN($n_T$). \Tone- and \Ttwo-weighted reconstructions in the IXI dataset at $R=4,8,12,16$ were considered. \TtwoPDTone, \TonePDTtwo, and \ToneTtwoPD~ synthesis tasks were considered. To prevent potential confounds, the optimal progression orders determined for the original ProvoGAN models were retained. 
\\~\\ 
\textbf{Generalizability of Progressive Volumetrization: }Experiments were conducted to demonstrate the generalizability of the proposed progressive volumetrization to another network architecture. Demonstrations were performed on the IXI dataset for \ToneTtwoPD, \TonePDTtwo, and \TtwoPDTone~synthesis tasks. A recent state-of-the-architecture, SC-GAN, with a U-Net backbone using intermittent self-attention layers \citep{str_artifacts} was considered. Variants of sGAN, vGAN, and ProvoGAN were implemented based on this architecture: sSC-GAN, vSC-GAN, ProvoSC-GAN. Again, three separate sSC-GAN models were trained in each orientation: sSC-GAN-A in the axial, sSC-GAN-C in the coronal, and sSC-GAN-S in the sagittal orientation.
\\~\\
\textbf{Statistical Assessments: }PSNR, SSIM, and opinion scores were utilized to quantitatively evaluate the recovery quality of the methods under comparison. Since the performance measurements from these metrics followed a non-normal distribution ($p<0.05$ with Shapiro-Willks test), significance of differences in quantitative metrics were evaluated using non-parametric statistical tests. Assessments of progression order in ProvoGAN were performed via Kruskal-Wallis tests, whereas performance comparisons among competing methods were performed via Wilcoxon signed-rank tests.

\begin{figure*}[t]
\centering
\includegraphics[width=0.80\linewidth]{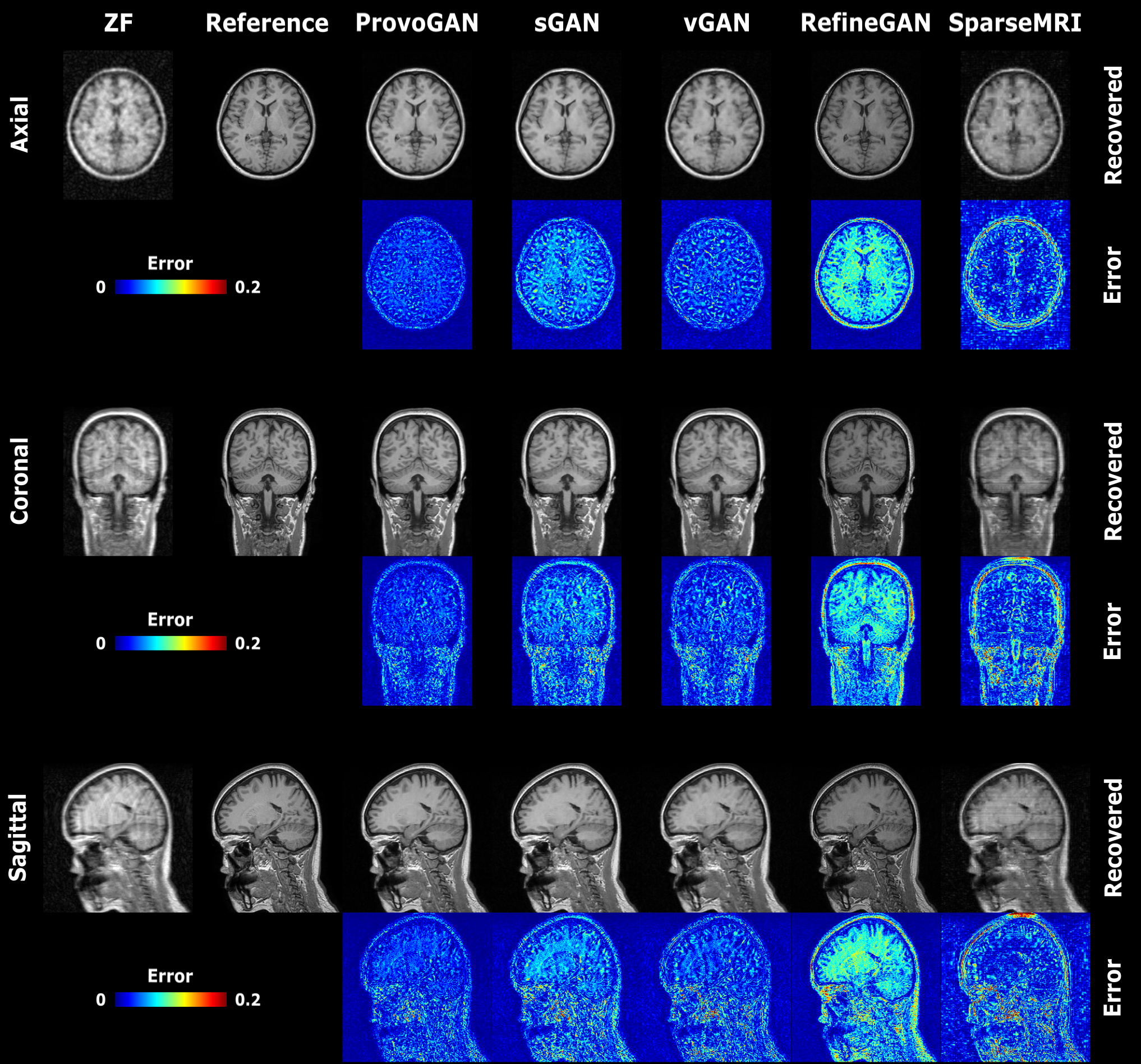}
\caption{The proposed ProvoGAN method is demonstrated on the IXI dataset for single-coil reconstruction of \Tone-weighted acquisitions undersampled at $R=8$. Representative results are displayed for all competing methods together with the zero-filled (ZF) undersampled source images (first column) and the reference target images (second column). The top two rows display results for the axial, the middle two rows for the coronal, and the last two rows for the sagittal orientation. Error was taken as the absolute difference between the reconstructed and reference images (see colorbar). Overall, the proposed ProvoGAN method offers delineation of tissues with higher acuity compared to the volumetric (vGAN) model, and alleviates undesirable discontinuities compared to cross-sectional models (sGAN, RefineGAN, SparseMRI) by improving reconstruction performance in all orientations. }
\label{fig:IXI_T1_recon_8x}
\end{figure*}

\section{Results}

\subsection{Task-Specific Progression Order}
ProvoGAN serially decomposes a given volumetric recovery task into cross-sectional mappings in three rectilinear orientations. Subsequent 2D mappings are residually learned based on outputs from earlier progressions. Please note that spatial distribution of the tissues and the correlations between the source-target images may vary uniquely across orientations for each recovery task. In this setup, if an earlier 2D model yields relatively higher artifacts, the task difficulty for the remaining progressions would be elevated. Contrarily, initiating the progression at a different orientation with lower artifacts can reduce task difficulty for the remaining stages.   Therefore, we predicted that the progression sequence in ProvoGAN can significantly affect task-specific recovery performance. 

To test this prediction, we performed reconstruction and synthesis experiments separately on the IXI, in vivo brain, and in vivo knee datasets (see Section \ref{datasets_section} for details). We comparatively evaluated performance of multiple independent ProvoGAN models for the six possible permutations of the progression sequence: 1)~\ACS, 2)~\ASC, 3)~\CAS, 4)~\CSA, 5)~\SAC, 6)~\SCA, where A denotes the axial, C denotes the coronal, and S denotes the sagittal orientation. Here, we considered volumetric PSNR measurements between the recovered and reference target volumes within the validation set. The highest and lowest performing ProvoGAN models yield an average PSNR difference of $3.44$ dB for single-coil reconstruction tasks in IXI and $3.42$ dB for multi-coil reconstruction tasks in the in vivo knee dataset (see Supp. Tables 1,2 for details). Meanwhile, the average PSNR difference between the highest and lowest performing ProvoGAN models is $1.46$ dB for synthesis in IXI, and $1.01$ dB for synthesis in the in vivo brain dataset (see Supp. Tables 3,4).  Optimization of the progression order enables a significant performance increase in both reconstruction ($p<0.05$, Kruskal-Wallis test) and synthesis ($p<0.05$). Therefore, the optimal orders were utilized for each recovery task in all evaluations thereafter unless otherwise stated. 

 Note that brain and knee MRI acquisitions were undersampled across the two phase-encoding dimensions in the axial plane (A/P, L/R) in the reconstruction experiments, so the reconstruction task in the axial plane is relatively more difficult. Accordingly, there is a general performance increase in progression orders that leave the axial orientation towards later stages of ProvoGAN, and this effect is particularly emphasized towards higher acceleration rates $R$ (see Supp. Tables 1,2). For synthesis, a factor that contributes to task difficulty is the level of structural details in the target contrast. Images in the IXI dataset and \Ttwo-weighted and FLAIR images in the in vivo brain dataset have relatively higher spatial resolution in the axial plane, but broader voxel dimensions in the longitudinal direction. Accordingly, a general performance increase is observed in progression orders that leave the axial orientation towards later stages, and these effects are more accentuated when the target is \Tone- and \Ttwo-weighted images that have relatively better capture of structural details compared to other contrasts such as PD- or \Tonec-weighted (see Supp. Tables 3,4). 

\renewcommand{\arraystretch}{1.15}
\begin{table*}[!t]
\centering
\caption{Quality of Reconstruction in the IXI Dataset: Volumetric PSNR (dB) and SSIM (\%) measurements between the reconstructed and ground truth images in the test set in the IXI dataset are given as mean $\pm$ std for the test set. The measurements are reported for zero-filled images (ZF), the proposed ProvoGAN and competing sGAN, vGAN, RefineGAN, and SparseMRI reconstruction methods for four distinct acceleration factors ($R=4,8,12,16$). Boldface indicates the best performing method.}
\fontsize{8}{10}\selectfont
\begin{tabular}{cccccccccccccc}
\cline{3-14}
& & \multicolumn{2}{c}{ProvoGAN} & \multicolumn{2}{c}{sGAN}  & \multicolumn{2}{c}{vGAN} & \multicolumn{2}{c}{RefineGAN} & \multicolumn{2}{c}{SparseMRI} & \multicolumn{2}{c}{ZF} \\ \cline{3-14}\vspace{-1mm}
& & PSNR & SSIM & PSNR & SSIM & PSNR & SSIM & PSNR & SSIM & PSNR & SSIM & PSNR & SSIM   \\ \hline
\multirow{4}{*}{R=4} & \multirow{2}{*}{\Tone} & \textbf{35.25} & \textbf{96.73} & 33.85 & 93.21 & 30.82 & 88.61 & 31.04 & 92.13 & 26.25 & 72.16 & 24.59 & 64.96  \\ 
& & \textbf{$\pm$ 1.78} & \textbf{$\pm$ 0.57} & $\pm$ 1.29 & $\pm$ 0.78 & $\pm$ 1.24 & $\pm$ 0.93 & $\pm$ 1.74 & $\pm$ 0.94 & $\pm$ 0.80 & $\pm$ 1.92 & $\pm$ 1.17 & $\pm$ 3.07 \\ \cline{2-14}
& \multirow{2}{*}{\Ttwo} & \textbf{35.50} & \textbf{96.08} & 32.95 & 86.44 & 33.09 & 93.13 & 33.96 & 94.00 & 28.08 & 82.44 & 27.54 & 75.14 \\ 
& & \textbf{$\pm$ 2.62} & \textbf{$\pm$ 1.07} & $\pm$ 1.50 & $\pm$ 1.26 & $\pm$ 1.73 & $\pm$ 1.20 & $\pm$ 0.91 & $\pm$ 0.57 & $\pm$ 0.73 & $\pm$ 1.42 & $\pm$ 1.04 & $\pm$ 2.09 \\ \hline
\multirow{4}{*}{R=8} & \multirow{2}{*}{\Tone}  & \textbf{31.38} & \textbf{94.93} & 30.08 & 91.18 & 29.71 & 88.73 & 27.63 & 91.81 & 25.92 & 72.26 & 23.04 & 62.53 \\
& & \textbf{$\pm$ 1.26} & \textbf{$\pm$ 0.86} & $\pm$ 1.32 & $\pm$ 1.01 & $\pm$ 0.88 & $\pm$ 1.23 & $\pm$ 1.56 & $\pm$ 0.68 & $\pm$ 0.36 & $\pm$ 1.52 & $\pm$ 1.05 & $\pm$ 3.32 \\ \cline{2-14}
& \multirow{2}{*}{\Ttwo} & \textbf{33.49} & \textbf{95.92} & 32.24 & 90.47 & 31.35 & 92.57 & 30.70 & 93.82 & 26.55 & 79.35 & 26.67 & 74.16 \\ 
& & \textbf{$\pm$ 2.21} & \textbf{$\pm$ 1.01} & $\pm$ 2.14 & $\pm$ 0.95 & $\pm$ 0.66 & $\pm$ 0.86 & $\pm$ 1.51 & $\pm$ 0.40 & $\pm$ 0.51 & $\pm$ 1.39 & $\pm$ 0.87 & $\pm$ 2.10 \\ \hline
\multirow{4}{*}{R=12} & \multirow{2}{*}{\Tone} & \textbf{29.67} & \textbf{92.48} & 27.34 & 86.23 & 27.56 & 82.70 & 27.48 & 88.56 & 23.84 & 60.77 & 20.76 & 52.72 \\ 
& & \textbf{$\pm$ 0.91} & \textbf{$\pm$ 0.90} & $\pm$ 1.06 & $\pm$ 1.36 & $\pm$ 1.20 & $\pm$ 1.70 & $\pm$ 1.08 & $\pm$ 1.07 & $\pm$ 0.28 & $\pm$ 1.74 & $\pm$ 1.04 & $\pm$ 3.62 \\ \cline{2-14}
& \multirow{2}{*}{\Ttwo} & \textbf{30.41} & \textbf{91.98} & 28.48 & 79.50 & 27.97 & 85.76 & 29.16 & 91.49 & 24.67 & 70.16 & 24.60 & 66.90 \\ 
& & $\pm$ \textbf{1.03} & \textbf{$\pm$ 1.40} & $\pm$ 1.06 & $\pm$ 2.15 & $\pm$ 0.79 & $\pm$ 1.74 &  $\pm$ 1.54 &  $\pm$ 0.87 &  $\pm$ 0.59 &  $\pm$ 1.67 & $\pm$ 0.70 & $\pm$ 2.12 \\ \hline
\multirow{4}{*}{R=16} & \multirow{2}{*}{\Tone} & \textbf{29.15} & \textbf{91.40} & 26.73 & 85.23 & 25.47 & 79.56 & 22.37 & 84.31 & 23.42 & 61.64 & 20.95 & 52.85 \\ 
& & \textbf{$\pm$ 1.09} & \textbf{$\pm$ 1.09} & $\pm$ 1.53 & $\pm$ 1.74 & $\pm$ 1.07 & $\pm$ 2.16 &  $\pm$ 1.07 &  $\pm$ 1.36 &  $\pm$ 0.47 &  $\pm$ 1.86 & $\pm$ 1.07 & $\pm$ 3.46 \\ \cline{2-14}
& \multirow{2}{*}{\Ttwo} & \textbf{30.66} & \textbf{93.74} & 29.05 & 83.38 & 27.66 & 85.47 & 28.12 & 91.07 & 24.37 & 69.15 & 24.44 & 66.70\\ 
& & \textbf{$\pm$ 1.60} & \textbf{$\pm$ 1.35} & $\pm$ 1.04 & $\pm$ 1.22 & $\pm$ 0.50 & $\pm$ 1.57 & $\pm$ 1.26 & $\pm$ 0.84 & $\pm$ 0.69 & $\pm$ 1.60 & $\pm$ 0.76 & $\pm$ 2.19 \\ \hline
\end{tabular}
\label{tab: IXI_recon_compare}
\end{table*}

\begin{figure*}[t]
\centering
\includegraphics[width=0.8\linewidth]{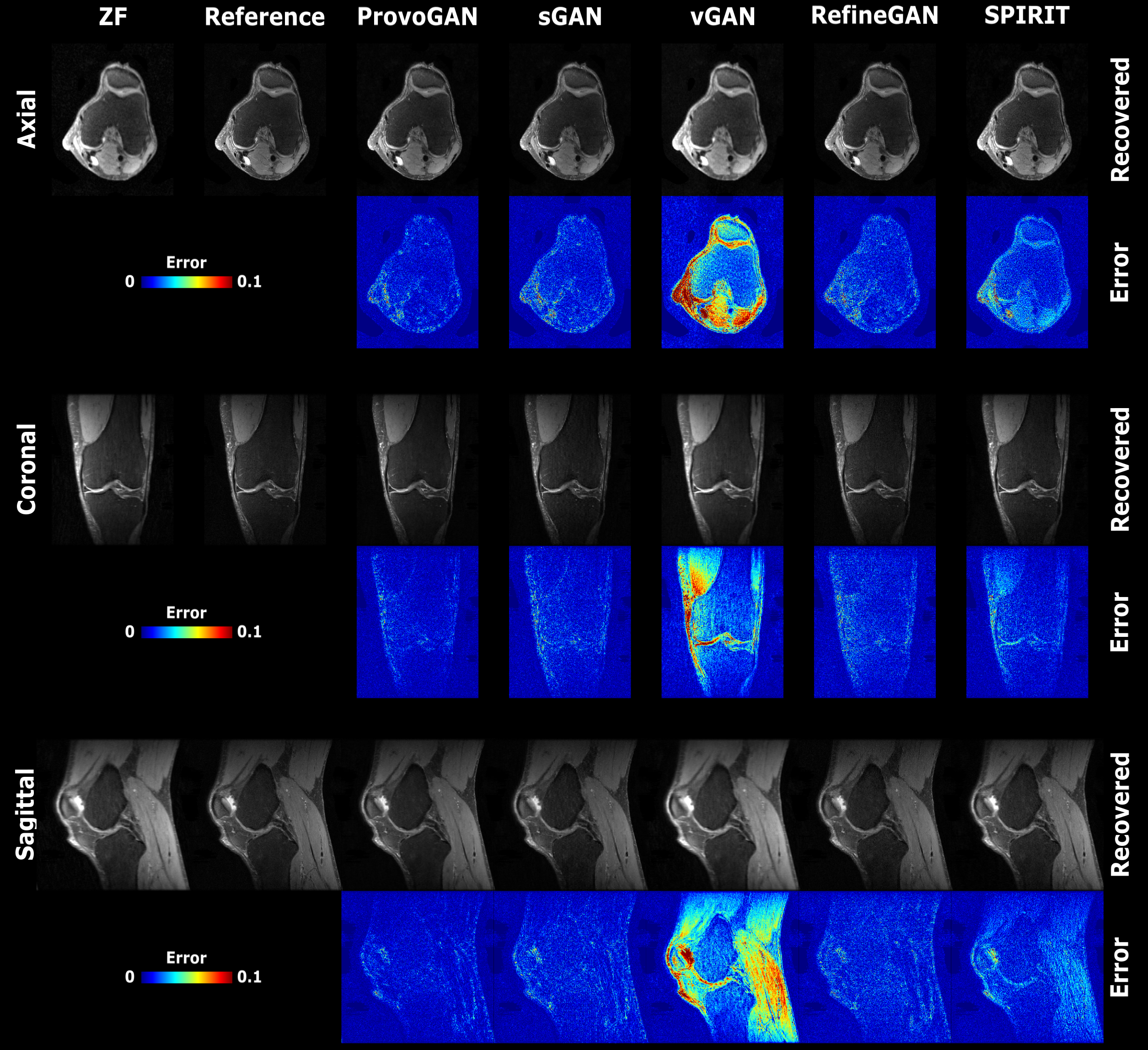}
\caption{ The proposed method is demonstrated on the in vivo multi-coil knee dataset for reconstruction at an acceleration ratio of $R=8$. Representative results are displayed for all competing methods together with the zero-filled (ZF) undersampled source images (first column) and the reference target images (second column). The top two rows display results for the axial, the middle two rows for the coronal, and the last two rows for the sagittal orientation. Error was taken as the absolute difference between the reconstructed and reference images (see colorbar). Overall, ProvoGAN achieves sharper tissue depiction compared to vGAN, and alleviates undesirable discontinuities compared to cross-sectional models (sGAN, RefineGAN, SPIRiT) by improving reconstruction performance in all orientations. }
\label{fig:knee_recon_8x}
\end{figure*}	

\begin{table*}[!t]
\centering
\caption{Quality of Reconstruction in the In vivo Knee Dataset: Volumetric PSNR (dB) and SSIM (\%) measurements between the reconstructed and ground truth images in the test set in the in vivo knee dataset are given as mean $\pm$ std for the test set. The measurements are reported for zero-filled images (ZF), the proposed ProvoGAN and competing sGAN, vGAN, RefineGAN, and SPIRiT methods for four distinct acceleration factors ($R=4,8,12,16$). Boldface indicates the best performing method.}
\fontsize{8}{10}\selectfont
\begin{tabular}{ccccccccccccc}
\cline{2-13}
& \multicolumn{2}{c}{ProvoGAN} & \multicolumn{2}{c}{sGAN}  & \multicolumn{2}{c}{vGAN} & \multicolumn{2}{c}{RefineGAN} & \multicolumn{2}{c}{SPIRiT} & \multicolumn{2}{c}{ZF} \\ \cline{2-13}\vspace{-1mm}
& PSNR & SSIM & PSNR & SSIM & PSNR & SSIM & PSNR & SSIM & PSNR & SSIM & PSNR & SSIM \\ \hline
\multirow{2}{*}{R=4} & \textbf{40.75} & \textbf{95.74} & 40.34 & 95.69 & 36.80 & 92.79 & 40.31 & 95.21 & 39.46 & 95.35 & 32.17 & 93.50 \\ 
& \textbf{$\pm$ 1.35} & \textbf{$\pm$ 0.94} & $\pm$ 1.43 & $\pm$ 0.87 & $\pm$ 1.69 & $\pm$ 1.36 & $\pm$ 1.50 & $\pm$ 1.15 & $\pm$ 1.39 & $\pm$ 1.12 & $\pm$ 2.38 & $\pm$ 1.85\\ \hline
\multirow{2}{*}{R=8} & \textbf{39.45} & \textbf{95.13} & 38.73 & 93.73 & 30.83 & 87.54 & 38.22 & 92.43 & 35.61 & 93.16 & 29.85 & 90.81 \\ 
& \textbf{$\pm$ 2.15} & \textbf{$\pm$ 1.08} & $\pm$ 1.01 & $\pm$ 0.98 & $\pm$ 1.44 & $\pm$ 2.80 & $\pm$ 1.43 & $\pm$ 1.58 & $\pm$ 2.70 & $\pm$ 1.60 & $\pm$ 1.77 & $\pm$ 1.88 \\ \hline
\multirow{2}{*}{R=12}  & \textbf{36.99} & \textbf{93.63} & 36.76 & 91.99 & 29.03 & 88.10 & 36.21 & 90.82 & 33.58 & 91.66 & 28.65 & 89.17 \\ 
& \textbf{$\pm$ 1.29} & \textbf{$\pm$ 0.95} & $\pm$ 1.18 & $\pm$ 0.97 & $\pm$ 1.49 & $\pm$ 1.78 & $\pm$ 0.88 & $\pm$ 1.47 & $\pm$ 3.17 & $\pm$ 1.86 & $\pm$ 1.25 & $\pm$ 1.64\\ \hline
\multirow{2}{*}{R=16} & \textbf{37.86} & \textbf{92.27} & 35.11 & 89.34 & 28.99 & 86.00 & 35.42 & 89.70 & 32.28 & 90.69 & 27.93 & 87.94 \\
& \textbf{$\pm$ 0.50} & \textbf{$\pm$ 1.45} & $\pm$ 1.06 & $\pm$ 1.61 & $\pm$ 2.21 & $\pm$ 2.28 & $\pm$ 0.55 & $\pm$ 1.13 & $\pm$ 2.63 & $\pm$ 1.85 & $\pm$ 1.25 & $\pm$ 1.60 \\ \hline 
\end{tabular}
\label{tab: multicoil_recon_compare}
\end{table*}


\subsection{Accelerated MRI Reconstruction}Next, we performed comprehensive experiments on the IXI and in vivo knee datasets for accelerated MRI reconstruction. We comparatively demonstrated the recovery quality of ProvoGAN against state-of-the-art cross-sectional (sGAN, RefineGAN, SparseMRI, and SPIRiT), and volumetric (vGAN) models (see Section \ref{comp_met_section} for details). We first assessed the performance of the competing methods quantitatively based on volumetric PSNR and SSIM measurements between the reconstructed and high-quality reference images in the test set. We considered single-coil reconstruction tasks in the IXI dataset for \Tone- and \Ttwo-weighted images with distinct acceleration factors  ($R=4,8,12,16$). The proposed ProvoGAN model offers enhanced recovery performance compared to competing methods ($p<0.05$), where it achieves in the range of $[1.85,6.55]$ dB higher PSNR and $[3.26,23.17]$ \% higher SSIM (see Table \ref{tab: IXI_recon_compare}). We then considered multi-coil reconstruction of PD-weighted images in the in vivo brain dataset with $R=4,8,12,16$. ProvoGAN again maintains superior performance to the competing methods ($p<0.05$), where it achieves in the range of $[1.03,7.35$] dB higher PSNR and [$1.48,5.59$] \% higher SSIM (see Table \ref{tab: multicoil_recon_compare}).

To corroborate quantitative assessments, we visually examined the reconstructed volumes from individual methods to identify the nature of reconstruction errors ProvoGAN alleviates. Representative results from the competing methods are shown in Fig. \ref{fig:IXI_T1_recon_8x} for IXI and in Fig. \ref{fig:knee_recon_8x} for the in vivo knee dataset. Overall, cross-sectional models (sGAN, RefineGAN, SparseMRI, SPIRiT) that perform 2D mapping via compressed sensing or deep learning suffer from discontinuity artifacts across individually recovered cross-sections and retrograded capture of fine-structural details. Meanwhile, the volumetric vGAN model performing 3D mapping suffers from loss of spatial resolution within the reconstructed volumes due to noticeable over-smoothing. In contrast, ProvoGAN reconstructs the target volumes with higher consistency across the cross-sections in all orientations and offers sharper delineation of brain and knee tissues. Taken together, these findings clearly outline ProvoGAN's potential to mitigate the limitations of volumetric and cross-sectional models for accelerated MRI reconstruction.

\subsection{Multi-Contrast MRI Synthesis}
We further conducted experiments on the IXI and in vivo brain datasets for multi-contrast MRI synthesis to demonstrate ProvoGAN against state-of-the-art cross-sectional (sGAN) and volumetric (vGAN, SC-GAN, REPLICA) models (see Section \ref{comp_met_section} for details). We again measured volumetric PSNR and SSIM between the synthesized and reference target images for quantitative performance evaluation. In the IXI dataset, we considered synthesis tasks of \TtwoPDTone, \TonePDTtwo, and \ToneTtwoPD. ProvoGAN outperforms the competing methods in all tasks ($p<0.05$), where it achieves in the range of [$1.20,2.90$] dB higher PSNR and [$2.08,4.37$] \% higher SSIM (see Table \ref{tab: IXI_many2one_compare}). In the in vivo brain dataset, we considered synthesis tasks of \TtwoFlairTonecTone, \ToneFlairTonecTtwo, \ToneTtwoTonecFlair, and \ToneTtwoFlairTonec. ProvoGAN again yields enhanced recovery performance in all tasks compared to the competing methods ($p<0.05$), where it maintains [$0.59,5.01$] dB higher PSNR and [$1.96,5.36$] \% higher SSIM (see Table \ref{tab: hacet_many2one_compare}). Note that the in vivo brain dataset was acquired under a diverse set of scanning protocols with varying spatial resolution where 
\Ttwo-weighted and FLAIR images were acquired with larger slice thickness (see Section \ref{datasets_section}). Here, models were built to synthesize \Tone-weighted and \Tonec-weighted images, where the thick-slices data were on the input side.  Models were also built to synthesize \Ttwo-weighted and FLAIR images where thick-slice data were on the output side (see Section \ref{experiments_section}). In both cases, ProvoGAN enhances recovery performance compared to sGAN. On average, ProvoGAN achieves 1.69 dB higher PSNR and 4.18 \% higher SSIM when thick-slice data are on the input side, and 0.70 dB higher PSNR and 1.85 \% higher SSIM when thick-slice data are on the output side.

\begin{figure*}[t]
\centering
\includegraphics[width=0.8\linewidth]{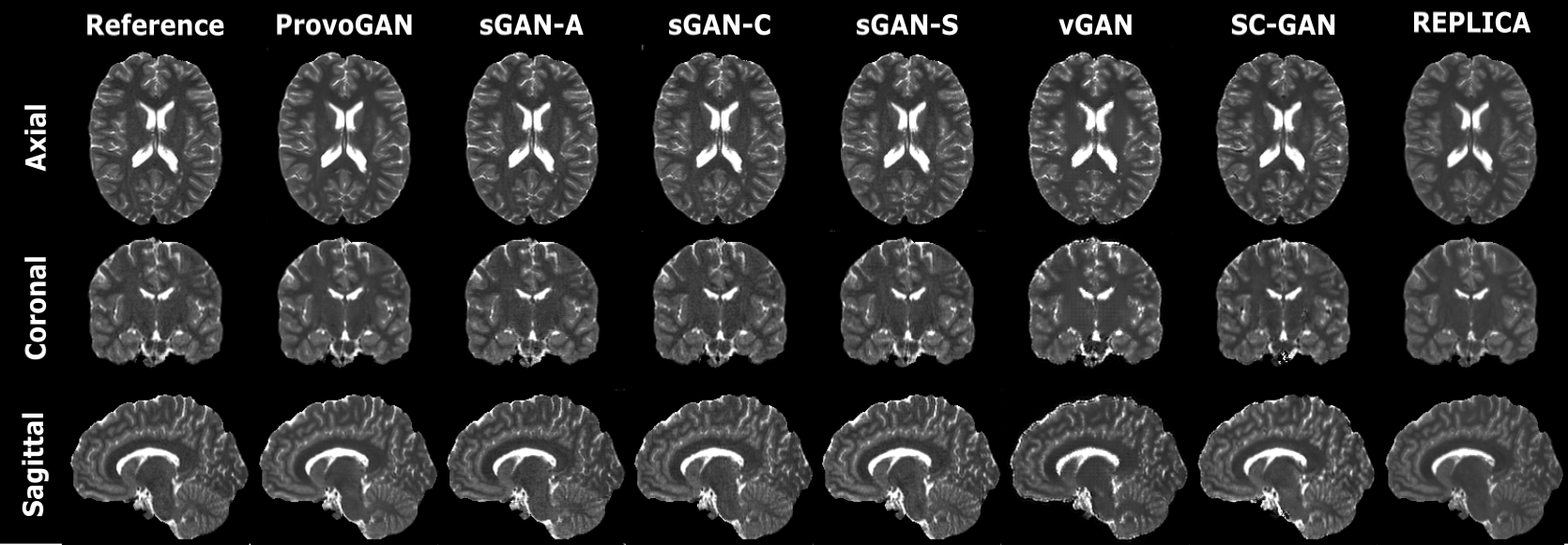}
\caption{The proposed method is demonstrated on the IXI dataset for \Ttwo-weighted image synthesis from \Tone- and PD-weighted images. Representative results are displayed for all competing methods together with the reference target images (first column). The first row displays results for the axial orientation, the second row for the coronal orientation, and the third row for the sagittal orientation. Overall, the proposed method delineates tissues with higher spatial resolution compared to volumetric vGAN, SC-GAN, and REPLICA models, and alleviates discontinuity artifacts by improving synthesis performance in all orientations compared to cross-sectional sGAN-A, sGAN-C, and sGAN-S models.}
\label{fig:T2syn_IXI_many2one}
\end{figure*}

\begin{figure*}[t]
\centering
\includegraphics[width=0.8\linewidth]{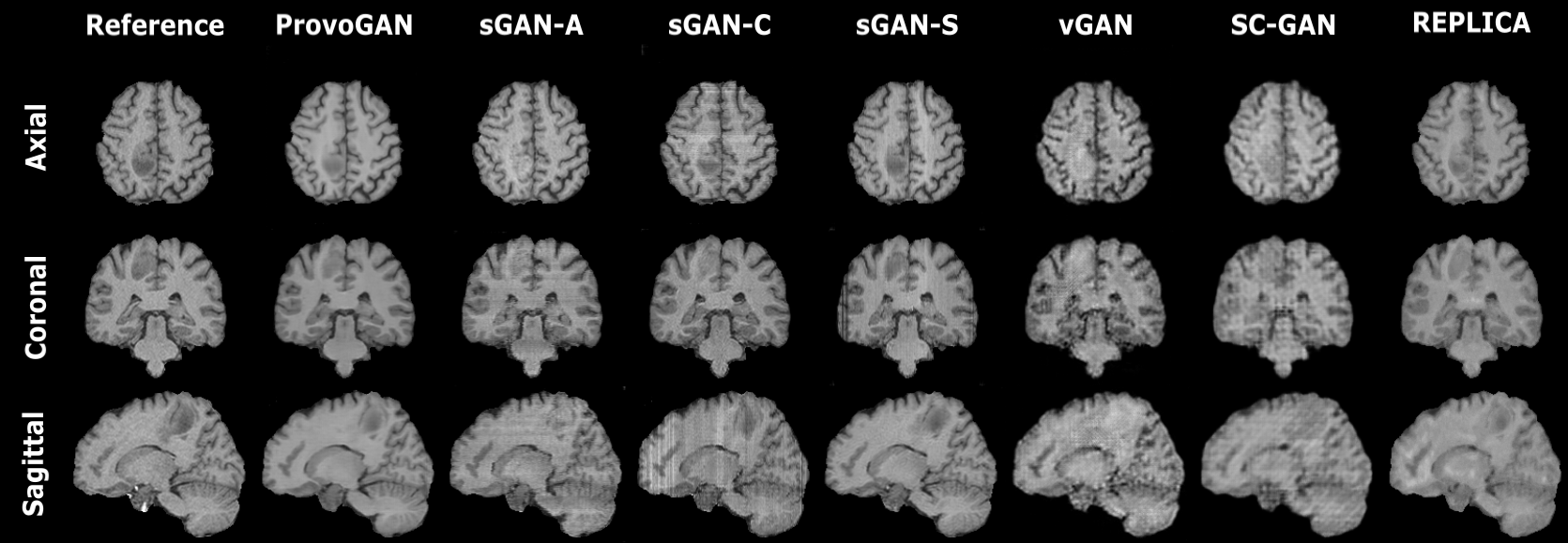}
\caption{The proposed method is demonstrated on the in vivo brain dataset for \Tone-weighted image synthesis from \Ttwo-, \Tonec-weighted and FLAIR images. Representative results are displayed for all competing methods together with the reference target images (first column). The first row displays results for the axial, the second row for the coronal, and the third row for the sagittal orientation. Overall, the proposed method delineates tissues with higher spatial resolution compared to volumetric vGAN, SC-GAN, and REPLICA models, and alleviates discontinuity artifacts by improving synthesis performance in all orientations compared to cross-sectional sGAN-A, sGAN-C, and sGAN-S models. Meanwhile, the proposed method achieves more accurate depictions for tumor regions, which are suboptimally recovered by the competing methods.}
\label{fig:T1syn_hacet}
\end{figure*}

The superior synthesis quality offered by ProvoGAN is clearly visible in representative results displayed in Fig. \ref{fig:T2syn_IXI_many2one} for the IXI dataset and Fig. \ref{fig:T1syn_hacet} for the in vivo brain dataset. These results indicate that the cross-sectional sGAN-A, sGAN-C, and sGAN-S models suffer from suboptimal recovery in the longitudinal dimension due to independent synthesis of cross-sections. Meanwhile volumetric vGAN, SC-GAN, and REPLICA models suffer from poor recovery of fine-structural details and loss of spatial resolution in the target images due to increased model complexity. In comparison to cross-sectional baselines, ProvoGAN alleviates discontinuity artifacts by pooling global contextual information via progressive execution of cross-sectional mappings. In comparison to volumetric baselines, ProvoGAN offers sharper and improved tissue depiction particularly near tumor regions due to its improved learning behavior. Overall, these findings demonstrate ProvoGAN's utility for diverse synthesis tasks in multi-contrast MRI exams.

\renewcommand{\tabcolsep}{3pt}
\begin{table*}[!t]
\centering
\caption{Quality of Synthesis in the IXI Dataset: Volumetric PSNR (dB) and SSIM (\%) measurements between the synthesized and ground truth images in the test set in the IXI dataset are given as mean $\pm$ std. The measurements are provided for the proposed and competing methods for all synthesis tasks: 1)~\TtwoPDTone, 2)~\TonePDTtwo, 3)~\ToneTtwoPD. sGAN-A denotes the sGAN model trained in the axial orientation, sGAN-C in the coronal orientation, and sGAN-S in the sagittal orientation. Boldface indicates the highest performing method.}
\fontsize{8}{10}\selectfont
\begin{tabular}{ccccccccccccccc}
\cline{2-15}
      & \multicolumn{2}{c}{ProvoGAN} & \multicolumn{2}{c}{sGAN-A} & \multicolumn{2}{c}{sGAN-C} & \multicolumn{2}{c}{sGAN-S} & \multicolumn{2}{c}{vGAN} & \multicolumn{2}{c}{SC-GAN} & \multicolumn{2}{c}{REPLICA} \\ \cline{2-15}
& PSNR & SSIM & PSNR & SSIM & PSNR & SSIM & PSNR & SSIM & PSNR & SSIM & PSNR & SSIM & PSNR & SSIM \\\hline
\multirow{2}{*}{\TtwoPDTone} & \textbf{24.15} & \textbf{90.33} & 23.20 & 85.81 & 22.58 & 86.60 & 23.65 & 87.71 & 23.35 & 85.48 & 22.58 & 85.32 & 21.14 & 86.30  \\
& \textbf{$\pm$ 2.80} & \textbf{$\pm$ 4.47} &  $\pm$ 2.08 & $\pm$ 3.95 & $\pm$ 2.11 & $\pm$4.05 & $\pm$1.98 & $\pm$4.15 & 2.89 & 4.18 & 2.99 & 4.00 & $\pm$4.15 & $\pm$ 3.80 \\ \hline
\multirow{2}{*}{\TonePDTtwo} & \textbf{28.97} & \textbf{94.17} & 27.64 & 92.49 & 27.74 & 92.67 & 27.93 & 93.28 & 25.97 & 90.61 & 25.29 & 89.81 & 26.98 & 92.51 \\
& \textbf{$\pm$ 2.91} & \textbf{$\pm$ 4.16} & $\pm$ 2.59 & $\pm$ 4.20 & $\pm$ 2.67 & $\pm$ 4.31 & $\pm$ 2.19 & $\pm$ 2.88 & $\pm$ 1.81 & $\pm$ 4.04 & $\pm$ 1.95 & $\pm$ 4.31 & $\pm$ 2.37 & $\pm$ 4.57  \\ \hline
\multirow{2}{*}{\ToneTtwoPD} & \textbf{29.81} & \textbf{95.41} & 27.69 & 93.64 & 29.00 & 94.21 & 27.12 & 92.67 & 26.17 & 90.70 & 26.36 & 92.12 & 26.96 & 94.10 \\
& \textbf{$\pm$ 2.96} & \textbf{$\pm$ 2.75} & 2.20 & $\pm$ 3.00 & $\pm$ 2.41 & $\pm$ 2.99 & $\pm$ 1.61 & $\pm$ 2.95 & $\pm$ 1.41 & $\pm$ 2.47 & $\pm$ 1.41 & $\pm$ 2.59 & $\pm$ 2.93 & $\pm$ 3.03 \\ \hline
\end{tabular}
\label{tab: IXI_many2one_compare}
\end{table*}

\renewcommand{\tabcolsep}{3pt}
\begin{table*}[!t]
\centering
\caption{Quality of Synthesis in the In vivo Brain Dataset: Volumetric PSNR (dB) and SSIM (\%) measurements between the synthesized and ground truth images in the test set of the in vivo brain dataset are given as mean $\pm$ std. The measurements are provided for proposed and competing methods for all many-to-one synthesis tasks: 1)~\TtwoFlairTonecTone, 2)~\ToneFlairTonecTtwo, 3)~\ToneTtwoTonecFlair, 4)~\ToneTtwoFlairTonec. sGAN-A denotes the sGAN model trained in the axial orientation, sGAN-C in the coronal orientation, and sGAN-S in the sagittal orientation. Boldface indicates the highest performing method.}
\fontsize{8}{10}\selectfont
\begin{tabular}{ccccccccccccccc}
\cline{2-15}
      & \multicolumn{2}{c}{ProvoGAN} & \multicolumn{2}{c}{sGAN-A} & \multicolumn{2}{c}{sGAN-C} & \multicolumn{2}{c}{sGAN-S} & \multicolumn{2}{c}{vGAN} & \multicolumn{2}{c}{SC-GAN} & \multicolumn{2}{c}{REPLICA} \\ \cline{2-15}
& PSNR & SSIM & PSNR & SSIM & PSNR & SSIM & PSNR & SSIM & PSNR & SSIM & PSNR & SSIM & PSNR & SSIM \\\hline
\multirow{2}{*}{\TtwoFlairTonecTone} & \textbf{26.92} & \textbf{94.24} & 24.17 & 88.23 & 25.31 & 90.78 & 26.22 & 91.17 & 22.73 & 87.73 & 21.70 & 85.60 & 17.14 & 83.34   \\ 
& \textbf{$\pm$ 4.55} & \textbf{$\pm$ 3.41} & $\pm$ 3.83 & $\pm$ 4.68 & $\pm$ 4.12 & $\pm$ 4.09 & $\pm$ 3.09 & $\pm$ 3.03 & $\pm$ 3.69 & $\pm$ 3.60 & $\pm$ 2.96 & $\pm$3.45 & $\pm$ 4.43 & $\pm$ 7.43  \\ \hline 

\multirow{2}{*}{\ToneFlairTonecTtwo} & \textbf{26.87} & \textbf{92.79} & 25.67 & 89.76 & 25.98 & 90.95 & 26.85 & 92.11 & 25.48 & 89.75 & 24.48 & 88.63 & 24.68 & 89.06 \\ 
& \textbf{$\pm$ 2.40} & \textbf{$\pm$ 4.22} & $\pm$ 1.75 & $\pm$ 3.20 & $\pm$ 2.18 & $\pm$ 4.10 & $\pm$ 2.38 & $\pm$ 4.11 & $\pm$ 1.82 & $\pm$ 3.93 & $\pm$ 1.72 & $\pm$ 3.54 & $\pm$ 1.94 & $\pm$ 3.16  \\ \hline 

\multirow{2}{*}{\ToneTtwoTonecFlair} & \textbf{25.52} & \textbf{90.39} & 24.50 & 87.21 & 24.95 & 88.09 & 24.81 & 88.39 & 22.94 & 85.67 & 23.07 & 85.88 & 22.70 & 87.63   \\ 
& \textbf{$\pm$ 2.22} & \textbf{$\pm$ 3.06} & $\pm$ 1.84 & $\pm$ 2.73 & $\pm$ 2.03 & $\pm$ 3.18 & $\pm$ 2.21 & $\pm$ 2.67 & $\pm$ 1.61 & $\pm$ 3.23 & $\pm$ 2.23 & $\pm$2.62 & $\pm$2.79 & $\pm$ 2.97 \\ \hline 

\multirow{2}{*}{\ToneTtwoFlairTonec} & \textbf{29.67} & \textbf{94.14} & 28.53 & 91.70 & 28.48 & 89.91 & 28.75 & 92.06 & 27.21 & 89.11 & 27.57 & 90.19 & 24.44 & 90.09 \\ 
& \textbf{$\pm$ 2.23} & \textbf{$\pm$ 2.09} & $\pm$ 1.98 & $\pm$ 2.34 & $\pm$ 2.11 & $\pm$ 3.09 & $\pm$ 2.23 & $\pm$ 2.44 & $\pm$ 1.46 & $\pm$ 1.74 & $\pm$ 1.68 & $\pm$ 2.00 & $\pm$ 2.49 & $\pm$ 3.03 \\ \hline 

\end{tabular}
\label{tab: hacet_many2one_compare}
\justify
{ }
\end{table*}

\subsection{Demonstrations Against Hybrid Models}
Having demonstrated the superior performance of ProvoGAN against several state-of-the-art cross-sectional and volumetric models, we conducted additional experiments to comparatively evaluate it against alternative volumetrization methods. In particular, ProvoGAN was compared with hybrid models based on fusion (M\textsuperscript{3}NET) and transfer learning strategies (TransferGAN) that both involve a mixture of cross-sectional and volumetric mappings (see Section \ref{comp_met_section} for details). Experiments were performed on the IXI dataset for accelerated MRI reconstruction and multi-contrast MRI synthesis. For reconstruction, \Tone- and \Ttwo-weighted image recovery tasks at four distinct acceleration factors ($R=4,8,12,16$) were examined. Table \ref{tab: IXI_recon_compare_vs_2.5D} reports performance measurements for the methods under comparison. ProvoGAN yields superior performance compared to both hybrid models in all reconstruction tasks ($p<0.05$), where on average it achieves $1.87$ dB higher PSNR and $4.34$ \% higher SSIM compared to M\textsuperscript{3}NET, and $1.83$ dB higher PSNR and $5.02$ \% higher SSIM compared to TransferGAN. Meanwhile, \ToneTtwoPD, \TonePDTtwo, \TtwoPDTone~recovery tasks were considered for synthesis. The respective measurements are reported in Table \ref{tab: IXI_many2one_compare_vs_2.5D}. We find that ProvoGAN again maintains enhanced recovery performance in all synthesis tasks ($p<0.05$), where it achieves an average of $2.94$ dB higher PSNR and $2.81$ \% higher SSIM compared to M\textsuperscript{3}NET, and $0.87$ dB higher PSNR and $1.76$ \% higher SSIM compared to TransferGAN.

\begin{figure*}[t]
\centering
\includegraphics[width=0.85\linewidth]{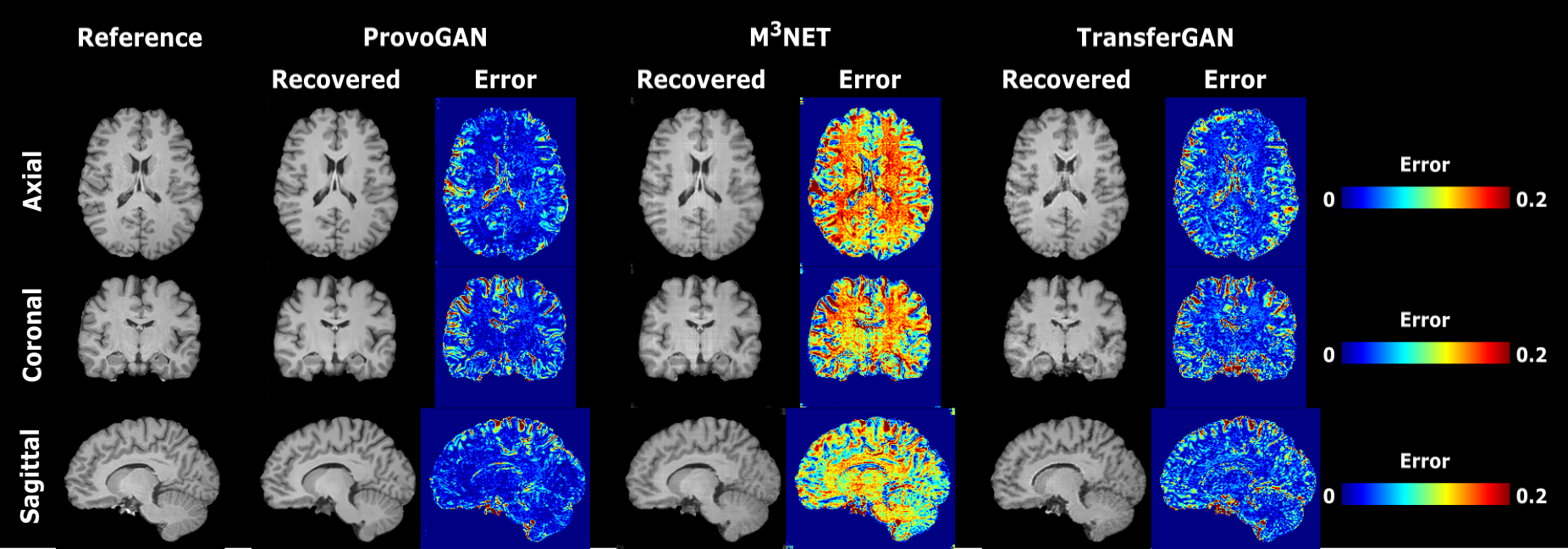}
\caption{ The proposed ProvoGAN method is demonstrated on the IXI dataset against hybrid models (M\textsuperscript{3}NET and TransferGAN) for \Tone-weighted image synthesis from \Ttwo- and PD-weighted images. Representative results are displayed for all methods under comparison together with the ground truth target images (first column). The first row displays results for the axial orientation, the second row for the coronal orientation, and the third row for the sagittal orientation. Error was taken as the absolute difference between the reconstructed and reference images (see colorbar). Overall, the proposed method offers sharper and more accurate delineation of tissues than the competing methods. Furthermore, ProvoGAN better alleviates residual discontinuity artifacts compared to M\textsuperscript{3}NET.}
\label{fig:T1syn_IXI_many2one_hybrid_compare}
\end{figure*}

Quantitative improvements that ProvoGAN offers are also visible in representative images displayed in Supp. Fig. 3 for reconstruction and in Fig. \ref{fig:T1syn_IXI_many2one_hybrid_compare} for synthesis. The M\textsuperscript{3}NET model that performs 3D fusion of 2D model outputs at separate orientations moderately increases contextual sensitivity, but suffers from residual discontinuity artifacts and over-smoothing. Meanwhile, the TransferGAN model that transfers pretrained weights from a 2D model to condition the final 3D model improves learning behavior, but it suffers from elevated model complexity leading to loss of spatial resolution and structural details. In contrast, ProvoGAN yields enhanced recovery performance in all orientations with higher contextual consistency and sensitivity to structural details in the recovered images.

\begin{figure*}[ht]
\centering
\includegraphics[width=0.99\linewidth]{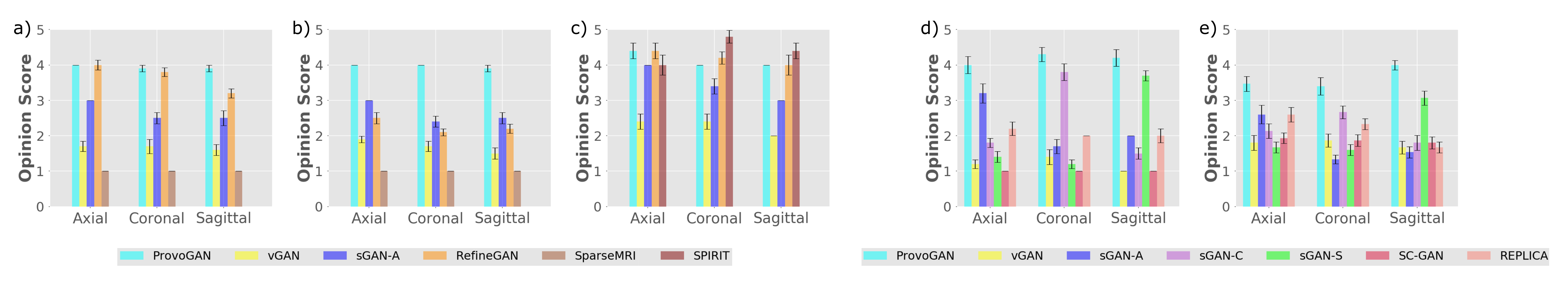}
\caption{Methods were compared in terms of radiological opinion scores for three reconstruction tasks: a) single-coil reconstruction of \Tone-weighted images undersampled by $R=8$ in the IXI dataset, b) single-coil reconstruction of \Ttwo-weighted images undersampled by $R=8$ in the IXI dataset, c) multi-coil reconstruction of PD-weighted images undersampled by $R=8$ in the in vivo knee dataset, and for two synthesis tasks: d) many-to-one synthesis task of \TtwoPDTone~in the IXI dataset, e) many-to-one synthesis task of \ToneTtwoTonecFlair~in the in vivo brain dataset. The quality of the recovered axial, coronal, and sagittal cross-sections were rated by an expert radiologist by assessing their similarity to the reference cross-sections via a five-point scale (0: unacceptable, 1: very poor, 2: limited, 3: moderate, 4: good, 5: perfect match). Figure legends denote the colors used for marking the methods under comparison.}
\label{fig:radeval}
\end{figure*}

\renewcommand{\tabcolsep}{5pt}
\begin{table}[!t]
\centering
\caption{ Comparison of Volumetrization Approaches for Reconstruction in the IXI Dataset: Volumetric PSNR (dB) and SSIM (\%) measurements between the reconstructed and ground truth images in the test set in the IXI dataset are given as mean $\pm$ std. The measurements are reported for the proposed ProvoGAN and competing M\textsuperscript{3}NET and TransferGAN methods for four distinct acceleration factors ($R=4,8,12,16$). Boldface indicates the best performing method.}
\fontsize{8}{10}\selectfont
\begin{tabular}{cccccccc}
\cline{3-8}
& & \multicolumn{2}{c}{ProvoGAN} & \multicolumn{2}{c}{M\textsuperscript{3}NET} & \multicolumn{2}{c}{TransferGAN}  \\ \cline{3-8}\vspace{-1mm}
& & PSNR & SSIM & PSNR & SSIM & PSNR & SSIM     \\ \hline
\multirow{4}{*}{R=4} & \multirow{2}{*}{\Tone} & \textbf{35.25} & \textbf{96.73} & 31.25 & 92.07 & 31.80 & 89.11   \\ 
& & \textbf{$\pm$ 1.78} & \textbf{$\pm$ 0.57} & $\pm$ 1.87 & $\pm$ 1.18 & $\pm$ 0.88 & $\pm$ 1.10  \\ \cline{2-8}
& \multirow{2}{*}{\Ttwo} & \textbf{35.50} & \textbf{96.08} & 33.85 & 92.88 & 34.18 & 94.12  \\ 
& & \textbf{$\pm$ 2.62} & \textbf{$\pm$ 1.07} & $\pm$ 2.73 & $\pm$ 1.12 & $\pm$ 1.30 & $\pm$ 0.84   \\ \hline
\multirow{4}{*}{R=8} & \multirow{2}{*}{\Tone}  & \textbf{31.38} & \textbf{94.93} & 29.61 & 90.78 & 30.37 & 88.44   \\
& & \textbf{$\pm$ 1.26} & \textbf{$\pm$ 0.86} & $\pm$ 1.03 & $\pm$ 1.05 & $\pm$ 1.22 & $\pm$ 0.99  \\ \cline{2-8}
& \multirow{2}{*}{\Ttwo} & \textbf{33.49} & \textbf{95.92} & 32.12 & 93.09 & 31.69 & 93.49  \\ 
& & \textbf{$\pm$ 2.21} & \textbf{$\pm$ 1.01} & $\pm$ 2.25 & $\pm$ 1.32 & $\pm$ 0.83 & $\pm$ 0.71  \\ \hline
\multirow{4}{*}{R=12} & \multirow{2}{*}{\Tone} & \textbf{29.67} & \textbf{92.48} & 28.55 & 87.61 & 28.79 & 85.31   \\ 
& & \textbf{$\pm$ 0.91} & \textbf{$\pm$ 0.90} & $\pm$ 1.05 & $\pm$ 1.46 & $\pm$ 0.89 & $\pm$ 1.54  \\ \cline{2-8}
& \multirow{2}{*}{\Ttwo} & \textbf{30.41} & \textbf{91.98} & 30.29 & 88.31 & 29.18 & 90.49  \\ 
& & $\pm$ \textbf{1.03} & \textbf{$\pm$ 1.40} &  $\pm$ 1.00 &  $\pm$ 1.08 &  $\pm$ 0.81 &  $\pm$ 0.94  \\ \hline
\multirow{4}{*}{R=16} & \multirow{2}{*}{\Tone} & \textbf{29.15} & \textbf{91.40} & 25.83 & 85.59 & 27.61 & 83.31  \\ 
& & \textbf{$\pm$ 1.09} & \textbf{$\pm$ 1.09} &  $\pm$ 1.13 &  $\pm$ 1.90 &  $\pm$ 0.88 &  $\pm$ 1.65  \\ \cline{2-8}
& \multirow{2}{*}{\Ttwo} & \textbf{30.66} & \textbf{93.74} & 29.02 & 88.03 & 27.28 & 88.84  \\ 
& & \textbf{$\pm$ 1.60} & \textbf{$\pm$ 1.35} & $\pm$ 1.24 & $\pm$ 1.71 & $\pm$ 1.69 & $\pm$ 1.36  \\ \hline
\end{tabular}
\label{tab: IXI_recon_compare_vs_2.5D}
\end{table}

\renewcommand{\tabcolsep}{3pt}
\begin{table}[!t]
\centering
\caption{ Comparison of Volumetrization Approaches for Synthesis in the IXI Dataset: Volumetric PSNR (dB) and SSIM (\%) measurements between the synthesized and ground truth images in the test set in the IXI dataset are given as mean $\pm$ std. The measurements are reported for the proposed ProvoGAN and competing M\textsuperscript{3}NET and TransferGAN methods for all synthesis tasks: 1)~\TtwoPDTone, 2)~\TonePDTtwo, 3)~\ToneTtwoPD. Boldface indicates the highest performing method.}
\fontsize{8}{10}\selectfont
\begin{tabular}{ccccccc}
\cline{2-7}
& \multicolumn{2}{c}{ProvoGAN} & \multicolumn{2}{c}{M\textsuperscript{3}NET} & \multicolumn{2}{c}{TransferGAN}  \\ \cline{2-7}\vspace{-1mm}
& PSNR & SSIM & PSNR & SSIM & PSNR & SSIM     \\ \hline
\multirow{2}{*}{\TtwoPDTone} & \textbf{24.15} & \textbf{90.33} & 20.85 & 86.81 & 23.84 & 87.09 \\
 & \textbf{$\pm$ 2.80} & \textbf{$\pm$ 4.47} & $\pm$ 4.08 & $\pm$ 5.00 & $\pm$ 3.37 & $\pm$ 4.09 \\ \hline
\multirow{2}{*}{\TonePDTtwo} & \textbf{28.97} & \textbf{94.17} & 23.79 & 89.82 & 27.78 & 93.10 \\
 & \textbf{$\pm$ 2.91} & \textbf{$\pm$ 4.16} & $\pm$ 1.23 & $\pm$ 3.12 & $\pm$ 2.47 & $\pm$ 3.98 \\ \hline
 \multirow{2}{*}{\ToneTtwoPD} & \textbf{29.81} & \textbf{95.41} & 29.48 & 94.84 & 28.71 & 94.45 \\
 & \textbf{$\pm$ 2.96} & \textbf{$\pm$ 2.75} & $\pm$ 2.24 & $\pm$ 2.41 & $\pm$ 1.89 & $\pm$ 2.52 \\ \hline
\end{tabular}
\label{tab: IXI_many2one_compare_vs_2.5D}
\end{table}

\subsection{Radiological Evaluation}
Quantitative performance assessments in MRI recovery tasks clearly indicate that ProvoGAN outperforms competing volumetric and cross-sectional models in terms of image quality metrics (i.e., PSNR, SSIM). Yet, an important question concerns to what extent these quantitative improvements will benefit diagnostic assessments. Given its ability to effectively capture global context as well as fine structural details, we hypothesized that ProvoGAN will recover MR images of equivalent or higher diagnostic value than competing models. To test this hypothesis, radiological evaluations were performed on images recovered via ProvoGAN, sGAN and vGAN, as well as SC-GAN, REPLICA for synthesis, and RefineGAN, compressed-sensing methods (SparseMRI and SPIRIT) for reconstruction (see Section 2.7 for details). Opinion scores for all methods in axial, coronal, and sagittal orientations denoted as (OS\textsubscript{A}, OS\textsubscript{C}, OS\textsubscript{S}) are reported in Fig. \ref{fig:radeval}a-c for reconstruction and in Fig. \ref{fig:radeval}d,e for synthesis. Across reconstruction tasks, ProvoGAN achieves average opinion scores of ($4.13, 3.97, 3.93$) where sGAN yields ($3.33, 2.77, 2.67$), vGAN yields ($2.00, 1.93, 1.70$), RefineGAN yields ($3.63, 3.67, 3.13$) and compressed-sensing reconstructions yield ($2.00, 2.27, 2.13$). Across synthesis tasks, ProvoGAN achieves average opinion scores of ($3.73, 3.85, 4.10$) whereas vGAN yields ($1.50, 1.63, 1.33$), SC-GAN yields ($1.47, 1.43, 1.40$) and REPLICA yields ($2.40, 2.17, 1.83$). Meanwhile, transverse sGAN models~\footnote{In radiological evaluation, an sGAN model is called transverse for those opinion scores given for the orientation where that sGAN model is trained, e.g., sGAN-A for OS\textsubscript{A}.} maintain ($2.90, 3.23, 3.38$) and longitudinal sGAN models~\footnote{In radiological evaluation, an sGAN model is called longitudinal for those opinion scores given for the orientation where that sGAN model is not trained, e.g., sGAN-A for OS\textsubscript{C} or OS\textsubscript{S}.} yield ($1.64, 1.81, 1.47$). Overall, ProvoGAN outperforms all competing methods in synthesis ($p<0.05$, Wilcoxon signed-ranked test) and reconstruction ($p<0.05$) tasks, except for RefineGAN and SPIRIT in the in-vivo knee dataset where the three methods perform similarly ($p>0.05$). In synthesis, ProvoGAN surpasses not only longitudinal sGAN models in transverse dimensions but also transverse sGAN models in transverse dimensions for which they have been optimized.

Radiological evaluations were also performed on recovered images from M\textsuperscript{3}NET and TransferGAN, as competing volumetrization baselines. Opinion scores for all volumetrization approaches in axial, coronal, and sagittal orientations denoted as (OS\textsubscript{A}, OS\textsubscript{C}, OS\textsubscript{S}) are reported for the IXI dataset in Supp. Fig. \ref{fig:rad_eval_for_hybrid_compare}. Across reconstruction tasks, ProvoGAN achieves average opinion scores of ($4.00, 3.95, 3.90$) where M\textsuperscript{3}NET yields ($2.95, 3.25, 3.50$) and TransferGAN yields ($2.80, 2.75, 2.8$). For synthesis, ProvoGAN achieves average opinion scores of ($4.00, 4.3, 4.2$) where M\textsuperscript{3}NET yields ($2.90, 3.20, 3.00$) and TransferGAN yields ($1.90, 2.10, 1.9$). Overall, ProvoGAN outperforms  competing volumetrization methods in both synthesis ($p<0.05$) and reconstruction ($p<0.05$) tasks. Taken together, these findings strongly suggest that ProvoGAN can offer improved diagnostic quality in accelerated multi-contrast MRI protocols.

\subsection{Complexity of Cross-Sectional Mappings}

Model complexity in deep neural networks depends on several architectural choices, including the number of layers, number of filters in each layer, and kernel size. To minimize bias in performance comparisons, here we aligned the architectural designs as closely as possible among the competing methods. To do this, the number of layers, number of filters, and kernel size were all kept fixed across methods, except that 2D convolutional kernels were used in sGAN and ProvoGAN whereas 3D convolutional kernels were used in vGAN (see Supp. Figs. 1,2). The precise parameter values were guided by the demanding vGAN model. We selected the parameter set that resulted in maximal model complexity while still allowing us to fit a single vGAN model into the VRAM of the GPUs used to conduct the experiments here. Thus, it is reasonable to consider that vGAN is at its performance limits (see Supp. Tables 5,6). That said, a relevant question is whether and how the relative performance benefits of ProvoGAN over sGAN change with model complexity. To examine this issue, we performed separate experiments on reconstruction and synthesis tasks (see Section \ref{experiments_section} for details) while systematically varying the complexity of the convolutional layers in both models by a factor of $n_f\in\{1/16,1/9,1/4,1,4,9,16\}$. This resulted in seven distinct pairs of models: ProvoGAN($n_f$)-sGAN($n_f$) with $n_f$-fold change in number of learnable network weights. PSNR and SSIM measurements between the recovered and reference volumes reported in Supp. Table 7 demonstrate that ProvoGAN achieves superior reconstruction performance to sGAN at all complexity levels, with on average $1.42$ dB higher PSNR and $3.20$\% higher SSIM ($p<0.05$, Wilcoxon signed-rank test). Meanwhile, PSNR and SSIM measurements reported in Supp. Table 8 indicate that ProvoGAN increases synthesis performance on average by $1.22$ dB in PSNR, and $2.82$\% in SSIM compared to sGAN across complexity levels ($p<0.05$). Taken together, these findings suggest that the benefits of ProvoGAN over sGAN in MRI recovery tasks are reliable across variations in complexity of network layers.

\par
An alternative approach to help improve performance of cross-sectional models without substantially altering model complexity would be to admit inputs from multiple neighboring cross-sections. Given several neighboring cross-sections as input, this would enable a 2D model to incorporate local context in the vicinity of the central cross-section. To examine the utility of this approach in cross-sectional processing, we implemented multi-cross-section variants of the two methods, namely ProvoGAN(multi) and sGAN(multi). Both variants received as input three consecutive cross-sections and learned to recover the central cross-section of the target volume. We postulated that while this approach might increase sGAN performance to a limited degree, ProvoGAN that leverages broad spatial priors across all orientations should still yield superior performance. To test this prediction, we performed comprehensive experiments on the IXI dataset for reconstruction and synthesis tasks (see Section \ref{experiments_section} for details). PSNR and SSIM measurements were performed between the recovered and reference target volumes (see Supp. Tables 9,10).  Overall, ProvoGAN enhances recovery performance compared to sGAN(multi) in both tasks ($p<0.05$), where it achieves on average $1.48$ dB higher PSNR and $6.87$\% higher SSIM in reconstruction, and $0.87$ dB higher PSNR and $1.71$\% higher SSIM in synthesis.  These findings reveal that ProvoGAN outperforms cross-sectional mappings implemented with extended spatial priors across the longitudinal dimension. Note that ProvoGAN and ProvoGAN(multi) perform similarly across tasks ($p>0.05$), whereas sGAN(multi) generally yields on par or higher performance than sGAN. This result suggests that sGAN processing each cross-section independently suffers from loss of spatial context across the longitudinal dimension, whereas sGAN(multi) improves performance by incorporating short-range context across this dimension. In contrast, ProvoGAN(multi) captures limited additional information from multiple neighboring cross-sections, since ProvoGAN readily captures global context across the volume.

\subsection{Data Efficiency}
Volumetric models characteristically involve a substantial amount of parameters that result in heavier demand for training data for successful model training. Instead, ProvoGAN comprises more compact cross-sectional models that can be efficiently trained on limited datasets. To demonstrate the data efficiency of ProvoGAN, we trained independent ProvoGAN and vGAN models while varying the number of training subjects in $n_T=\{5,15,25\}$, yielding ProvoGAN($n_T$) and vGAN($n_T$). For reconstruction, \Tone- and \Ttwo-weighted image recovery tasks in the IXI dataset at four distinct acceleration factors ($R=4,8,12,16$) were considered. Table \ref{tab: IXI_recon_compare_data_efficiency} lists reconstruction performance for all models. As expected, model performance drops for both ProvoGAN and vGAN as number of training subjects is reduced. That said, ProvoGAN($n_T$) outperforms vGAN($n_T$) at all $n_T$ values and in all tasks ($p<0.05$, Wilcoxon signed-rank test), with $4.47$ dB higher PSNR and $11.70$ \% higher SSIM on average. Furthermore, the performance drop due to training with $n_T$=5 versus $n_T$=25 is merely $0.70$ dB PSNR and $1.12$ \% SSIM for ProvoGAN, and $2.70$ dB PSNR and $7.87$ \% SSIM for vGAN. Therefore, ProvoGAN better maintains its reconstruction performance on limited datasets to the extent that ProvoGAN models trained with $n_T=5$ outperform vGAN models trained with $n_T=25$. For synthesis, many-to-one recovery tasks of \TtwoPDTone, \TonePDTtwo, and \ToneTtwoPD~ in the IXI dataset were considered. Measurements reported in Supp. Table 11 again indicate that ProvoGAN($n_T$) outperforms vGAN($n_T$) at all $n_T$ values and in all tasks ($p<0.05$), except for PSNR in \Tone-weighted image recovery. On average, ProvoGAN models yield $1.76$ dB higher PSNR and $2.51$ \% higher SSIM compared to corresponding vGAN models. Moreover, the performance drop due to training with $n_T$=5 versus $n_T$=25 is merely $0.68$ dB PSNR and $1.63$ \% SSIM for ProvoGAN, and $1.19$ dB PSNR and $2.32$ \% SSIM for vGAN. Taken together, these results suggest that the proposed progressive volumetrization approach offers enhanced data efficiency during model training compared to volumetric models. 

The primary factor contributing to the enhanced data efficiency of ProvoGAN is the reduced number of parameters in 2D versus 3D network architectures that elicit improved learning behavior. For the various reconstruction and synthesis tasks examined here, Supp. Table 12 lists comparisons between sGAN, ProvoGAN and vGAN in terms of model complexity, memory load, number of floating point operations per second (FLOPS), and total training time. Compared to vGAN, ProvoGAN reduces model complexity by 3-fold, memory load by 20-fold, FLOPS by 80-fold approximately. Collectively, these benefits empower ProvoGAN to offer improved learning behavior and computationally efficient inference while enhancing the capture of global context in volumetric images. Compared to sGAN, each progression in ProvoGAN naturally maintains the same model complexity and memory load. Yet, it has 3-fold higher FLOPS and train duration than sGAN due to ProvoGAN's sequential learning and inference process across the three rectilinear orientations, which also pushes its train duration beyond that of vGAN. As such, ProvoGAN offers a favorable compromise between contextual sensitivity and data efficiency, albeit at the expense of a prolonged training procedure.  

\begin{table}[!t]
\centering
\caption{ Data Efficiency of ProvoGAN for Reconstruction in the IXI Dataset: Volumetric PSNR (dB) and SSIM (\%) measurements between the reconstructed and ground truth images in the IXI dataset are given as mean $\pm$ std across the test set. Measurements are reported for ProvoGAN and vGAN trained with varying number of subjects, and at four distinct acceleration factors ($R=4,8,12,16$). ProvoGAN($n_T$) and vGAN($n_T$) denote models trained with $n_t\in\{5,15,25\}$ subjects.}
\fontsize{6.5}{10}\selectfont

\begin{tabular}{cccccccccc}
\cline{3-10}
 & & \multicolumn{2}{c}{R=4} & \multicolumn{2}{c}{R=8}  & \multicolumn{2}{c}{R=12} & \multicolumn{2}{c}{R=16}  \\ \cline{3-10}\vspace{-1mm}
 & & PSNR & SSIM & PSNR & SSIM & PSNR & SSIM & PSNR & SSIM    \\ \hline
\multirow{12}{*}{\Tone} &\multirow{2}{*}{ProvoGAN(25)} & 34.37 & 95.47 & 31.63 & 94.79 & 29.07 & 88.03 & 28.29 & 86.73  \\
& & $\pm$ 1.51 & $\pm$ 0.63 & $\pm$ 1.23 & $\pm$ 1.02 & $\pm$ 0.99 & $\pm$ 1.61 & $\pm$ 0.92 & $\pm$ 1.58 \\ \cline{2-10}
 &\multirow{2}{*}{ProvoGAN(15)} & 34.04 & 94.00 & 31.13 & 94.42 & 28.96 & 87.74 & 27.96 & 86.44  \\ 
 & & $\pm$ 1.18 & $\pm$ 0.84 & $\pm$ 1.15 & $\pm$ 1.05 & $\pm$ 0.94 & $\pm$ 1.61 & $\pm$ 1.17 & $\pm$ 1.50 \\ \cline{2-10}
 &\multirow{2}{*}{ProvoGAN(5)} & 33.65 & 92.93 & 30.72 & 93.28 & 28.97 & 89.83 & 28.21 & 88.10  \\ 
 & & $\pm$ 1.33 & $\pm$ 1.03 & $\pm$ 1.40 & $\pm$ 0.98 & $\pm$ 0.99 & $\pm$ 1.39 & $\pm$ 1.08 & $\pm$ 1.35 \\ \cline{2-10}
  &\multirow{2}{*}{vGAN(25)} & 28.90 & 83.27 & 26.75 & 81.91 & 27.14 & 79.92 & 24.95 & 77.50  \\ 
 & & $\pm$ 1.29 & $\pm$ 1.36 & $\pm$ 1.29 & $\pm$ 1.84 & $\pm$ 0.89 & $\pm$ 2.40 & $\pm$ 1.43 & $\pm$ 2.73 \\ \cline{2-10}
 &\multirow{2}{*}{vGAN(15)} & 28.52 & 81.67 & 26.26 & 80.33 & 24.59 & 75.38 & 23.98 & 74.68  \\ 
 & & $\pm$ 1.71 & $\pm$ 1.59 & $\pm$ 1.44 & $\pm$ 2.22 & $\pm$ 1.39 & $\pm$ 2.79 & $\pm$ 1.28 & $\pm$ 2.53 \\ \cline{2-10}
 &\multirow{2}{*}{vGAN(5)} & 26.61 & 74.71 & 24.43 & 73.56 & 23.64 & 68.63 & 23.64 & 69.31  \\ 
 & & $\pm$ 1.25 & $\pm$ 2.33 & $\pm$ 1.14 & $\pm$ 2.20 & $\pm$ 1.27 & $\pm$ 3.16 & $\pm$ 1.18 & $\pm$ 2.63 \\ \hline
 \multirow{12}{*}{\Ttwo} &\multirow{2}{*}{ProvoGAN(25)} & 36.32 & 96.76 & 32.23 & 93.83 & 30.02 & 91.90 & 30.59 & 93.58  \\
 &  & $\pm$ 2.35 & $\pm$ 0.72 & $\pm$ 1.46 & $\pm$0.96 & $\pm$ 1.04 & $\pm$ 1.20 & $\pm$ 1.41 & $\pm$ 1.27 \\ \cline{2-10}
 &\multirow{2}{*}{ProvoGAN(15)} & 35.89 & 96.49 & 32.05 & 93.50 & 29.89 & 91.31 & 28.99 & 90.81  \\ 
 &  & $\pm$ 2.69 & $\pm$ 0.99 & $\pm$ 1.27 & $\pm$ 0.98 & $\pm$ 1.08 & $\pm$ 1.48 & $\pm$ 0.89 & $\pm$ 1.25 \\ \cline{2-10}
 &\multirow{2}{*}{ProvoGAN(5)} & 35.83 & 96.00 & 31.45 & 92.82 & 29.54 & 89.91 & 28.53 & 89.28  \\ 
 &  & $\pm$ 3.03 & $\pm$ 1.46 & $\pm$1.33 & $\pm$ 1.26 & $\pm$ 0.82 & $\pm$ 1.53 & $\pm$ 0.89 & $\pm$ 1.80 \\ \cline{2-10}
  &\multirow{2}{*}{vGAN(25)} & 31.14 & 88.65 & 29.75 & 89.11 & 27.47 & 84.13 & 27.29 & 85.72  \\ 
 &  & $\pm$1.21 & $\pm$1.57 & $\pm$ 0.73 & $\pm$ 1.28 & $\pm$ 0.60 & $\pm$ 1.71 & $\pm$ 0.79 & $\pm$ 1.33\\ \cline{2-10}
 &\multirow{2}{*}{vGAN(15)} & 30.25 & 86.21 & 28.76 & 86.58 & 26.89 & 83.13 & 26.67 & 81.47  \\ 
 &  & $\pm$ 1.07 & $\pm$ 1.87 & $\pm$ 0.56 & $\pm$ 1.50 & $\pm$ 0.50 & $\pm$ 1.57 & $\pm$ 0.44 & $\pm$ 1.81 \\ \cline{2-10}
 &\multirow{2}{*}{vGAN(5)} & 28.70 & 82.03 & 27.63 & 82.67 & 25.54 & 79.22 & 21.62 & 77.10  \\ 
 &  & $\pm$ 0.55 & $\pm$ 2.24 & $\pm$ 0.67 & $\pm$ 1.50 & $\pm$ 0.66 & $\pm$ 1.85 & $\pm$ 1.05 & $\pm$ 1.74 \\ \hline
\end{tabular}
\label{tab: IXI_recon_compare_data_efficiency}
\end{table}

\subsection{Generalizability of Progressive Volumetrization}
\begin{figure*}[t]
\centering
\includegraphics[width=0.8\linewidth]{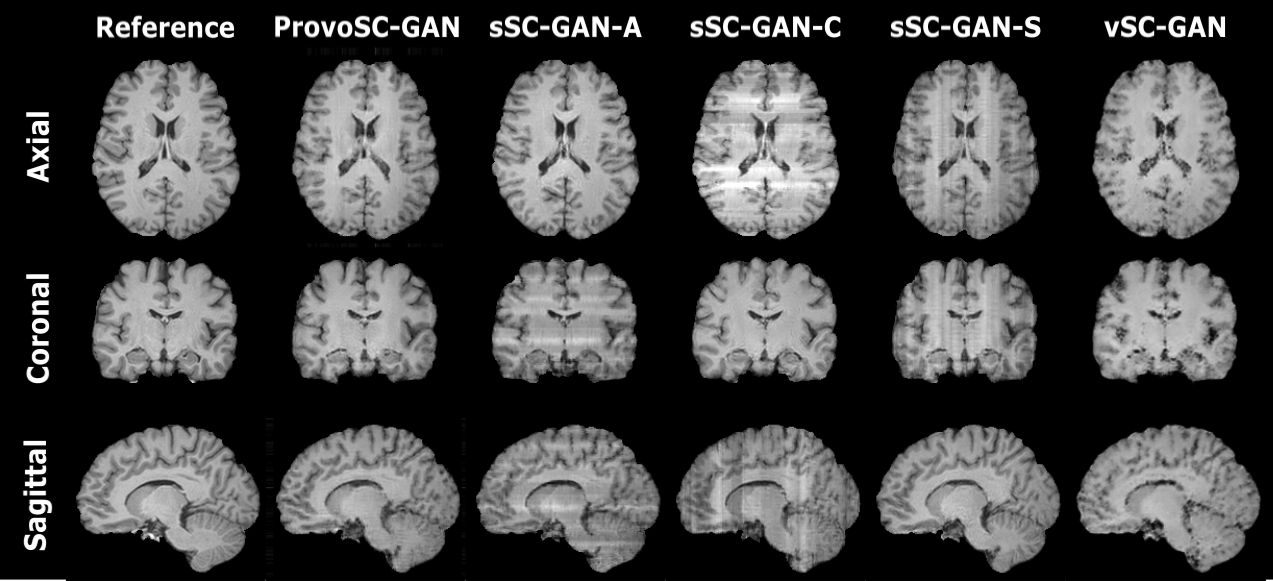}
\caption{ Progressive volumetrization was performed on recently proposed SC-GAN architecture. Representative results for \Tone-weighted image synthesis from \Ttwo- and PD-weighted images in the IXI dataset are displayed. Results are shown for progressively volumetrized (ProvoSC-GAN), cross-sectional (sSC-GAN), and volumetric (vSC-GAN) models, along with the ground truth target images (first column). The first row displays results for the axial orientation, the second row for the coronal orientation, and the third row for the sagittal orientation. Overall, ProvoSC-GAN improves delineation of structural details compared to vSC-GAN, and enhances contextual consistency in the longitudinal dimensions compared to sSC-GAN models.}
\label{fig:T1syn_IXI_many2one_generalization}
\end{figure*}

Here we primarily implemented ProvoGAN on a recent conditional GAN architecture with a ResNet backbone within the generator \citep{Dar2019}. That said, progressive volumetrization can be viewed as a model-agnostic approach that can be adapted to various 2D network architectures. To illustrate the generalizability of ProvoGAN, we performed progressive volumetrization on another state-of-the-art architecture SC-GAN with a U-Net backbone injected with self-attention layers \citep{str_artifacts}. Cross-sectional (sSC-GAN), volumetric (vSC-GAN) and volumetrized (ProvoSC-GAN) variants of this architecture were built. Demonstrations were performed for \ToneTtwoPD, \TonePDTtwo, and \TtwoPDTone~synthesis tasks on the IXI dataset. Resulting PSNR and SSIM measurements are listed in Table \ref{tab: IXI_many2one_compare_with_Unet}, where ProvoSC-GAN achieves in the range [$0.35,2.42$] dB higher PSNR and [$0.47,3.05$] \% higher SSIM compared to sSC-GAN and vSC-GAN ($p<0.05$). The superior synthesis quality offered by ProvoSC-GAN is also visible in representative results displayed in Fig. \ref{fig:T1syn_IXI_many2one_generalization}. Specifically, sSC-GAN models manifest discontinuity artifacts across the respective longitudinal dimensions, and vSC-GAN is suboptimal in recovering fine-structural details. In contrast, ProvoSC-GAN alleviates the limitations of both sSC-GAN and vSC-GAN models to enable more detailed and spatially-coherent tissue depiction. Taken together, these results strongly suggest that the proposed progressive volumetrization strategy can be extended to other network architectures while preserving its advantages against cross-sectional and volumetric mappings.

\renewcommand{\tabcolsep}{3pt}
\begin{table*}[!t]
\centering
\caption{ Progressive Volumetrization of the SC-GAN Architecture: Volumetric PSNR (dB) and SSIM (\%) measurements between the synthesized and ground truth images in the test set in the IXI dataset are given as mean $\pm$ std. Measurements are provided for proposed and competing methods for all many-to-one synthesis tasks: 1)~\TtwoPDTone, 2)~\TonePDTtwo, 3)~\ToneTtwoPD. sSC-GAN-A denotes the sSC-GAN model trained in the axial orientation, sSC-GAN-C in the coronal orientation, and sSC-GAN-S in the sagittal orientation. Boldface indicates the highest performing method.}
\fontsize{8}{10}\selectfont
\begin{tabular}{ccccccccccc}
\cline{2-11}
   & \multicolumn{2}{c}{ProvoSC-GAN} & \multicolumn{2}{c}{sSC-GAN-A} & \multicolumn{2}{c}{sSC-GAN-C} & \multicolumn{2}{c}{sSC-GAN-S} & \multicolumn{2}{c}{vSC-GAN}  \\ \hline

& PSNR & SSIM & PSNR & SSIM & PSNR & SSIM & PSNR & SSIM & PSNR & SSIM \\ \hline
\multirow{2}{*}{\TtwoPDTone} &  \textbf{23.64} & \textbf{88.27} & 22.74 & 85.63 & 23.03 & 86.94 & 23.30 & 87.53 & 22.58 & 85.32   \\ 
& \textbf{$\pm$ 3.12} & \textbf{$\pm$ 4.50} & $\pm$ 2.27 & $\pm$ 3.74 & $\pm$ 2.42 & $\pm$ 3.80 & $\pm$ 2.95 & $\pm$ 4.46 & $\pm$ 2.99 & $\pm$ 4.00  \\ \hline

\multirow{2}{*}{\TonePDTtwo} & \textbf{28.20} & \textbf{93.26} & 27.79 & 92.64 & 27.72 & 92.45 & 28.09 & 93.22 & 25.29 & 89.81  \\ 
& \textbf{$\pm$ 2.97} & \textbf{$\pm$ 4.53} & $\pm$ 2.71 & $\pm$ 4.44 & $\pm$ 2.74 & $\pm$ 4.53 & $\pm$ 2.86 & $\pm$ 4.35 & $\pm$ 1.95 & $\pm$ 4.31 \\ \hline

\multirow{2}{*}{\ToneTtwoPD} & \textbf{29.66} & \textbf{94.86} & 28.83 & 94.02 & 28.93 & 93.93 & 29.05 & 94.21 & 26.36 & 92.12     \\ 
& \textbf{$\pm$ 2.37} & \textbf{$\pm$ 2.96} & $\pm$ 2.12 & $\pm$ 2.86 & $\pm$ 2.11 & $\pm$ 2.95 & $\pm$ 1.92 & $\pm$ 2.46 & $\pm$ 1.41 & $\pm$ 2.59 \\ \hline 

\end{tabular}
\label{tab: IXI_many2one_compare_with_Unet}
\end{table*}

\section{Discussion}
Here, we introduced a progressively volumetrized deep generative model (ProvoGAN) for accelerated MRI that decomposes complex volumetric image recovery tasks into a series of cross-sectional mappings task-optimally ordered across individual rectilinear orientations. This progressive decomposition empowers ProvoGAN to learn both global contextual priors and fine-structural details in each orientation with enhanced data efficiency. Comprehensive evaluations on brain and knee MRI datasets illustrate the superior performance of ProvoGAN against state-of-the-art volumetric and cross-sectional models. Compared to volumetric models, ProvoGAN better captures fine structural details while at the same time maintaining lower instantaneous model complexity. As subtasks in ProvoGAN take single cross-sections as separate training samples, the effective size of the training set is expanded. Therefore, for a given model complexity, ProvoGAN demands an order of magnitude lower memory load than volumetric models. Compared to cross-sectional models, ProvoGAN mitigates discontinuity artifacts across the longitudinal dimensions and extends reliable capture of structural details from transverse onto longitudinal dimensions. Importantly, ProvoGAN offers this advanced recovery performance for the same budget of model complexity and memory load as cross-sectional models,  albeit at the expense of a three-times prolonged training procedure due to sequential learning.  
\par

Several recent studies in medical image processing have focused on improving learning behavior in volumetric models. An earlier group of studies proposed spatially-focused 3D models to process volumetric patches during MRI recovery \citep{ex_mod_prop,dict_learning_im_synth,patch_based_one_to_one_4,Jog2017b,les_seg,example_based,patch_based_one_to_one_2,unsup_cross_model_synth,modality_prop}. Patch-based models that restrict the spatial extent of network inputs-outputs can reduce model size to offer performance improvements. That said, a compact 3D patch incorporates context along the longitudinal axis at the expense of narrowing coverage in the in-plane dimensions. Since patches are processed independently, the predicted volumes might also manifest discontinuity artifacts. These limitations can undercut potential benefits of patch-based processing for 3D models. Later studies proposed hybrid models to bridge 2D and 3D models in an effort to combine their strengths \citep{volume_volume_fuse,provolike_1,shan20183,liu20183d}. Among hybrid methods are fusion models that aggregate the outputs of parallel 2D models in multiple orientations \citep{provolike_1,volume_volume_fuse}. Fusion models employ a cascade of 2D and 3D processing, so they incur high computational complexity, and sensitivity to fine structural details might be limited by the aggregation process across orientations. An alternative approach is transfer learning from 2D onto 3D models to facilitate model training \citep{shan20183,liu20183d}. A full-scale 3D model is leveraged in transfer learning methods that lead to elevated model complexity, and a similar computational footprint to conventional 3D models. In contrast, ProvoGAN is composed of a sequence of 2D models, without any 3D module, resulting in substantially lower model complexity and computational load.

\par
 An alternative approach to volumetrization in medical imaging tasks has been to revise cross-sectional models to help them better incorporate spatial context.  In \citep{Zheng2018}, enhanced spatial consistency during cardiac image segmentation was aimed by performing cross-sectional mapping on short-axis images sequentially across neighboring cross-sections. The segmentation map from the earlier cross-section was used to initialize the map for the current cross-section \citep{Zheng2018}. While benefits were demonstrated over 2D processing, this approach limits accumulation of contextual information to a single direction and to neighboring cross-sections. A different strategy for cardiac MRI segmentation was to perform cross-sectional mapping in short-axis orientation while latent representations captured via an autoencoder on a multitude of view orientations were fused at intermediate layers \citep{shape_priors}. The complexity of the resulting models scales with the number of additional views included, and this promising approach might be limited in applications where a multitude of different views on the same anatomy are unavailable. In \citep{voxel_voxel_segment_1,voxel_voxel_segment_2,multi_view_nodule_classification}, MR images at three rectilinear views that span across a target voxel were incorporated as inputs to a cross-sectional model during segmentation or classification tasks. Classifying a center voxel by fusing information across orientations might limit flow of contextual information from nearby voxels not covered by the input image views. 

\par

Our analyses involved brain and knee MRI datasets mostly collected at near-isotropic resolution. That said, the in vivo brain dataset was acquired under a diverse set of imaging protocols with varying spatial resolution. Note that images from \Ttwo-weighted and FLAIR acquisitions had considerably poorer resolution in the longitudinal dimension, although they were registered to the MNI template with 1-mm isotropic resolution prior to modeling, In our experiments, models were built to synthesize \Tone-weighted and \Tonec-weighted images, where thick-slice acquisitions were on the input side. Models were also built to synthesize \Ttwo-weighted and FLAIR images where thick-slice acquisitions were on the output side. We find that ProvoGAN offers enhanced recovery in both cases comprising a mixture of isotropic and anisotropic resolutions. Yet, contextual dependencies in the longitudinal dimension might be weaker for datasets uniformly acquired with thick slices, which in turn can limit the benefits of volumetrization. We plan to investigate this important issue in future studies by evaluating volumetrization performance on datasets with systematically varied slice thickness.

\par
Several technical lines of development can be taken to further improve the performance and reliability of progressive volumetrization. In this study, ProvoGAN was independently demonstrated for mainstream MRI reconstruction and synthesis tasks. ProvoGAN can also be adopted for a joint reconstruction-synthesis task to further improve the utility and practicality of accelerated multi-contrast MRI protocols \citep{darsynergistic,iglesias2021joint,semisupervised_my}. Here ProvoGAN was trained using a fully-supervised learning framework, which assumes the availability of datasets containing high-quality ground truth target images. However, compiling large datasets with high-quality references might prove difficult due to various concerns such as patient motion or examination costs \citep{liu2021rethinking}. An alternative would be to train ProvoGAN in a self-supervised setting for reconstruction tasks \citep{selfsupervised,demirel2021improved,cole2020unsupervised,korkmaz2021unsupervised} or in a semi-supervised setting for synthesis tasks \citep{semisupervised_my} to alleviate dependency on high-quality training datasets.  Another avenue of development concerns the generalization of ProvoGAN to work on nonrectilinear orientations \citep{ramzi2021density,sun2020non,motyka2021k}. While ProvoGAN was mainly demonstrated for rectilinear acquisitions in this work, similar decompositions can be viable for nonrectilinear sampling schemes in MRI such as radial and spiral acquisitions. Additionally, the number of total progressions in ProvoGAN can be adaptively modified together with the specific ordering of the orientations used in the progressions to enhance task-optimal recovery performance. Instead of performing a separate sequential training of each progression, an end-to-end training of the whole network can also be performed for improved performance by leveraging advanced model parallelism techniques \citep{parallelism_1}.

\par
In this work, we demonstrated the proposed progressive volumetrization approach via a data-driven deep generative model that performs recovery in the image domain. Although image-to-image learning of deep models proves popular in MRI recovery tasks \citep{wang2021deep,Yang2018,dai2020multimodal,Quan2018c,cole2021analysis,wang2020synthesize}, there are other successful approaches to MRI processing based on k-space-to-k-space learning \citep{han2020}, k-space-to-image learning \citep{eo2018kiki,zhu2018image,Akcakaya2019,wang2019dimension}, or model-based learning with unrolled network architectures \citep{sun2016deep,zhang2018ista,aggarwal2018modl,duan2019vs,yang2018admm}. In principle, ProvoGAN can also be implemented to volumetrize models based on these recent powerful approaches. Thus, it remains important future work to investigate the potential benefits of progressive volumetrization to the contextual sensitivity of a broader family of recovery methods \citep{wang2021deep,yang2021synthesizing,chen2021synthesizing,dalmaz2021resvit,tavaf2021grappa,hu2021run}. 

\par
In summary, here we introduced a progressive volumetrization framework for deep network models to process 3D imaging datasets. The superior learning behavior of ProvoGAN was demonstrated for inverse problem solutions in two mainstream MRI tasks, reconstruction and synthesis. Yet, our framework can be adopted to other imaging modalities and tasks with minimal effort \citep{jcye_new,dharmony,voxel_voxel_segment_2,multi_view_nodule_classification,rueckert_1,3Dreview,prince_new,Zheng2018,zhou2020review,narnhofer2021bayesian}. As the key idea of subtasking across cross-sectional orientations is domain general, ProvoGAN has further implications for computer vision applications that rely on 3D processing such as style transfer, semantic segmentation and video processing \citep{aberman2020unpaired,chen2021indoor}.  

\section*{Acknowledgments}
This study was supported in part by a TUBITAK 1001 Research Grant (118E256), an EMBO Installation Grant (3028), a TUBA GEBIP 2015 fellowship, a BAGEP 2017 fellowship, and by nVidia under GPU grant.

\bibliographystyle{model2-names}\biboptions{authoryear}
\bibliography{provo}

\newgeometry{left=2cm,right=2cm,bottom=0.1cm}
\onecolumn

\clearpage
\newpage
\setcounter{page}{1}
\begin{center}
\LARGE{Supplementary Materials for "Progressively Volumetrized Deep Generative Models for Data-Efficient Contextual Learning of MR Image Recovery"}
\\~\\
\normalsize{Mahmut Yurt\textsuperscript{a,b,f}, Muzaffer Özbey\textsuperscript{a,b,f}, Salman U. H. Dar\textsuperscript{a,b}, Berk Tınaz\textsuperscript{a,b,c}, Kader K. Oğuz\textsuperscript{b,d}, Tolga Çukur\textsuperscript{a,b,e,*}}\\
\textsuperscript{a} Department of Electrical and Electronics Engineering, Bilkent University, Ankara 06800, Turkey\\
\textsuperscript{b} National Magnetic Resonance Research Center (UMRAM), Bilkent University, Ankara 06800, Turkey\\
\textsuperscript{c} Department of Electrical and Computer Engineering, University of Southern California, Los Angeles 90089, USA\\
\textsuperscript{d} Department of Radiology, Hacettepe University, Ankara 06100, Turkey\\
\textsuperscript{e} Neuroscience Program, Bilkent University, Ankara 06800, Turkey\\
\textsuperscript{f}{denotes equal contribution}
\end{center}
\captionsetup[table]{name= Supp. Table}
\setcounter{table}{0} \renewcommand{\thetable}{\arabic{table}}

\captionsetup[figure]{name= Supp. Figure}
\setcounter{figure}{0} \renewcommand{\thefigure}{\arabic{figure}}

\renewcommand{\tabcolsep}{9pt}
\renewcommand{\arraystretch}{1.6}

\clearpage
\newpage
\section*{Supp. Text 1}

\begin{figure}[h]
\centering
\includegraphics[width=0.99\linewidth]{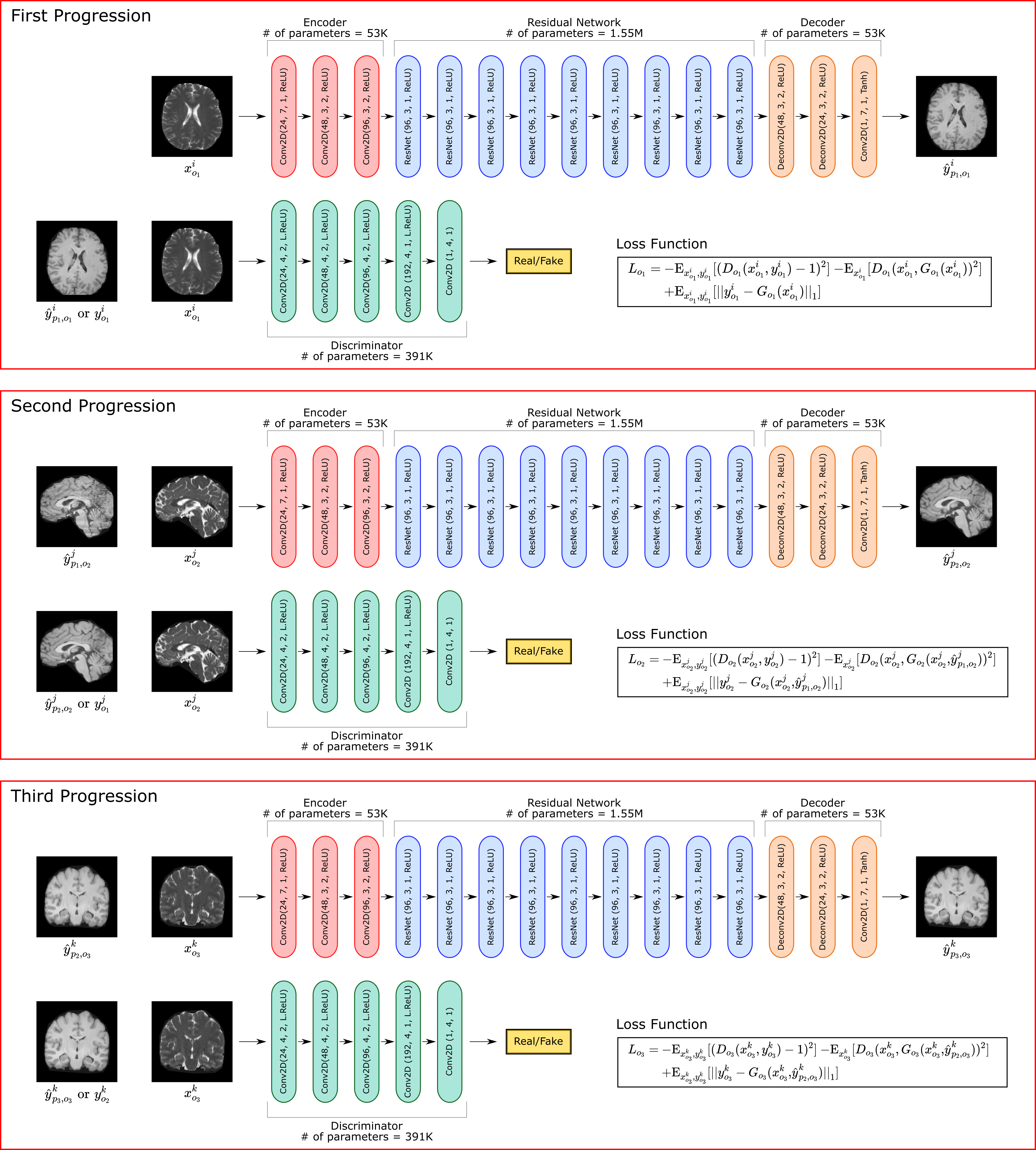}
\caption{The architectural details of the generator and discriminator submodules in ProvoGAN are displayed (here a progression order of \ASC~is used for illustration). The generator architectures are the same for the progressions and consist of an encoder with $3$ convolutional layers: $\mathrm{conv2D}(k=7, f=24, s=1, a=\mathrm{ReLU})$, $\mathrm{conv2D}(k=3, f=48, s=2, a=\mathrm{ReLU})$, $\mathrm{conv2D}(k=3, f=96, s=2, a=\mathrm{ReLU})$, a residual network of $9$ ResNet blocks: $9\times \mathrm{ResNet2D}(k=3, f=96, s=1, a=\mathrm{ReLU})$, and a decoder of $3$ convolutional layers: $\mathrm{deconv2D}(k=3, f=48, s=2, a=\mathrm{ReLU}), \mathrm{deconv2D}(k=3, f=24, s=2, a=\mathrm{ReLU}), \mathrm{conv2D}(k=7, f=1, s=1, a=\mathrm{Tanh})$, where $k$ denotes kernel size, $f$ denotes number of filters, $s$ denotes stride, and $a$ denotes activation function. Similarly, the discriminator architectures are identical for the progresions and consist of a convolutional network of $5$ convolutional layers in series: $\mathrm{conv2D}(k=4, f=24, s=2, a=\mathrm{leakyReLU})$, $\mathrm{conv2D}(k=4, f=48, s=2, a=\mathrm{leakyReLU})$, $\mathrm{conv2D}(k=4, f=96, s=2, a=\mathrm{leakyReLU})$, $\mathrm{conv2D}(k=4, f=192, s=1, a=\mathrm{leakyReLU})$, $\mathrm{conv2D}(k=4, f=1, s=1, a=\mathrm{none})$.}
\label{fig:provo_arch}
\end{figure}

\noindent\textbf{ProvoGAN: Network Architecture}
\\~\\
The proposed ProvoGAN model is based on conditional generative adversarial networks (GANs) \citep{condgans}. Each GAN model within the progressions consisted of a generator that contains an encoder of $3$ convolutional layers, a residual network of $9$ ResNet blocks, and a decoder of $3$ convolutional layers, and a discriminator that contains a convolutional network of $5$ convolutional layers, all with 2D kernels. The kernel size, number of filters, stride, activation function, and connections of the network layers in the generators and discriminators are provided in Supp. Fig. \ref{fig:provo_arch}. The generator in the first progression received as input the cross-sections of the source volume in the first orientation, whereas the generators in the second and third progressions received as input the cross-sections of both source and previously recovered target volumes in the second and third orientations, respectively. Conditional PatchGAN discriminator architectures were used to effectively incorporate priors. The discriminator received as input the concatenation of source and target cross-sectional images. The source cross-sections were identical to the inputs of the corresponding generator. Ground truth target images were designated as real target images, whereas generator-recovered target images were designated as the fake target images.
\\~\\
\textbf{ProvoGAN: Implementation Details}
\\~\\The training procedure of ProvoGAN comprised three progressive phases, and within each phase the respective pair of generator and discriminator architectures were trained to learn cross-sectional recovery in the given orientation. Hyperparameter-selection procedures for all phases were adopted from \cite{Dar2019}. The number of epochs and relative weighting of the loss terms were optimized via PSNR measurements in the validation set. The generator and discriminator were trained with the ADAM optimizer \citep{adam} ($\beta_1=0.5$, $\beta_2=0.999$) for $100$ epochs. The learning rate was set to $2\times10^{-4}$ in the first $50$ epochs and was linearly decayed to $0$ in the last $50$ epochs. All cross-sectional training samples were processed by the networks in every training epoch with a batch size of $1$, and instance normalization was performed. The optimal relative weighing of the pixel-wise loss to the adversarial loss was selected as $100$. 

\clearpage
\newpage
\section*{Supp. Text 2}

\begin{figure}[h]
\centering
\includegraphics[width=0.99\linewidth]{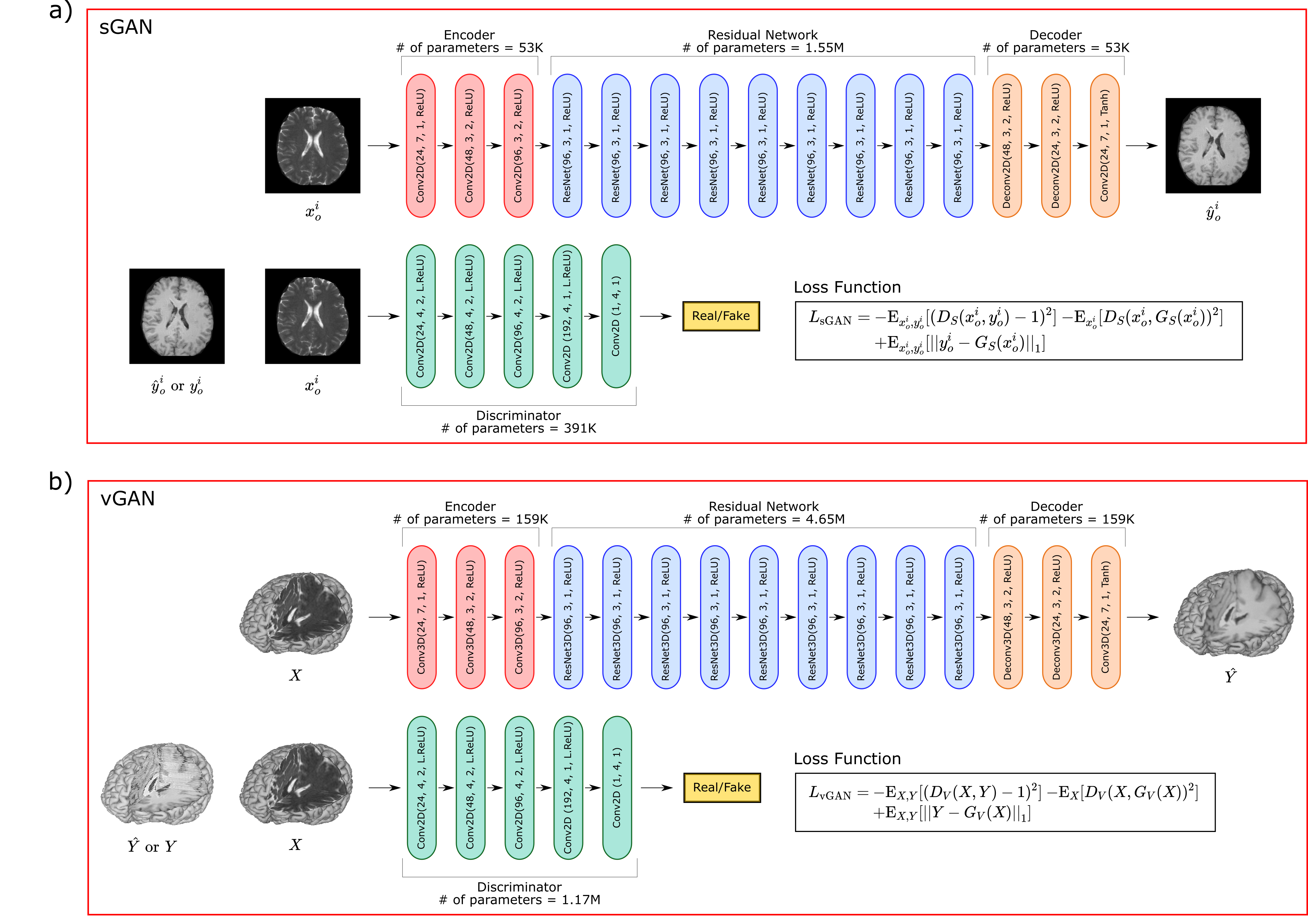}
\caption{a) The architectural details of the generator and discriminator in sGAN are displayed (here an sGAN model trained in the axial orientation referred to as sGAN-A is used for illustration). The generator consists of an encoder with $3$ convolutional layers: $\mathrm{conv2D}(k=7, f=24, s=1, a=\mathrm{ReLU})$, $\mathrm{conv2D}(k=3, f=48, s=2, a=\mathrm{ReLU})$, $\mathrm{conv2D}(k=3, f=96, s=2, a=\mathrm{ReLU})$, a residual network of $9$ ResNet blocks: $9\times \mathrm{ResNet2D}(k=3, f=96, s=1, a=\mathrm{ReLU})$, and a decoder of $3$ convolutional layers: $\mathrm{deconv2D}(k=3, f=48, s=2, a=\mathrm{ReLU}), \mathrm{deconv2D}(k=3, f=24, s=2, a=\mathrm{ReLU}), \mathrm{conv2D}(k=7, f=1, s=1, a=\mathrm{Tanh})$, where $k$ denotes kernel size, $f$ denotes number of filters, $s$ denotes stride, and $a$ denotes activation function. The discriminator consists of a convolutional network of $5$ convolutional layers in series: $\mathrm{conv2D}(k=4, f=24, s=2, a=\mathrm{leakyReLU})$, $conv2D(k=4, f=48, s=2, a=\mathrm{leakyReLU})$, $\mathrm{conv2D}(k=4, f=96, s=2, a=\mathrm{leakyReLU})$, $\mathrm{conv2D}(k=4, f=192, s=1, a=\mathrm{leakyReLU})$, $\mathrm{conv2D}(k=4, f=1, s=1, a=\mathrm{none})$. \\~\\b) The architectural details of the generator and discriminator in vGAN are displayed. The generator consists of an encoder with $3$ convolutional layers: $\mathrm{conv3D}(k=7, f=24, s=1, a=\mathrm{ReLU})$, $\mathrm{conv3D}(k=3, f=48, s=2, a=\mathrm{ReLU})$, $\mathrm{conv3D}(k=3, f=96, s=2, a=\mathrm{ReLU})$, a residual network of $9$ ResNet blocks: $9\times \mathrm{ResNet3D}(k=3, f=96, s=1, a=\mathrm{ReLU})$, and a decoder of $3$ convolutional layers: $\mathrm{deconv3D}(k=3, f=48, s=2, a=\mathrm{ReLU}), \mathrm{deconv3D}(k=3, f=24, s=2, a=\mathrm{ReLU}), \mathrm{conv3D}(k=7, f=1, s=1, a=\mathrm{Tanh})$. The discriminator consists of a convolutional network of $5$ convolutional layers in series: $\mathrm{conv3D}(k=4, f=24, s=2, a=\mathrm{leakyReLU})$, $\mathrm{conv3D}(k=4, f=48, s=2, a=\mathrm{leakyReLU})$, $\mathrm{conv3D}(k=4, f=96, s=2, a=\mathrm{leakyReLU})$, $\mathrm{conv3D}(k=4, f=192, s=1, a=\mathrm{leakyReLU})$, $\mathrm{conv3D}(k=4, f=1, s=1, a=\mathrm{none})$.}
\label{fig:sGAN_vGAN_arch}
\end{figure}

\noindent \textbf{Competing Methods: Network Architectures}\\~\\The cross-sectional sGAN and RefineGAN models as well as cross-sectional components of the hybrid M\textsuperscript{3}NET and TransferGAN models were based on 2D conditional GANs \citep{condgans} with a ResNet backbone \citep{Dar2019}. The resultant GAN model consisted of a generator that contains an encoder of 3 convolutional layers, a residual network of 9 ResNet blocks, and a decoder of 3 convolutional layers, and a discriminator that contains a convolutional network of 5 convolutional layers, all formed with 2D kernels. The kernel size, number of filters, stride, activation function, and connections of the network layers in the generators and discriminators are provided in Supp. Fig \ref{fig:sGAN_vGAN_arch}a. The generators in these models received as input cross-sections of the source contrast in the selected orientation. The discriminators received as input either recovered or ground truth cross-sections of the target contrast concatenated with cross-sections of the source contrast in the same orientation.

On the other hand, the volumetric vGAN model and the volumetric component of the hybrid TransferGAN model were based on 3D conditional GANs \citep{condgans} again with a ResNet backbone. GAN models consisted of a generator that contains an encoder of 3 convolutional layers, a residual network of 9 ResNet blocks, and a decoder of 3 convolutional layers, and a discriminator that contains a convolutional network of 5 convolutional layers, all with 3D kernels. The kernel size, number of filters, stride, activation function, and connections of the network layers in the generators and discriminators are provided in Supp. Fig \ref{fig:sGAN_vGAN_arch}b. For maximized performance, their values were optimized via cross-validation among possible architectures fitting in a single GPU with 11 GB of VRAM (see Supp. Tables \ref{tab: vgan_opt_recon},\ref{tab: vgan_opt_synthesis} for details). The generators in these volumetric models received as input the entire volume of the source contrast. The discriminators received as input the entire volume of either recovered or ground truth target contrast concatenated with the entire volume of the source contrast. Note that an alternative approach for potentially lowering data requirements of volumetric processing might be patch-based 3D models. A compact 3D patch incorporates context along the longitudinal axis at the expense of narrowing coverage in the in-plane dimensions. Since patches are processed independently, the predicted volumes might also manifest discontinuity artifacts. We observed that patch-based 3D models do not offer significant performance benefits over vGAN, and at the same time they can be susceptible to discontinuity artifacts due to patch splitting. Therefore, we preferred not to consider patch-based models in the current study.

The architecture of the hybrid M\textsuperscript{3}NET method was adopted from \cite{volume_volume_fuse}. Meanwhile, SC-GAN embodied a generator based on a U-Net architecture and a discriminator based on a PatchGAN architecture. Both subnetworks contained 3D self-attention modules as described in \cite{str_artifacts}. The generators in these volumetric models again received as input the entire volume of the source contrast. The discriminators received as input the entire volume of either recovered or ground truth target contrast concatenated with the entire volume of the source contrast. The compressed-sensing-based REPLICA, SparseMRI, and SPIRiT methods were implemented as described in \cite{Jog2017b}, \cite{lustig2007}, and \cite{Lustig2010}, respectively. REPLICA received as input the entire volume of the source contrast to recover the target volume. Meanwhile, SparseMRI and SPIRiT received as input cross-sections of the undersampled source volume to recover cross-sections of the target volume.
\\~\\
\textbf{Competing Methods: Implementation Details}
\\~\\For the cross-sectional sGAN and RefineGAN models and cross-sectional components of hybrid M\textsuperscript{3}NET and TransferGAN models, generator-discriminator pairs were trained to learn a recovery task in the selected orientation. Hyperparameter selection procedures were again adopted from \cite{Dar2019}. The number of epochs and relative weighting of the loss terms were optimized via PSNR measurements in the validation set. The generator and discriminator were trained with the ADAM optimizer \citep{adam} ($\beta_1=0.5$, $\beta_2=0.999$) for $100$ epochs. The learning rate was set to $2\times10^{-4}$ in the first $50$ epochs and was linearly decayed to $0$ in the last $50$ epochs. All cross-sectional training samples were processed by the models in every training epoch with a batch size of $1$, and instance normalization was performed.  The relative weighing of the pixel-wise loss to the adversarial loss was selected as $100$ for sGAN and cross-sectional components of M\textsuperscript{3}NET and TransferGAN. For RefineGAN, the relative weighing of the pixel-wise loss and cycle consistency loss of acquired k-space samples to the adversarial loss \citep{Quan2018c} were taken as $10$ and $1$. 
\newpage
For the volumetric vGAN and volumetric component of TransferGAN, generator-discriminator pairs were trained to learn a volumetric recovery task. The number of epochs and relative weighting of the loss terms were optimized via PSNR measurements in the validation set. The generator and discriminator were trained with the ADAM optimizer \citep{adam} ($\beta_1=0.5$, $\beta_2=0.999$) for $200$ epochs. The learning rate was set to $2\times10^{-4}$ for synthesis tasks and $5\times10^{-4}$ for reconstruction tasks in the first $100$ epochs, and was linearly decayed to $0$ in the last $100$ epochs. All training samples were processed by the models in every training epoch with a batch size of $1$, and instance normalization was performed. The relative weighing of the pixel-wise loss to the adversarial loss was selected as $50$ for synthesis tasks and $200$ for reconstruction tasks.

Training procedures of compressed-sensing-based REPLICA, SparseMRI and SPIRIT methods were implemented following procedures outlined in \cite{Jog2017b}, \cite{lustig2007}, and \cite{Lustig2010}, respectively. For REPLICA, hyperparameter selections reported in \cite{Jog2017b} were adopted as they were observed to yield high performance. For SparseMRI and SPIRIT, number of iterations and relative weighting of the loss terms were optimized via PSNR measurements in the validation set. For SparseMRI, number of iterations was selected as $40$. L1 regularization weight of wavelet coefficients was set to 0.002 for \Tone reconstruction with R=4x, 0.004 for \Ttwo  reconstruction with R=16x, and $0.01$ for remaining reconstruction tasks. For SPIRIT, number of iterations was selected as $30$ and interpolation kernel size was set to $5$. L1 and L2 regularization weights were respectively selected as $0.01-0.1$ for R=4,8, $0.01-0.001$ for R=12, and $0.01-0.0001$ for R=16.

\clearpage
\newpage
\section*{Supp. Text 3}

\textbf{Input-Output Image Sizes}
\begin{itemize}
    \item \textbf{Synthesis in the IXI Dataset:} Sizes of the input-output images were ($192\times160\times160$) for vGAN, SC-GAN, REPLICA, M\textsuperscript{3}NET, and TransferGAN, ($192\times160$) for sGAN-A, ($160\times160$) for sGAN-C, ($160\times192$) for sGAN-S where ProvoGAN progressively received images of sizes ($192\times160$), ($160\times160$), and ($160\times192$). 
    \item \textbf{Synthesis in the In vivo Brain Dataset:} Sizes of the input-output images were ($192\times160\times160$) for vGAN, SC-GAN, and REPLICA, ($192\times160$) for sGAN-A, ($160\times160$) for sGAN-C, ($160\times192$) for sGAN-S, where ProvoGAN progressively received images of sizes ($192\times160$), ($160\times160$), and ($160\times192$). 
    \item \textbf{Reconstruction in the IXI Dataset:} Sizes of the input-output images were ($256\times150\times256$) for vGAN, M\textsuperscript{3}NET, and TransferGAN, ($256\times150$) for sGAN, RefineGAN, and SparseMRI, where ProvoGAN progressively received images of sizes ($256\times150$), ($256\times256$), and ($150\times256$). 
    \item \textbf{Reconstruction in the In vivo Knee Dataset:} Sizes of the input-output images were ($320\times320\times256$) for vGAN, ($320\times256$) for sGAN, RefineGAN, and SPIRiT, where ProvoGAN progressively received images of sizes ($320\times256$), ($320\times320$), and ($320\times256$). 
    \item \textbf{Processing:} All models leveraged convolutional input layers that flexibly adapt to varying input image sizes. Thus, no cropping or zero-padding was performed on the original images prior to modeling. 
\end{itemize}
Note that input-output image sizes in the IXI dataset differ between synthesis and reconstruction due to registration to the MNI template and skull-stripping in synthesis (see Section 2.5 for details).

\clearpage
\newpage

\clearpage
\newpage

\clearpage
\newpage
	\begin{figure}[h]
		\centering
		\includegraphics[width=0.65\linewidth]{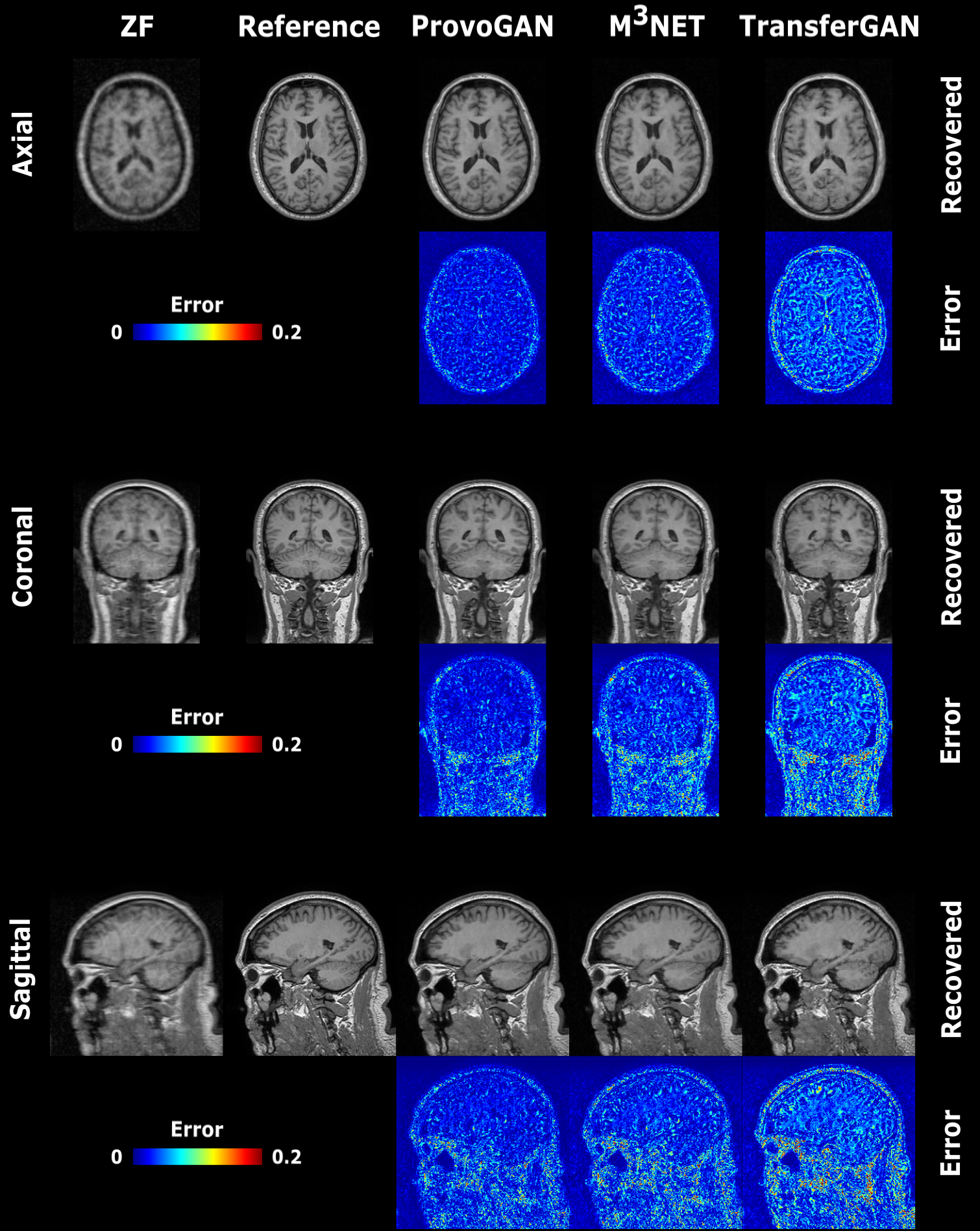}
		\caption{The proposed ProvoGAN method is demonstrated against hybrid models (M\textsuperscript{3}NET and TransferGAN) on the IXI dataset for reconstruction of \Tone-weighted acquisitions undersampled at $R=8$. Representative results are displayed for the methods under comparison together with the undersampled zero-filled source images (first column) and the reference target images (second column). The top two rows display results for the axial, the middle two rows for the coronal, and the last two rows for the sagittal orientation. Error was taken as the absolute difference between the reconstructed and reference images (see colorbar). Overall, the proposed method enables sharper tissue delineation against competing methods, and it improves mitigation of discontinuity artifacts compared to M\textsuperscript{3}NET.}
		\label{fig:IXI_t1_recon_hybrid_compare}
	\end{figure}

\clearpage
\newpage
	\begin{figure}[h]
		\centering
		\includegraphics[width=0.8\linewidth]{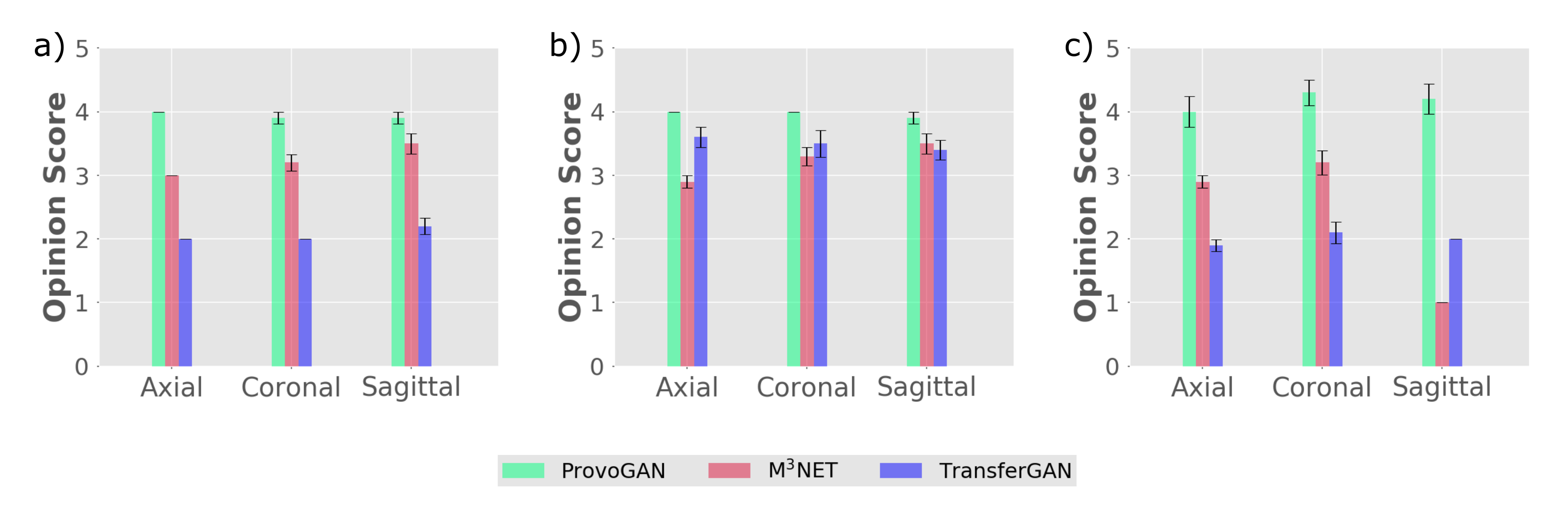}
		\caption{ Volumetrization approaches were compared in terms of radiological opinion scores for three representative tasks: a) reconstruction of \Tone-weighted images undersampled by $R=8$ in the IXI dataset, b) reconstruction of \Ttwo-weighted images undersampled by $R=8$ in the IXI dataset, c) \TtwoPDTone~ synthesis in the IXI dataset. The quality of the recovered axial, coronal, and sagittal cross-sections were rated by an expert radiologist by assessing their similarity to the reference cross-sections via a five-point scale (0: unacceptable, 1: very poor, 2: limited, 3: moderate, 4: good, 5: perfect match). Figure legend denotes the colors used for the methods under comparison.}
		\label{fig:rad_eval_for_hybrid_compare}
	\end{figure}

\clearpage
\newpage

\begin{table}[h]
\caption{Task-Optimal Progression Order for Reconstruction in the IXI Dataset:  Volumetric PSNR (dB) measurements between the reconstructed and ground truth images in the validation set in the IXI dataset are given as mean $\pm$ std. The measurements are provided for all possible progression orders: 1)~\ACS, 2)~\ASC, 3)~\SAC, 4)~\SCA, 5)~\CSA, 6)~\CAS~and acceleration factors: $R=4,8,12,16$. Boldface indicates the highest performing progression sequence.}
\centering
\resizebox{0.85\textwidth}{!}{%
\begin{tabular}{cccccccc}
\cline{2-8}
&  &\textbf{ \ACS} & \textbf{\ASC} & \textbf{\SAC} & \textbf{\SCA} & \textbf{\CSA} & \textbf{\CAS} \\ \hline
\multirow{2}{*}{R=4} & \Tone & 35.12 $\pm$ 1.73 & \textbf{36.12 $\pm$ 1.22} & 35.80 $\pm$ 1.81 & 35.50 $\pm$ 1.63 & 35.84 $\pm$ 1.05 & 35.21 $\pm$ 1.87 \\ \cline{2-8} 
& \Ttwo & 36.42 $\pm$ 1.84 & 36.62 $\pm$ 1.24 & \textbf{36.69} $\pm$ 1.72 & 26.92 $\pm$ 1.02 & 29.50 $\pm$ 0.64 & 36.51 $\pm$ 2.68 \\ \hline
\multirow{2}{*}{R=8} & \Tone & 32.48 $\pm$ 0.42 & 33.51 $\pm$ 0.75 & \textbf{33.52 $\pm$ 0.70}  & 33.41 $\pm$ 0.49 & 32.85 $\pm$ 0.83 & 32.66 $\pm$ 0.33 \\ \cline{2-8} 
& \Ttwo & 32.66 $\pm$ 1.66 & \textbf{33.73 $\pm$ 2.67} & 32.71 $\pm$ 2.13 & 28.52 $\pm$ 1.14 & 30.77 $\pm$ 0.92 & 33.10 $\pm$ 1.78 \\ \hline
\multirow{2}{*}{R=12} & \Tone & 30.43 $\pm$ 0.44 & 30.20 $\pm$ 0.72 & 30.94 $\pm$ 0.94 & 30.18 $\pm$ 0.73 & 30.04 $\pm$ 0.70 & \textbf{31.26 $\pm$ 1.08} \\ \cline{2-8} 
& \Ttwo & 30.76 $\pm$ 0.76 & 30.57 $\pm$ 1.09 & 30.86 $\pm$ 1.13 & 30.75 $\pm$ 2.27  & 27.79 $\pm$ 0.59  & \textbf{31.09 $\pm$ 0.78} \\ \hline
\multirow{2}{*}{R=16} & \Tone & 29.44  $\pm$ 0.69 & 29.11  $\pm$ 0.97 & 29.35  $\pm$ 0.73 & 29.82  $\pm$ 0.61  & 29.39  $\pm$ 0.90 & \textbf{30.38  $\pm$ 1.37}  \\ \cline{2-8} 
& \Ttwo  & 30.74  $\pm$ 0.77 & 29.73  $\pm$ 0.38 & \textbf{31.84  $\pm$ 0.62} & 27.15  $\pm$ 2.19 & 29.26  $\pm$ 1.37 & 30.27  $\pm$ 0.61 \\ \hline
\end{tabular}%
}
\label{tab: orientation_compare_IXI_recon}
\vskip 0.3em
\justify
{}
\end{table}

\clearpage
\newpage
\begin{table}[h]
\centering
\caption{Task-Optimal Progression Order for Reconstruction in the In vivo Knee Dataset: Volumetric PSNR (dB) measurements between the reconstructed and ground truth images in the validation set in the in vivo knee dataset are given as mean $\pm$ std. The measurements are provided for all possible progression orders: 1)~\ACS, 2)~\ASC, 3)~\SAC, 4)~\SCA, 5)~\CSA, 6)~\CAS~and acceleration factors: $R=4,8,12,16$. Boldface indicates the highest performing progression sequence.}
\resizebox{0.85\textwidth}{!}{%
\begin{tabular}{ccccccc}
\cline{2-7}
&\textbf{ \ACS} & \textbf{\ASC} & \textbf{\SAC} & \textbf{\SCA} & \textbf{\CSA} & \textbf{\CAS} \\ \hline
R=4 & 37.56 $\pm$ 1.09 & 37.91 $\pm$ 0.49 & \textbf{39.72 $\pm$ 1.63} & 38.73 $\pm$ 1.60 & 38.18 $\pm$ 1.99 & 38.27 $\pm$ 1.52  \\ \hline
R=8 & 38.80 $\pm$ 1.39 & 38.25 $\pm$ 1.13 & 36.38 $\pm$ 1.09 & \textbf{39.95 $\pm$ 0.64} & 37.44 $\pm$ 1.18 & 35.59 $\pm$ 1.19 \\ \hline
R=12 & 35.20 $\pm$ 0.04 & 38.36 $\pm$ 0.42 & 37.22 $\pm$ 1.29 & \textbf{39.14 $\pm$ 0.44} & 37.49 $\pm$ 0.81 & 37.61 $\pm$ 1.07 \\ \hline
R=16 & 37.26 $\pm$ 1.19 & 35.05 $\pm$ 3.30 & 37.38 $\pm$ 1.10 & 38.01 $\pm$ 0.27 & \textbf{38.28 $\pm$ 0.25} & 37.49 $\pm$ 0.94 \\ \hline
\end{tabular}
}
\label{tab: orientation_compare_multicoil_recon}
\end{table}

\clearpage
\newpage
\begin{table}[h]
\centering
\caption{Task-Optimal Progression Order for Synthesis in the IXI Dataset: Volumetric PSNR (dB) measurements between the synthesized and ground truth images in the validation set in the IXI dataset are given as mean $\pm$ std. The measurements are provided for all possible progression orders: 1)~\ACS, 2)~\ASC, 3)~\SAC, 4)~\SCA, 5)~\CSA, 6)~\CAS~and all many-to-one synthesis tasks: 1)~\TtwoPDTone, 2)~\TonePDTtwo, 3)~\ToneTtwoPD. Boldface indicates the highest performing progression sequence.}
\resizebox{0.85\textwidth}{!}{%
\begin{tabular}{ccccccc}
\cline{2-7}
&\textbf{ \ACS} & \textbf{\ASC} & \textbf{\SAC} & \textbf{\SCA} & \textbf{\CSA} & \textbf{\CAS} \\ \hline
{\TtwoPDTone} & 23.38 $\pm$ 2.38 & 23.75 $\pm$ 2.59 & 24.24 $\pm$ 3.81 & 24.01 $\pm$ 3.65 & \textbf{25.35 $\pm$ 2.23} & 24.90 $\pm$ 2.06 \\ \hline
{\TonePDTtwo} & 28.35 $\pm$ 1.65 & 28.34 $\pm$ 1.67 & 28.50 $\pm$ 1.16 & 28.37 $\pm$ 1.04 & \textbf{29.51 $\pm$ 1.41} & 28.62 $\pm$ 1.08 \\ \hline
{\ToneTtwoPD} & 30.42 $\pm$ 1.29 & \textbf{31.50 $\pm$ 1.61} & 30.46 $\pm$ 1.08 & 30.50 $\pm$ 1.14 & 31.44 $\pm$ 1.87 & 30.25 $\pm$ 1.51 \\ \hline
\end{tabular}
}
\label{tab: orientation_compare_many2one}
\end{table}

\clearpage
\newpage
\begin{table}[h]
\centering
\caption{Task-Optimal Progression Order for Synthesis in the In vivo Brain Dataset: Volumetric PSNR (dB) measurements between the synthesized and ground truth images in the validation set in the in vivo brain dataset are given as mean $\pm$ std. The measurements are provided for all possible progression orders: 1)~\ACS, 2)~\ASC, 3)~\SAC, 4)~\SCA, 5)~\CSA, 6)~\CAS~and all many-to-one synthesis tasks: 1)~\TtwoFlairTonecTone, 2)~\ToneFlairTonecTtwo, 3)~\ToneTtwoTonecFlair, 4)~\ToneTtwoFlairTonec. Boldface indicates the highest performing progression sequence.}
\resizebox{0.85\textwidth}{!}{%
\begin{tabular}{ccccccc}
\cline{2-7}
&\textbf{ \ACS} & \textbf{\ASC} & \textbf{\SAC} & \textbf{\SCA} & \textbf{\CSA} & \textbf{\CAS} \\ \hline
\TtwoFlairTonecTone & 24.48 $\pm$ 2.91 & 24.49 $\pm$ 2.85 & 24.48 $\pm$ 3.32 & 24.45 $\pm$ 3.27 & \textbf{25.63 $\pm$ 3.53} & 24.81 $\pm$ 3.14 \\ \hline
\ToneFlairTonecTtwo & 27.37 $\pm$ 2.96 & 27.14 $\pm$ 2.86 & 27.66 $\pm$ 2.91 & 27.19 $\pm$ 2.76 & 26.96 $\pm$ 2.81 & \textbf{27.68 $\pm$ 2.99} \\ \hline
\ToneTtwoTonecFlair & 24.93 $\pm$ 2.40 & \textbf{25.58 $\pm$ 2.47} & 24.90 $\pm$ 3.36 & 25.35 $\pm$ 3.45 & 25.51 $\pm$ 2.93 & \textbf{25.58 $\pm$ 3.09} \\ \hline
\ToneTtwoFlairTonec & 28.83 $\pm$ 2.08 & \textbf{29.88 $\pm$ 2.44} & 28.75 $\pm$ 2.54 & 28.83 $\pm$ 2.34 & 28.94 $\pm$ 2.26 & 28.43 $\pm$ 2.14 \\ \hline
\end{tabular}
}
\label{tab: orientation_compare_many2one_hacet}
\end{table}

\clearpage
\newpage
\begin{table}[h]
\centering
\caption{vGAN Architecture Optimization for Reconstruction Tasks: Volumetric PSNR (dB) measurements between the reconstructed and ground truth images in the validation set of IXI are given as mean $\pm$ std. Measurements are provided for vGAN implemented using various combinations of number of residual layers, number of filters and convolutional kernel size. Three different network depths (3-6-9 residual blocks in the generator), three different numbers of filters (1x, 1.25x, 1.5x that of the base model) and three different kernel sizes ($3\times3\times3$ - $5\times5\times5$ - $7\times7\times7$) were examined. Note that in deeper networks higher numbers of filters cannot be implemented due to constraints related to the GPU memory. Boldface indicates the highest performing vGAN architecture.}
\resizebox{0.55\textwidth}{!}{%
\begin{tabular}{ccccc}
\cline{2-5}
& $n_f$ &\textbf{ $3\times3\times3$ } & \textbf{$5\times5\times5$} & \textbf{$7\times7\times7$}  \\ \hline
9 ResNet & 1x & \textbf{30.48 $\pm$ 0.61} & 27.75 $\pm$ 0.42 & 27.66 $\pm$ 1.21 \\ \hline
\multirow{2}{*}{6 ResNet} & 1x & 28.03 $\pm$ 1.36 & 28.55 $\pm$ 0.41 & 26.26 $\pm$ 1.77 \\ \cline{2-5}
 & 1.25x & 27.50 $\pm$ 1.31 & 28.71 $\pm$ 0.60 & 24.83 $\pm$ 1.14 \\ \hline
 \multirow{3}{*}{3 ResNet} & 1x & 29.56 $\pm$ 0.45 & 28.03 $\pm$ 1.12 & 28.01 $\pm$ 0.77 \\ \cline{2-5}
 & 1.25x & 28.14 $\pm$ 1.10 & 27.61 $\pm$ 1.25 & 29.42 $\pm$ 0.76 \\ \cline{2-5}
 & 1.5x & 28.62 $\pm$ 0.80 & 30.48 $\pm$ 0.38 & 27.54 $\pm$ 1.31 \\ \hline

\end{tabular}
}
\label{tab: vgan_opt_recon}
\end{table}

\clearpage
\newpage
\begin{table}[h]
\centering
\caption{vGAN Architecture Optimization for Synthesis Tasks: Volumetric PSNR (dB) measurements between the synthesized and ground truth images in the validation set of IXI are given as mean $\pm$ std. Measurements are provided for vGAN implemented using various combinations of number of residual layers, number of filters and convolutional kernel size. Three different network depths (3-6-9 residual blocks in the generator), three different numbers of filters (1x, 1.25x, 1.5x that of the base model) and three different kernel sizes ($3\times3\times3$ - $5\times5\times5$ - $7\times7\times7$) were examined. Note that in deeper networks higher numbers of filters cannot be implemented due to constraints related to the GPU memory. Boldface indicates the highest performing vGAN architecture.}
\resizebox{0.55\textwidth}{!}{%
\begin{tabular}{ccccc}
\cline{2-5}
& $n_f$ &\textbf{ $3\times3\times3$ } & \textbf{$5\times5\times5$} & \textbf{$7\times7\times7$}  \\ \hline
9 ResNet & 1x & \textbf{26.98 $\pm$ 0.73} & 24.80 $\pm$ 1.56 & 24.67 $\pm$ 0.75 \\ \hline
\multirow{2}{*}{6 ResNet} & 1x & 23.82 $\pm$ 1.11 & 24.54 $\pm$ 0.51 & 22.97 $\pm$ 0.87 \\ \cline{2-5}
 & 1.25x & 21.51 $\pm$ 1.29 & 23.64 $\pm$ 0.67 & 24.23 $\pm$ 0.87 \\ \hline
 \multirow{3}{*}{3 ResNet} & 1x & 22.12 $\pm$ 0.60 & 23.86 $\pm$ 0.53 & 21.30 $\pm$ 1.08 \\ \cline{2-5}
 & 1.25x & 23.30 $\pm$ 0.58 & 21.08 $\pm$ 1.56 & 22.55 $\pm$ 0.82 \\ \cline{2-5}
 & 1.5x & 19.56 $\pm$ 1.21 & 19.71 $\pm$ 1.19 & 22.73 $\pm$ 0.54 \\ \hline

\end{tabular}
}
\label{tab: vgan_opt_synthesis}
\end{table}

\clearpage
\newpage
\begin{table}[h]
  \centering
  \caption{Model Complexity Analysis for Reconstruction in the IXI Dataset: Volumetric PSNR (dB) and SSIM (\%) measurements between the reconstructed and ground truth images in the test set of the IXI dataset are given as mean $\pm$ std for reconstruction of \Tone-weighted acquisitions undersampled at $R=8$. The measurements are reported for ProvoGAN and sGAN while varying the complexity of the convolutional layers in both models by $n_f\in\{1/16,1/9,1/4,1,4,9,16\}$, yielding seven distinct ProvoGAN-sGAN pairs: ProvoGAN($n_f$)-sGAN($n_f$) with $n_f$ times folded number of learnable network weights. Boldface indicates the highest performing method.}
  \resizebox{0.95\textwidth}{!}{%
\begin{tabular}{ccccccccc}
      \cline{3-9}
      &      & $1/16$  & $1/9$ & $1/4$ & 1 & 4 & 9 & 16 \\ \hline
      \multirow{2}{*}{ProvoGAN} & PSNR & \textbf{31.53 $\pm$ 1.31} & \textbf{31.02 $\pm$ 0.96} & \textbf{30.96 $\pm$ 0.85} & \textbf{31.38 $\pm$ 1.26} & \textbf{32.16 $\pm$ 1.80}  & \textbf{32.23 $\pm$ 1.81} & \textbf{32.52 $\pm$ 1.76}    \\ \cline{2-9} 
      & SSIM & \textbf{93.74 $\pm$ 1.10} & \textbf{91.29 $\pm$ 1.02} & \textbf{90.62 $\pm$ 1.10} & \textbf{94.93 $\pm$ 0.86} & \textbf{95.68 $\pm$ 0.97}  & \textbf{95.78 $\pm$ 1.00} & \textbf{95.81 $\pm$ 1.05}       \\ \hline
      \multirow{2}{*}{sGAN-A} & PSNR & 29.58 $\pm$ 0.99 & 30.14 $\pm$ 0.93 & 30.09 $\pm$ 1.45 & 30.08 $\pm$ 1.32 & 30.17 $\pm$ 1.56  & 30.82 $\pm$ 1.61 & 30.96 $\pm$ 1.53    \\ \cline{2-9} 
      & SSIM & 89.07 $\pm$ 1.19 & 89.84 $\pm$ 1.03 & 90.50 $\pm$ 1.07 & 91.18 $\pm$ 1.01 & 91.47 $\pm$ 1.15  & 91.73 $\pm$ 1.20 & 91.66 $\pm$ 1.13       \\ \hline
      
    \end{tabular}
  }
  \label{tab: IXI_recon_T1_8x_parametre_compare}
  \vskip 0.3em
\end{table}

\newpage
\clearpage
\begin{table}[h]
  \centering
  \caption{Model Complexity Analysis for Synthesis in the IXI Dataset: Volumetric PSNR (dB) and SSIM (\%) measurements between the synthesized and ground truth images in the test set of the IXI dataset are given as mean $\pm$ std for \Tone-weighted image synthesis from \Ttwo- and PD-weighted images. The measurements are reported for ProvoGAN and sGAN while varying the complexity of the convolutional layers in both models by $n_f\in\{1/16,1/9,1/4,1,4,9,16\}$, yielding seven distinct ProvoGAN-sGAN pairs: ProvoGAN($n_f$)-sGAN($n_f$) with $n_f$ times folded number of learnable network weights. sGAN-A denotes the sGAN model trained in the axial orientation, sGAN-C in the coronal orientation, and sGAN-S in the sagittal orientation. Boldface indicates the highest performing method.}
  \resizebox{0.95\textwidth}{!}{%
\begin{tabular}{ccccccccc}
      \cline{3-9}
      &      & $1/16$  & $1/9$ & $1/4$ & 1 & 4 & 9 & 16 \\ \hline
      \multirow{2}{*}{ProvoGAN} & PSNR & \textbf{24.25 $\pm$ 2.12} & \textbf{24.32 $\pm$ 2.11} & \textbf{24.10 $\pm$ 2.14} & \textbf{24.15 $\pm$ 2.80} & \textbf{24.74 $\pm$ 2.59}  & \textbf{25.27 $\pm$ 2.65} & \textbf{25.36 $\pm$ 2.14}    \\ \cline{2-9} 
      & SSIM & \textbf{88.61 $\pm$ 4.16} & \textbf{88.49 $\pm$ 4.30} & \textbf{88.53 $\pm$ 4.23} & \textbf{90.33 $\pm$ 4.47} & \textbf{89.51 $\pm$ 4.39}  & \textbf{90.05 $\pm$ 4.36} & \textbf{89.92 $\pm$ 4.34}       \\ \hline
      \multirow{2}{*}{sGAN-A} & PSNR & 23.13 $\pm$ 2.03 & 23.13 $\pm$ 1.95 & 22.90 $\pm$ 1.78 & 23.20 $\pm$ 2.08 & 23.63 $\pm$ 2.22  & 23.96 $\pm$ 87.34 & 24.22 $\pm$ 1.76     \\ \cline{2-9} 
      & SSIM & 84.84 $\pm$ 3.52 & 84.66 $\pm$ 3.61 & 85.08 $\pm$ 3.71 & 85.81 $\pm$ 3.95 & 86.51 $\pm$ 4.03  & 87.34 $\pm$ 3.95 & 87.38 $\pm$ 3.96       \\ \hline
      \multirow{2}{*}{sGAN-C} & PSNR & 22.75 $\pm$ 2.51 & 22.82 $\pm$ 2.19 & 22.84 $\pm$ 2.04 & 22.58 $\pm$ 2.11 & 23.32 $\pm$ 2.35  & 23.77 $\pm$ 2.27 & 23.32 $\pm$ 2.51     \\ \cline{2-9} 
      & SSIM & 85.24 $\pm$ 3.66 & 85.75 $\pm$ 3.93 & 86.21 $\pm$ 3.88 & 86.60 $\pm$ 4.05 & 87.47 $\pm$ 3.94  & 88.19 $\pm$ 4.04 & 88.10 $\pm$ 4.10       \\ \hline
      \multirow{2}{*}{sGAN-S} & PSNR & 23.08 $\pm$ 1.63 & 22.33 $\pm$ 1.81 & 23.74 $\pm$ 1.76 & 23.65 $\pm$ 1.98 & 24.07 $\pm$ 2.14  & 24.14 $\pm$ 2.18 & 24.30 $\pm$ 2.17     \\ \cline{2-9} 
      & SSIM & 85.23 $\pm$ 3.65 & 81.14 $\pm$ 4.65 & 87.43 $\pm$ 4.26 & 87.71 $\pm$ 4.15 & 88.76 $\pm$ 4.74  & 88.70 $\pm$ 4.81 & 88.90 $\pm$ 4.84       \\ \hline       
      
    \end{tabular}
  }
  \label{tab: IXI_many2one_T1_syn_parametre_compare}
\end{table}

\clearpage
\newpage

\begin{table}[h]
  \centering
  \caption{Comparison of Single versus Multi-Cross-Section ProvoGAN and sGAN Models for Reconstruction in the IXI Dataset: Volumetric PSNR (dB) and SSIM (\%) measurements between the reconstructed and ground truth images in the test set of the IXI dataset are given as mean $\pm$ std. The measurements are reported for the proposed ProvoGAN model, its multi-cross-section variant, the cross-sectional sGAN model and its multi-cross-section variant for all single-coil reconstruction tasks: \Tone-weighted and \Ttwo-weighted image reconstruction at $R=4,8,12,16$. sGAN and sGAN(multi) are trained in the axial orientation given axial readout direction. Boldface indicates the highest performing method.}
  \resizebox{0.95\textwidth}{!}{%
    \begin{tabular}{cccccccccc}
      \cline{3-10}
      &     & \multicolumn{2}{c}{R=4} & \multicolumn{2}{c}{R=8} & \multicolumn{2}{c}{R=12} & \multicolumn{2}{c}{R=16} \\ \cline{3-10}
      &      & \Tone & \Ttwo & \Tone & \Ttwo & \Tone & \Ttwo & \Tone & \Ttwo  \\ \hline
      \multirow{2}{*}{ProvoGAN} & PSNR & \textbf{35.25 $\pm$ 1.78} & 35.50 $\pm$ 2.62 & \textbf{31.83 $\pm$ 1.26} & \textbf{33.49 $\pm$ 2.21} & \textbf{29.67 $\pm$ 0.91} & \textbf{30.28 $\pm$ 1.31} & \textbf{29.14 $\pm$ 1.09} & \textbf{30.66 $\pm$ 1.60}  \\ \cline{2-10} 
      & SSIM & \textbf{96.73 $\pm$ 0.57} & 96.08 $\pm$ 1.07 & 94.93 $\pm$ 0.86 & \textbf{95.92 $\pm$ 1.01} & \textbf{92.48 $\pm$ 0.90} & 91.96 $\pm$ 1.40 & \textbf{91.40 $\pm$ 1.09} & \textbf{93.74 $\pm$ 1.35}      \\ \hline
      \multirow{2}{*}{ProvoGAN(multi)} & PSNR & 34.65 $\pm$ 1.61 & \textbf{36.36 $\pm$ 2.05} & 31.71 $\pm$ 1.31 & 32.63 $\pm$ 1.58 & 29.26 $\pm$ 0.92 & 30.24 $\pm$ 1.05 & 28.44 $\pm$ 1.05 & 29.29 $\pm$ 1.09  \\ \cline{2-10} 
      & SSIM & 96.27 $\pm$ 0.63 & \textbf{96.73 $\pm$ 0.70} & \textbf{95.08 $\pm$ 0.95} & 94.45 $\pm$ 0.91 & 88.58 $\pm$ 1.26 & \textbf{92.10 $\pm$ 1.26} & 88.11 $\pm$ 1.50 & 91.74 $\pm$ 1.27    \\ \hline
      \multirow{2}{*}{sGAN} & PSNR & 33.85 $\pm$ 1.29 & 32.95 $\pm$ 1.50 & 30.08 $\pm$ 1.32 & 32.24 $\pm$ 2.14 & 27.34 $\pm$ 1.06 & 28.48 $\pm$ 1.06 & 26.73 $\pm$ 1.53 & 29.05 $\pm$ 1.04  \\ \cline{2-10}  
      & SSIM & 93.21 $\pm$ 0.78 & 86.44 $\pm$ 1.23 & 91.18 $\pm$ 1.01 & 90.47 $\pm$ 0.95 & 86.23 $\pm$ 1.36 & 79.50 $\pm$ 2.15 & 85.23 $\pm$ 1.75 & 83.38 $\pm$ 1.22      \\ \hline
\multirow{2}{*}{sGAN(multi)} & PSNR & 33.71 $\pm$ 1.35 & 34.11 $\pm$ 1.58 & 31.00 $\pm$ 1.05 & 29.51 $\pm$ 0.93 & 28.43 $\pm$ 1.04 & 29.28 $\pm$ 1.27 & 27.80 $\pm$ 0.96 & 29.83 $\pm$ 1.28  \\ \cline{2-10}  
      & SSIM & 93.69 $\pm$ 0.65 & 88.86 $\pm$ 1.24 & 92.17 $\pm$ 0.77 & 80.54 $\pm$ 1.88 & 88.03 $\pm$ 1.07 & 82.51 $\pm$ 1.69 & 87.07 $\pm$ 1.32 & 85.43 $\pm$ 1.22      \\ \hline
      
    \end{tabular}
  }
  \label{tab: IXI_recon_compare_with_multi_cross}
  \vskip 0.3em
\end{table}

\clearpage
\newpage

\begin{table}[h]
  \centering
  \caption{Comparison of Single and Multi-Cross-Section ProvoGAN and sGAN Models for Synthesis in the IXI Dataset: Volumetric PSNR (dB) and SSIM (\%) measurements between the synthesized and ground truth images in the test set in the IXI dataset are given as mean $\pm$ std. The measurements are reported for the proposed ProvoGAN model, its multi-cross-section variant, the cross-sectional sGAN model and its multi-cross-section variant for all many-to-one synthesis tasks:  1)~\TtwoPDTone, 2)~\TonePDTtwo, 3)~\ToneTtwoPD. sGAN-A and sGAN(multi)-A denote the models trained in the axial orientation, sGAN-C and sGAN(multi)-C denote the models trained in the coronal orientation, and sGAN-S and sGAN(multi)-S denote the models trained in the sagittal orientation. Boldface indicates the highest performing method.}
  \resizebox{0.65\textwidth}{!}{%
    \begin{tabular}{ccccc}
      \cline{3-5}
      &      & \TtwoPDTone & \TonePDTtwo & \ToneTtwoPD  \\ \hline
      \multirow{2}{*}{ProvoGAN} & PSNR & 24.15 $\pm$ 2.80 & \textbf{28.97 $\pm$ 2.91} & \textbf{29.81 $\pm$ 2.96}  \\ \cline{2-5} 
      & SSIM &  90.33 $\pm$ 4.47 & \textbf{94.17 $\pm$ 4.16} & \textbf{95.41 $\pm$ 2.75}      \\ \hline
      \multirow{2}{*}{ProvoGAN(multi)} & PSNR & 24.23 $\pm$ 2.56 & 28.32 $\pm$ 2.69 & 29.64 $\pm$ 2.42  \\ \cline{2-5} 
      & SSIM &  \textbf{90.49 $\pm$ 4.50} & 93.52 $\pm$ 3.93 & 94.96 $\pm$ 2.62      \\ \hline
      \multirow{2}{*}{sGAN-A} & PSNR & 23.20 $\pm$ 2.08 & 27.64 $\pm$ 2.60 & 27.70 $\pm$ 2.20  \\ \cline{2-5} 
      & SSIM &  85.81 $\pm$ 3.95 & 92.94 $\pm$ 4.20 & 93.64 $\pm$ 3.00      \\ \hline
      \multirow{2}{*}{sGAN(multi)-A} &  PSNR & 23.55 $\pm$ 2.24 & 27.87 $\pm$ 2.54 & 28.69 $\pm$ 2.38  \\ \cline{2-5} 
      & SSIM &  86.53 $\pm$ 4.26 & 93.09 $\pm$ 3.89 & 93.89 $\pm$ 3.07      \\ \hline
      \multirow{2}{*}{sGAN-C} & PSNR & 22.56 $\pm$ 2.11 & 27.74 $\pm$ 2.67 & 29.00 $\pm$ 2.41  \\ \cline{2-5} 
      & SSIM &  86.60 $\pm$ 4.05 & 92.67 $\pm$ 4.31 & 94.21 $\pm$ 2.99      \\ \hline
      \multirow{2}{*}{sGAN(multi)-C} &  PSNR & 22.96 $\pm$ 2.07 & 27.97 $\pm$ 2.61 & 29.18 $\pm$ 2.42  \\ \cline{2-5}
      & SSIM &  87.53 $\pm$ 4.27 & 92.97 $\pm$ 4.06 & 94.46 $\pm$ 2.97      \\ \hline
      \multirow{2}{*}{sGAN-S} & PSNR & 23.65 $\pm$ 1.98 & 27.93 $\pm$ 2.19 & 27.12 $\pm$ 1.61  \\ \cline{2-5}
      & SSIM &  87.71 $\pm$ 4.15 & 93.28 $\pm$ 2.88 & 92.67 $\pm$ 2.95      \\ \hline
      \multirow{2}{*}{sGAN(multi)-S} &  PSNR & \textbf{24.33 $\pm$ 2.29} & 28.08 $\pm$ 2.18 & 28.36 $\pm$ 1.98  \\ \cline{2-5}
      & SSIM &  88.67 $\pm$ 4.45 & 93.43 $\pm$ 2.90 & 93.80 $\pm$ 2.87      \\ \hline
    \end{tabular}
  }
  \label{tab: IXI_many2one_compare_with_multi_cross}
\end{table}

\clearpage
\newpage

\begin{table}[h]
  \centering
  \caption{Data Efficiency of ProvoGAN and vGAN Models for Synthesis in the IXI Dataset: Volumetric PSNR (dB) and SSIM (\%) measurements between the synthesized and ground truth images in the test set in the IXI dataset are given as mean $\pm$ std. Measurements are reported for the proposed ProvoGAN and competing vGAN methods trained with varying number of subjects for the following synthesis tasks: 1)~\TtwoPDTone, 2)~\TonePDTtwo, 3)~\ToneTtwoPD. ProvoGAN-($n_T$) and vGAN($n_T$) denote models trained with $n_T$ subjects. }
  \resizebox{0.95\textwidth}{!}{%
    \begin{tabular}{ccccccc}
    \cline{2-7}
    & \multicolumn{2}{c}{\TtwoPDTone} & \multicolumn{2}{c}{\TonePDTtwo}  & \multicolumn{2}{c}{\ToneTtwoPD}   \\ \cline{2-7}\vspace{-1mm}
    & PSNR & SSIM & PSNR & SSIM & PSNR & SSIM    \\ \hline
    ProvoGAN(25) & 22.53 $\pm$ 3.39 & 89.11 $\pm$ 4.62 & 28.35 $\pm$ 2.87 & 93.23 $\pm$ 4.51 & 29.96 $\pm$ 3.06 & 95.45 $\pm$ 3.06   \\ \hline
    ProvoGAN(15) & 22.21 $\pm$ 3.13 & 87.67 $\pm$ 4.40 & 28.17 $\pm$ 2.80 & 93.17 $\pm$ 4.56 & 29.95 $\pm$ 2.60 & 95.50 $\pm$ 3.04   \\ \hline
    ProvoGAN(5) & 21.71 $\pm$ 3.26 & 85.25 $\pm$ 4.12 & 27.74 $\pm$ 2.73 & 92.87 $\pm$ 4.41 & 29.35 $\pm$ 2.53 & 94.80 $\pm$ 3.24   \\ \hline
    vGAN(25) & 22.69 $\pm$ 2.88 & 86.33 $\pm$ 4.06 & 26.67 $\pm$ 2.48 & 91.83 $\pm$ 4.34 & 27.09 $\pm$ 1.35 & 92.95 $\pm$ 2.76   \\ \hline
    vGAN(15) & 21.73 $\pm$ 3.28 & 85.13 $\pm$ 4.08 & 26.17 $\pm$ 2.16 & 91.34 $\pm$ 4.06 & 26.90 $\pm$ 1.58 & 92.75 $\pm$ 2.53   \\ \hline
    vGAN(5) & 21.75 $\pm$ 2.39 & 83.33 $\pm$ 3.30 & 24.98 $\pm$ 1.43 & 89.23 $\pm$ 3.46 & 26.17 $\pm$ 1.30 & 91.60 $\pm$ 2.15   \\ \hline
  
    \end{tabular}
  }
  \label{tab: IXI_many2one_compare_data_efficiency}
\end{table}

\clearpage
\newpage

\begin{table}[h]
\caption{The proposed ProvoGAN and competing sGAN and vGAN methods are evaluated in terms instantaneous model complexity (millions of parameters, M), GPU VRAM use (gigabytes, GB), FLOPs (billions of floating point operation, G), and train duration (hours). The model complexity of the methods is given as $(p_g,p_d)$, where $p_g$ denotes the number of free parameters in the generator and $p_d$ denotes the number of free parameters in the discriminator. The number of operations (FLOPs) of the methods is given as $(F_g,F_d)$, where $F_g$ denotes the number of operations in the generator and $F_d$ denotes the number of operations in the discriminator. The input-output volume sizes and the number training subjects are also reported.}
\centering
\begin{subtable}[h]{\textwidth}
\centering
\captionsetup{justification=centering}
\caption{Reconstruction in the IXI Dataset (Input Volume Size: $256\times150\times256$, Output Volume Size: $256\times150\times256$, Number of Training Subjects: $37$ )}
\resizebox{0.6\textwidth}{!}{
\begin{tabular}{c|ccc}
& sGAN & vGAN & ProvoGAN \\ \hline
Model Complexity (M) & (1.60, 0.39)     &  (4.8, 1.17)    &  (1.6, 0.39)        \\ \hline
GPU VRAM Usage (GB)        &  0.6  & 19    & 0.6         \\ \hline
FLOPs (G) & (4.66, 0.26)  & (1428.30, 31.79) & (17.55, 0.98) \\ \hline
Train Duration (hours)           &   11   &  20    &  33        \\ \hline
\end{tabular}
}
\end{subtable}
\vfill
\vskip 5mm
\centering
\begin{subtable}[h]{\textwidth}
\centering
\captionsetup{justification=centering}
\caption{Reconstruction in the In vivo Knee Dataset (Input Subvolume Size: $320\times320\times256$, Output Subvolume Size:  $320\times320\times256$, Number of Training Subjects: $12$ )}
\resizebox{0.6\textwidth}{!}{
\begin{tabular}{c|ccc}
& sGAN & vGAN & ProvoGAN \\ \hline
Model Complexity (M) & (1.60, 0.39)     &  (4.8, 1.17)    &  (1.6, 0.39)        \\ \hline
GPU VRAM Usage (GB)    &  0.8    &   19   & 0.8         \\ \hline
FLOPs (G) & (10.07, 0.58)  & (3860.27, 88.73) & (33.12, 1.90) \\ \hline
Train Duration (hours)       &  7    & 28     &  22        \\ \hline
\end{tabular}
}
\end{subtable}
\vfill
\vskip 5mm
\centering
\centering
\begin{subtable}[h]{\textwidth}
\centering
\captionsetup{justification=centering}
\caption{Synthesis in the IXI Dataset (Input Volume Size: $2\times192\times160\times160$, Output Volume Size:  $192\times160\times160$, Number of Training Subjects: $37$ )}
\resizebox{0.6\textwidth}{!}{
\begin{tabular}{c|ccc}
& sGAN & vGAN & ProvoGAN \\ \hline
Model Complexity (M) & (1.60, 0.39)     &  (4.8, 1.17)    &  (1.6, 0.39)        \\ \hline
GPU VRAM Usage (GB)   & 0.6   &  9   &   0.6       \\ \hline
FLOPs (G) & (3.57, 0.20)  & (723.80, 16.27) & (10.77, 0.60) \\ \hline
Train Duration (hours) & 4     &  12    &  12        \\ \hline
\end{tabular}
}
\end{subtable}
\vfill
\vskip 5mm
\centering
\begin{subtable}[h]{\textwidth}
\centering
\captionsetup{justification=centering}
\caption{Synthesis in the In-vivo Brain Dataset (Input Volume Size: $3\times192\times160\times160$, Output Volume Size:  $192\times160\times160$, Number of Training Subjects: $237$ )}
\resizebox{0.6\textwidth}{!}{
\begin{tabular}{c|ccc}
& sGAN & vGAN & ProvoGAN \\ \hline
Model Complexity (M) & (1.60, 0.39)     &  (4.8, 1.17)    &  (1.6, 0.39)        \\ \hline
GPU VRAM Usage (GB)   &  0.6   &   9   &  0.6        \\ \hline
FLOPs (G) & (3.60, 0.20)  & (764.26, 16.98) & (10.87, 0.61) \\ \hline
Train Duration (hours) &  6    &  16    &  18        \\ \hline
\end{tabular}
}
\end{subtable}
\label{tab: complexity_table_comp}
\end{table}

\end{document}